\documentclass[12pt,a4paper,twoside]{report}

\usepackage[a4paper,hmargin=2.8cm,vmargin=2.0cm,includeheadfoot]{geometry}
\usepackage{textpos}
\usepackage{natbib} 
\usepackage{tabularx,longtable,multirow,subfigure,caption}
\usepackage{fncylab} 
\usepackage{fancyhdr} 
\usepackage{url} 
\usepackage[english]{babel}
\usepackage{amsmath}
\usepackage{graphicx}
\usepackage{dsfont}
\usepackage{epstopdf} 
\usepackage{backref} 
\usepackage{array}
\usepackage{latexsym}
\usepackage[pdftex,pagebackref,hypertexnames=false,colorlinks]{hyperref} 

\hypersetup{pdftitle={},
  pdfsubject={}, 
  pdfauthor={},
  pdfkeywords={}, 
  pdfstartview=FitH,
  pdfpagemode={UseOutlines},
  bookmarksnumbered=true, bookmarksopen=true, colorlinks,
    citecolor=black,%
    filecolor=black,%
    linkcolor=black,%
    urlcolor=black}

\usepackage[all]{hypcap}

\def\@makechapterhead#1{%
  \vspace*{10\p@}%
  {\parindent \z@ \raggedright \sffamily
    \interlinepenalty\@M
    \Huge\bfseries \thechapter \space\space #1\par\nobreak
    \vskip 30\p@
  }}

\def\@makeschapterhead#1{%
  \vspace*{10\p@}%
  {\parindent \z@ \raggedright
    \sffamily
    \interlinepenalty\@M
    \Huge \bfseries  #1\par\nobreak
    \vskip 30\p@
  }}

\allowdisplaybreaks
\usepackage{amssymb}
\usepackage{listings}
\newcommand{\estiloPython}{
\lstset{
    frame=tb,
    language=Python,
    aboveskip=3mm,
    belowskip=3mm,
    basicstyle=\ttfamily\small,
    commentstyle=\color{verde},
    morecomment=[s][\color{blue}]{/**}{*/},
    extendedchars=true,
    showspaces=false,
    showstringspaces=false,
    numbers=left,
    numberstyle=\tiny,
    breaklines=true,
    breakautoindent=true,
    captionpos=b,
    xleftmargin=0pt,
    tabsize=2
}

}
\usepackage[table,xcdraw]{xcolor}

\usepackage[utf8x]{inputenc}
\usepackage[T1]{fontenc}
\usepackage{float}
\usepackage{booktabs}
\usepackage{multirow}
\usepackage{booktabs}
\usepackage{tabularx}
\usepackage{ltablex}
\usepackage{tikz}
\def\checkmark{\tikz\fill[scale=0.4](0,.35) -- (.25,0) -- (1,.7) -- (.25,.15) -- cycle;}
\newcolumntype{s}{>{\hsize=0.30\hsize}X}

\usepackage[colorinlistoftodos]{todonotes}

\usepackage{datetime}
\newdateformat{monthyeardate}{%
  \monthname[\THEMONTH], \THEYEAR}
\title{Aff-Wild Database and AffWildNet}
\author{Mengyao Liu}

\begin{document}

\begin{titlepage}

\newcommand{\HRule}{\rule{\linewidth}{0.5mm}} 


\includegraphics[width=4cm]{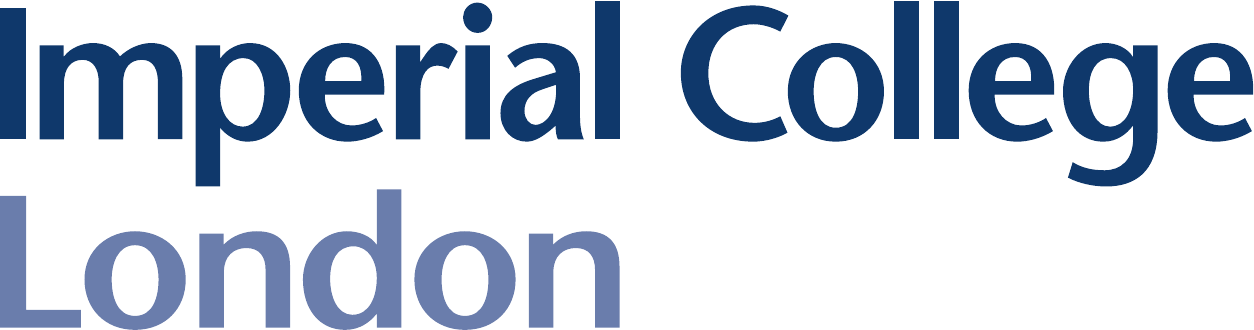}\\[0.5cm] 
 

\center 


\textsc{\Large Imperial College London}\\[0.5cm] 
\textsc{\large Department of Computing}\\[0.5cm] 

\makeatletter
\HRule \\[0.4cm]
{ \huge \bfseries \@title}\\[0.4cm] 
\HRule \\[1.5cm]
 

\begin{minipage}{0.4\textwidth}
\begin{flushleft} \large
\emph{Author:}\\
\@author 
\end{flushleft}
\end{minipage}
~
\begin{minipage}{0.4\textwidth}
\begin{flushright} \large
\emph{Supervisor:} \\
Mr. Dimitrios Kollias \\[1.2em] 
\end{flushright}
\end{minipage}\\[2cm]
\makeatother



\vfill 
Submitted in partial fulfillment of the requirements for the MSc degree in
Computing Science/Software Engineering of Imperial College London\\[0.5cm]

\monthyeardate\today

\end{titlepage}

\pagenumbering{roman}
\clearpage{\pagestyle{empty}\cleardoublepage}
\setcounter{page}{1}
\pagestyle{fancy}

\begin{abstract}
In the context of Human Computer Interaction(HCI), building an automatic system to recognize affect of human facial expression in real-world condition is very crucial to make machine interact with a man in a naturalistic way. However, existing databases of facial emotion usually contain facial expression in the limited scenario under well-controlled condition. Aff-Wild is currently the largest database consisting of spontaneous facial expression in the wild annotated with valence and arousal value. \newline

The first contribution of this project is the completion of extending Aff-Wild database which is fulfilled by collecting videos from YouTube on which the videos have spontaneous facial expressions in the wild, annotating videos with valence and arousal ranging in [-1.0, 1.0], detecting faces in frames using FFLD2 detector and partitioning the whole data set into train, validate and test set, having 527056, 94223 and 135145 frames respectively. The diversity is guaranteed regarding age, ethnicity and the values of valence and arousal. The ratio of male to female is close to 1:1. \newline

Regarding the techniques used to build the automatic system, deep learning is outstanding since almost all winning methods in emotion challenges adopt deep neural network techniques. The second contribution of this project is that an end-to-end deep learning neural network is constructed to have joint CNN and RNN block and gives the estimation on valence and arousal for each frame in sequential input data. VGGFace, ResNet, DenseNet with the corresponding pre-trained model for CNN block and LSTM, GRU, IndRNN, Attention mechanism for RNN block are experimented aiming to find the best combination. Fine tuning and transfer learning techniques are also tried out. By comparing the CCC evaluation value on test data, the best model is found out to be VGGFace pre-trained model connected with two layers GRU with attention mechanism. The test performance of this model achieves 0.555 CCC value for valence with sequence length 80 and 0.499 CCC value for arousal with sequence length 70.\newline
\end{abstract}

\cleardoublepage
\section*{Acknowledgments}
I am very grateful to my supervisor, Mr. Kollias. His passion for facial emotion recognition has deeply infected me. Through out the project, I received professional guidance and encouragement from him.\newline

I would also like to thank Dr. Anandha Gopalan for giving me MS cloud computing credits.\newline

Thanks to CSG for giving me detailed response when I encounter problems with GPU clusters.\newline

Thanks to Bingwen Wu for lending me the Surface Book for annotating videos.\newline

Thanks to all my friends and family.\newline

\clearpage{\pagestyle{empty}\cleardoublepage}


\fancyhead[RE,LO]{\sffamily {Table of Contents}}
\tableofcontents

\clearpage{\pagestyle{empty}\cleardoublepage}
\pagenumbering{arabic}
\setcounter{page}{1}
\fancyhead[LE,RO]{\slshape \rightmark}
\fancyhead[LO,RE]{\slshape \leftmark}

\chapter{Introduction}
This chapter mainly talks about the motivation and aims for this project in Section \ref{sec:motiaim}. A brief summary of the contribution is demonstrated in Section \ref{sec:contribution}. In the end, the arrangement of the whole report is listed in Section \ref{sec:arrangement}.

\section{Motivation and Objective}
\label{sec:motiaim}
In the context of Human Computer Interaction(HCI), understanding human emotions is essential to analyze human behaviours. Modern research on facial affect recognition aims to build an automatic system to interact with the human in a naturalistic way (\cite{goudelis2013exploring,mylonas2009using}). There are a lot of practical scenarios for this task. For instance, in an online education system, students' facial expression may show whether or not they catch up with the material taught or how much they are satisfied with the lecture content. If the facial expression of students can be analyzed automatically, the education system can give specific adjust to a particular student. Moreover, mental health diagnoses can also be accomplished by such a facial emotion analysis system \citep{kollias13,tagaris1,tagaris2}. However, building a system to analyze human facial emotions automatically is extremely difficult due to facial expressions are complex to describe and environment dependent. To achieve this task, such a system should be capable of interpreting spontaneous facial behaviour under uncontrolled conditions like diverse recording quality or various lighting conditions. Existing databases related to facial emotions are limited in studying behaviours in limited scenarios and under highly controlled recording conditions. Some representatives are listed here. SEMAINE \citep{mckeown2012semaine} database contains machine and human interactions under lab condition; RECOLA \citep{Ringeval2013IntroducingInteractions} database recorded video conference of pairs of people in well-controlled conditions; SEWA\footnote{http://sewaproject.eu} database consists of audiovisual data of people talking about commercial advertisements. Concerning the measurements used to model the facial emotions, there are different approaches to measure facial expression like using Action Units(AUs) \citep{friesen1978facial} and Seven Basic Emotion Categories \citep{ekman1982emotions}. Nonetheless, using the dimensional value like valence and arousal(VA) \citep{russell1980circumplex} to measure the emotion is more subtle. Besides, valence and arousal also cover a much broader spectrum of emotions in comparison to Seven Basic Emotions or AUs. The Aff-Wild database is currently the ideal database in facial emotion recognition. Aff-Wild database is the largest existing in the wild facial expression database annotated with VA values, containing spontaneous facial expressions under uncontrolled conditions since the included videos are sourced from YouTube. The \textbf{first objective} of this project is to extend Aff-Wild database. \newline

When it comes to the potential techniques which can be used to build the emotion analysis system, deep learning technique \citep{kollias10,kollias11,kollias12} is considered since this technique is applied successfully in recent emotion challenges like EmotiW \citep{dhall2017individual}, FERA \citep{valstar2015fera}, AVEC \citep{ringeval2017avec} and Aff-Wild \citep{kollias1,kollias3}. In \cite{jaiswal2016deep}, \cite{kollias7} and \cite{Chen2017MultimodalRecognition}, recurrent neural network is proved to have impressive performance in modelling temporal dynamics in sequential data. Moreover, deep CNNs, like VGGNet\citep{simonyan2014very}, ResNet\citep{he2016deep} and DenseNet \citep{huang2017densely}, usually have state-of-the-art performance at extracting visual features. In terms of face recognition, pre-trained model, like VGGFace network \citep{VGGFace2Cao18}, is also confirmed to have great ability in extracting facial features as said in \cite{kollias2} and \cite{Chen2017MultimodalRecognition}. In \cite{kollias2}, a joint CNN and RNN architecture is trained as a whole with pre-trained model and fine-tuning technique and achieved the best performance compared with other candidates' results on Aff-Wild Challenge. So this CNN joint RNN architecture is adopted to design the deep network in this project. The \textbf{second objective} for this project is implementing an end-to-end deep neural network having joint CNN and RNN architecture to estimate the valence and arousal value for the facial expression in each frame of videos, which belong to the database created by fulfilling the first objective. Plenty of experiments on different combinations of CNNs and RNNs are executed to find the best performing combination. 

\section{Contribution}
\label{sec:contribution}
\begin{enumerate}
    \item The \textbf{first contribution} of this project is the completion of the Aff-Wild database extension. The newly created database has 100, 29 and 30 videos sourced from YouTube in train, validation and test data set, having 527056, 94223 and 135145 frames respectively. All frames contain spontaneous facial expression having the annotation of continuous valence and arousal value ranging in [-1000, 1000] which is scaled to [-1.0, 1.0] later. The value of valence is more balanced regarding positive and negative distribution than arousal, and arousal mainly contains positive values. The created database also ensures diversity in the age and ethnicity of the subjects in the videos. The male and female ratio is close to 1:1. The videos all have the MP4 format and 30 FPS. The length of the videos is in the range of 0.10 to 15.04 minutes. The procedure of developing this database consists of collecting, annotating, pre-processing and partitioning. One dominant challenge is to label the frames in videos as accurately as possible. Because the facial expressions in the video are really difficult to distinguish and the expression changes very quickly in most cases. Overcoming the difficulties here mainly relies on studying video subtitles and manually modifying the annotation files. Another challenge is that in addition to the face, other objects are detected by the detector and cropped. This makes it difficult to get a standard database. The method used here to solve this problem is to calculate the features of all detected objects of one frame, such as RGB histogram, and then compare them with the face feature that really wants to be cropped. For each frame, only the object which has the most similar feature with the real face will be kept. 
    
    \item The \textbf{second contribution} of this project is an end-to-end deep learning neural network is built to have the best architectures found through experiments and its weights which can be used to accept a sequence of frames as input and give the output of VA values for each frame. To be concrete, batches of sequential frames ${F \in {\rm I\!R}^{Sequence Length \times 96\times96\times3}}$ will be first fed into the VGGFace \citep{Parkhi2015DeepRecognition} network so that the visual features can be extracted, then the sequential output features from CNN block will enter the 2 layers GRU block with attention mechanism so that the temporal dynamics will be captured. Finally, the output from RNN block will be transformed into the prediction ${P \in [-1, 1]^{Sequence Length \times 2}}$ by a fully connected layer which has two neurons. The CNN block is initialized with the VGGFace pre-trained model and RNN block is initialized with the model pre-trained on the created database in this project. The sequence length is 80 and the number of hidden units of GRU is 128. Attention mechanism mainly focuses on the last 30 frames outputs to predict current frame output. This designed architecture is trained as a whole by applying Adam optimizer with a learning rate of 0.0001. The final test performance with respect to CCC \citep{lawrence1989concordance} value is 0.555 for valence and 0.499 for arousal. The experiments for finding the best combination of CNN and RNN blocks are detailed in Chapter \ref{experimentChapter}. The main challenge during experiments stage is the training time is very long which leads to difficulty in adjusting the hyper-parameters or making a modification to the training program in a short time. To solve this, training and evaluating program are running on two separate GPUs which can speed up the training process a little bit.

\end{enumerate}

\section{Report Arrangements}
\label{sec:arrangement}
The report arrangements are shown as follows:
\begin{enumerate}
    \item Chapter \ref{backgroundChapter} first gives an overview of all potential emotion measurements, their related databases, relevant challenges and corresponding winning methods. Second, deep neural architectures including CNNs and RNNs which are very likely to be experimented with later in the project are demonstrated. A brief summary can be found in Section \ref{backgroundSummary}.
    
    \item Chapter \ref{databaseChapter} shows the whole workflow of how the database was created. The videos were collected from YouTube and annotated first. Then FFld2 detector was used to detect the faces in every frame. Finally, the database was partitioned into the train, validate and test set. A brief summary can be found in Section \ref{databasesummary}.
    
    \item Chapter \ref{designChapter} specifies the details of how the whole deep neural network is designed. The concrete configuration of alternative CNN blocks and RNN blocks which will be used in later experiments are also demonstrated. A brief summary can be found in Section \ref{designsummary}.
    
    \item Chapter \ref{implementationChapter} illustrates the implementation of the whole training, evaluating and testing procedure, including how the data set is processed and fed into the network, how the various CNN and RNN blocks are constructed and how the train, evaluate and test program are set up to execute experiments. A brief summary can be found in Section \ref{implementationsummary}.
    
    \item Chapter \ref{experimentChapter} tells how the experiments on the candidate CNN and RNN blocks are executed and provides evaluations on the experiments result. In the end, the determined best performance architecture is explained. A brief summary can be found in Section \ref{experimentsummary}.
    
    \item Chapter \ref{conclusionChapter} reports the achievements and challenges of this project. The evaluation of the achievements and future work are also pointed out.
    
\end{enumerate}

\chapter{Background}
\label{backgroundChapter}
This chapter aims to fully study the emotion recognition related work. In this background chapter, to start with, emotion recognition relevant measurements will be introduced, including Seven Basic Emotions, Action Units in section \ref{otherMeasurements} and Valence and Arousal in section \ref{VAmeasurements}, together with corresponding associated databases and challenges. In addition, the state-of-the-art winning methods for these challenges are discussed. To address the affect analysis problem later on, the advanced CNNs and RNNs are also investigated in section \ref{architecturesSection}. At the end, a evaluation of these related work is summarized in section \ref{architecturesSection}.

\section{Other Emotion Recognition Measurements}
\label{otherMeasurements}
\subsection{Seven Basic Emotion}
\label{sec:sevenbasicemotion}
Emotions can be judged in a categorical way. Six basic emotions consisting of anger, disgust, fear, happiness, sadness and surprise are universally displayed and accepted in the same way \citep{ekman1982emotions}. Together with neutral expression, there are seven basic emotions as shown in Figure \ref{SBE}:

\begin{figure}[H]
\centering
\includegraphics[height=5cm,width=13cm]{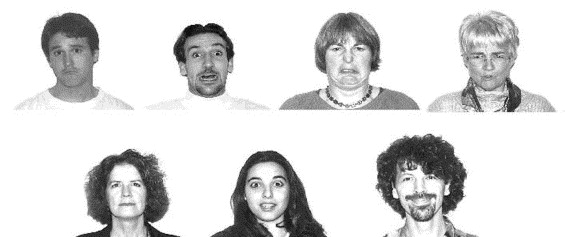}
\caption{Representatives of Seven Basic Emotions. From top to down and left to right, the emotion is of sadness, fear, disgust, anger, neutral, surprise and happiness in turn. \citep{batty2003early}}
\label{SBE}
\end{figure}

\subsection{Basic Emotion Related Databases}
Acted Facial Expressions in the Wild (AFEW) is an acted facial expressions dataset in challenging conditions\citep{dhall2012collecting}. The video data in AFEW are movies collected from WWW and annotated with six basic expressions anger, disgust, fear, happiness, sadness, surprise and the neutral class. Although the movies are in controlled conditions to some extent, AFEW is still more close to real world environments compared to other datasets collected in lab conditions.

\subsection{EmotiW Challenge}
One categorical emotion recognition related challenge is the Emotion Recognition in the Wild (EmotiW) challenge \footnote{https://sites.google.com/site/emotiwchallenge/}. EmotiW aims to provide a common benchmark for researchers analyzing affect and evaluating various approaches in `in the wild' conditions. Audio-Video Sub-challenge is included in The fifth EmotiW challenge in 2017 \citep{dhall2017individual}.

\subsubsection{Audio-Video Sub-challenge}
The Audio-Video Sub-challenge's assignment is to classify a piece of audio-video clip into one of the seven basic categories mentioned in \ref{sec:sevenbasicemotion}. The data used in this sub-challenge was AFEW \citep{dhall2012collecting} which was created by analyzing sentiment of subtitles in films and TV series directly. The AFEW data was divided into three data partitions: Train, Validate and Test, having 773, 383 and 653 samples respectively. For test data, sitcom TV series data was also added. And it is worth noting that the data in the three sets are independent from each other \citep{kollias6} in terms of the source of movies/TV and actors. The evaluation metric is unweighted classification accuracy. The video only baseline system reaches 38.81\% and 41.07\% classification accuracy for the Validate and Test sets. One of Audio-Video sub-challenge winners presents a new learning method named Supervised Scoring Ensemble (SSE) \citep{hu2017learning} which enhanced deep CNNs for improving this challenge. They first developed the idea of recent deep supervision to solve this sub-challenge emotion recognition problem. In more details, they applied supervision mechanism not only on the output layer but also on middle layers and shallow layers. In this way, the training of neural networks could become more sufficient. In addition, they propose an advanced fusion structure in which class-wise activating scores at various complementary feature layers are concatenated together and the connected results later used as the incoming value for secondary supervision, functioning as a deep feature ensemble within a single CNN network\citep{hu2017learning}. In this 2017 audio-video emotion recognition task, the average recognition classification accuracy rate of their best submission is 60.34\%.

\subsection{Action Units}
\label{FACS}
\begin{figure}[H]
\centering
\includegraphics[height=10cm,width=10cm]{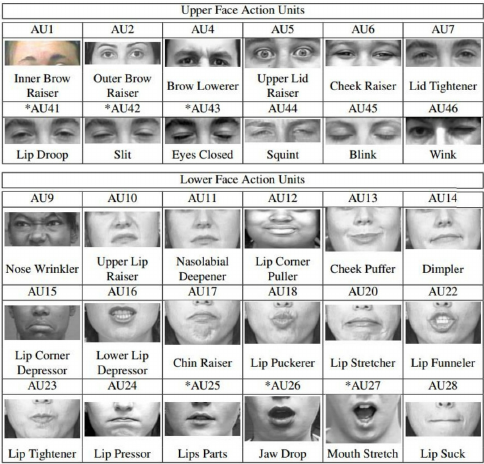}
\caption{The set of action units needed for basic emotions}
\label{AUs}
\end{figure}

Emotions can be measured by facial movements. Facial Action Coding System (FACS) established by Ekman and Friesen \citep{friesen1978facial} illustrates all possible facial muscle activations that lead to a visible change in the facial appearance. In this way, the human facial expressions can be well described. Action Units, as the basic measurement units used in FACS, represent one muscles or a group of muscles. In figure \ref{AUs}, it shows examples of Action Units.


\subsection{Action Units Related Databases}
\subsubsection{BP4D-Spontaneous Database} 
\label{BP4D}
BP4D-Spontaneous database \citep{zhang2014bp4d} is a high-quality spontaneous 3D dynamic facial expression database, whereas most available databases contain 2D static facial images or have posed facial behaviours. The video data consists of young adults give reactions to eight emotion-elicitation tasks for generating target expressions and behaviours which designed by professional director, captured by Di3D dynamic face capturing system in lab conditions. The annotations for every frame regarding facial actions was obtained using the FACS. 41 participants (including 23 women and 18 men) were recruited from the departments of psychology and computer science as well as from the school of engineering. They age 18–29 years, with 11 were Asian, 6 were African-American, 4 were Hispanic, and 20 were Euro-American.

\subsubsection{SEMAINE Database}
\label{SEMAINE}
Sustained Emotionally colored Machine-human Interaction using Nonverbal Expression(SEMAINE) database\footnote{SEMAINE is available form http://semaine-db.eu} \citep{mckeown2012semaine} is an annotated audio-video database consists of the recorded conversations between participants and operator simulating a Sensitive Artificial Listener(SAL) agent whose emotional state is rigid under different configurations. Resulting database has 959 conversations recorded by 150 users, with about 5 minutes each. The frames per second is 49.979 for video, having resolution of 780 x 580 pixels, while audio was recorded at 48 kHz. FACS annotation, mentioned in \ref{FACS}, was also given to eight character interactions. Selected frames received labels for Action Units appearance and whether they are combined with other Action Units. Results are 577 Action Unit codings in 181 frames.

\subsection{FERA Challenges}
Facial Expression Recognition and Analysis challenge (FERA) 2015 \citep{valstar2015fera} has three sub-challenges: the detection of AU occurrence, the estimation of AU intensity for pre-segmented data, and fully automatic AU intensity estimation. The training, development and test data for the FERA 2015 challenge are sourced from two databases: the BP4D-Spontaneous database \citep{zhang2014bp4d} and the SEMAINE database \citep{mckeown2012semaine}. The training and development sets drawn from SEMAINE database have 16 sessions and 15 sessions respectively. Deriving from BP4D-Spontaneous, the training and development have 21 subjects and 20 subjects respectively. The test set is drawn from part of the SEMAINE database which includes 12 sessions and an extended version of BP4D which includes videos capturing 20 participants reactions following similar procedure as stated in \ref{BP4D}. The entire data set is split into train and test parts. Only the train set is publicly available. The test set is held back by the challenge organizers. Participants provide their well-trained models and the FERA 2015 organizers test their submitted models on this test data to create a fair comparison on the model's performance. 

\subsubsection{Occurrence Sub-challenge}
\begin{figure}[H]
\centering
\includegraphics[height=3cm,width=10cm]{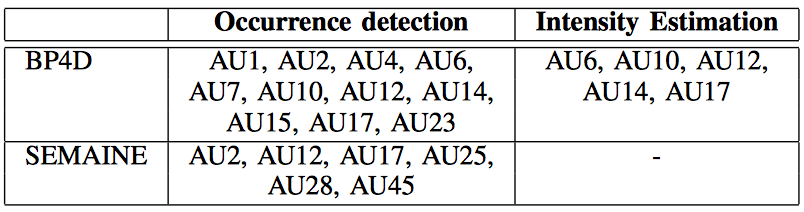}
\caption{OVERVIEW OF AUS INCLUDED IN THE THREE SUB-CHALLENGES
\citep{valstar2015fera}}
\label{AUDB}
\end{figure}

The Occurrence Detection sub-challenge requires candidates to detect 11 AUs from the BP4D database and 6 from the SEMAINE database (see Figure \ref{AUDB}). The performance measure used here to judge participants for AU occurrence is the F1-measure. Final scores are computed on the results of the two databases as a weighted mean on the total number of samples in each database. The results of baseline system had been obtained using linear SVM with geometric feature which derived from tracked facial point locations and appearance feature which extracted by adopting local LGBP\citep{zhang2005local} descriptor. For baseline results, the weighted mean value of F1 score for detection performance is 0.444 using geometric feature and 0.400 using appearance feature on the test partition. The winning model \citep{jaiswal2016deep} for occurrence sub-challenge propose a innovative method to Facial Action Unit detection using a deep learning architecture of joint CNN and BiLSTM, which learns visual cues as well as temporal dynamics. Moreover, they invent a new method to model shape features by utilizing binary image masks computed from the facial landmarks locations. The weighted average performance on BP4D and SEMAINE for AU occurrence achieved by this winning model is 0.5478. 

\section{Emotion Recognition By VA}
\label{VAmeasurements}
\subsection{Valence and Arousal}
\label{VAdefinition}
\begin{figure}[H]
\centering
\includegraphics[height=10cm,width=12cm]{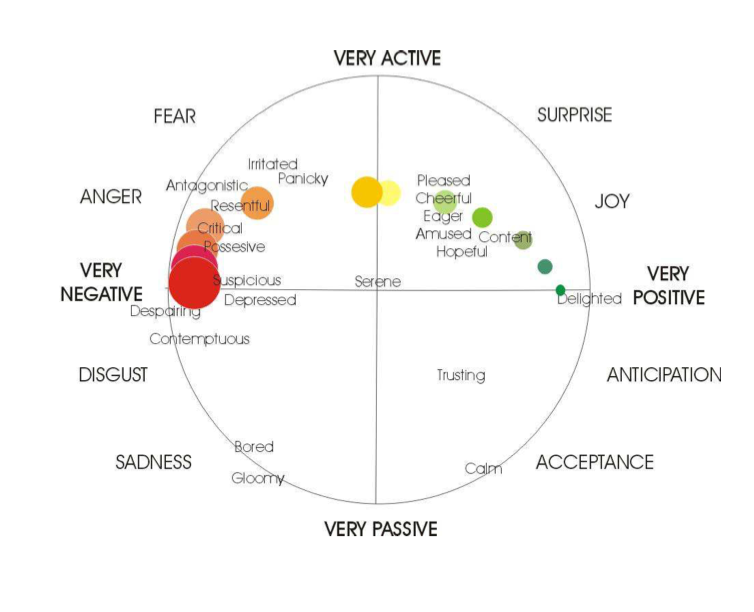}
\caption{The 2-D Emotion Wheel\citep{plutchik1980emotion}}
\label{EW}
\end{figure}

Emotion states can be measured in two dimensions : valence and arousal \citep{russell1980circumplex}. As stated in \cite{kollias5}, the valence measures how positive or negative an emotion is and the arousal evaluates the power of the activation of the emotion. And it is supported that these two dimensions are highly correlated \citep{russell1978evidence}. In Figure \ref{EW}, it shows the emotion in terms of the valence and arousal dimensions. We can see in this dimensional way, subtle emotions and expansive scope of emotions can be captured \citep{kollias8,kollias9}.

\subsection{Related Database}
\subsubsection{SEMAINE Database}
\label{SEMAINEVA}
Sustained Emotionally colored Machine-human Interaction using Nonverbal Expression(SEMAINE) database\footnote{SEMAINE is available form http://semaine-db.eu} \citep{mckeown2012semaine} is an annotated audio-video database consists of the recorded conversations between participants and operator simulating a Sensitive Artificial Listener(SAL) agent whose emotional state is rigid under different configurations as stated in \ref{SEMAINE}. Especially, in Solid SAL scenario, where a human act as the character of a SAL agent, consists of 21 sessions, including 75 character interactions, that were labelled with FEELTRACE for valence and arousal\citep{Kossaifi2017AFEW-VAIn-the-wild}.


\subsubsection{RECOLA Database}
The RECOLA dataset\citep{Ringeval2013IntroducingInteractions} contains multimodal data, which includes audio and video, of spontaneous collaborative and emotinal interactions in French produced by recording 46 participants(but only 27 of them agree to share their data)  collaborate in pairs during a video conference under well-controlled conditions. The annotations are provided by 6 annotators for first 5 minutes of each video on two dimensions: arousal and valence and are ranged from -1 to 1 with a step of 0.01. 

\subsubsection{AFEW-VA Database}
AFEW-VA\citep{Kossaifi2017AFEW-VAIn-the-wild} contains 600 video clips chosen from films which are mostly recorded in challenging conditions. Although the performance of actors are somewhat controlled, the actors live in their roles. In this case, the facial expressions of actors can be regarded as spontaneous emotions. And the actors whose movies are included in this AFEW-VA data set have various age and ethnicity. Valence and arousal annotations are provided in frame by frame approach which is highly accurate and ranged from -10 to 10, totally 21 levels. The number of frames of videos range from 10 to 120, which is relatively short to analyze the dynamics in clips.

\subsubsection{SEWA Database}
\label{SEWA}
Sentiment Analysis in the Wild (SEWA) database\footnote{http://sewaproject.eu} contains audiovisual data recording human-human interactions showing spontaneous emotions `in the wild', namely the data is collected at home or office with standard webcams and microphones from the computers belonging to subjects. The conversation of a pair of people is about one commercial advertisement, having maximum duration of 3 minutes.

\subsection{AVEC Challenge}
The Audio/Visual Emotion Challenge(AVEC) in 2017 \cite{ringeval2017avec} aims to provide a common benchmark for evaluating different approaches solving depression and emotion recognition problem using multi-modal information in the wild. In Affect Sub-Challenge (ASC) \cite{ringeval2017avec}, participants are expected to provide continuous emotion prediction of three emotional dimensions: Arousal, Valence, and Likability, which represents how much the participants like the commercial product. The data set used for ASC is the subset of SEWA database mentioned in \ref{SEWA}, only contains recordings of 64 German subjects whose age are ranged from 18 to 60. This subset is partitioned into training having 36 subjects, development having 14 subjects and testing having 16 subjects. The three dimensions, which are valence, arousal and liking, for this subset are time-continuous values in range from -1 to 1 annotated by 6 annotators who are all German native speakers annotated using a joystick continuously. In addition, three features consisting of video, audio and text are provided for participants to use freely. The evaluation function for ASC is CCC which is defined as in Formulation \ref{CCCPearson}:

\begin{equation}
\begin{aligned}
&\rho_{c}=\frac{2 \rho s_{x}s_{y}}{s_{x}^2 + s_{y}^2 + (\Bar{x}-\Bar{y})^2}
\end{aligned}
\label{CCCPearson}
\end{equation}

where ${\rho}$ is the Pearson correlation coefficient between ground truth and predictions. ${s_{x}^2}$ and ${\Bar{x}}$ are the variances and mean value of the predicted values, ${s_{y}^2}$ and ${\Bar{y}}$ are the variances and mean value of the ground truth values. With this CCC evaluation metric, the baseline performance for ASC is 0.375 for arousal, 0.466 for valence and 0.246 for liking. The winning method for ASC is published in \citep{Chen2017MultimodalRecognition}. Their approach outperform other solutions in three aspects. First, they combine the learned features of deep learning model and features obtained by applying feature engineering from all three modalities consisting of audio, video and text. Second, they compare the temporal model LSTM-RNN and non-temporal model and come to a conclusion that LSTM-RNN is good at modelling time dependent features and achieves better recognition performance than non-temporal model. Third, since the different dimensions are highly correlated with each other, they apply multi-task learning approach to predict multiple emotion dimensions with common representations. Their solution give final result of 0.675, 0.756 and 0.509 on arousal, valence, and likability in terms of CCC metric.


\subsection{Aff-Wild Challenge}
\label{AffWildChallenge}
The Affect-in-the-Wild (Aff-Wild) Challenge \citep{kollias1} proposes a
new comprehensive benchmark for assessing the performance of facial affect analysis `in-the-wild' which represents the scenario in which facial expressions analyzed are spontaneous and under un-controlled conditions. The Aff-Wild benchmark contains 298 videos (more than 30 hours) annotated by 6-8 lay experts with dimensional values of valence and arousal ranged continuously from −1 to +1. The main source for these videos is YouTube and the videos are searched using the keyword "reaction" so that the collected facial expressions are in arbitrary recording conditions and naturalistic. In total, Aff-Wild database has 200 subjects, with 130 males and 70 females. The measurement used to judge the participants' performance are CCC \citep{lawrence1989concordance} and Mean Squared Error(MSE). In particular, CCC is defined as:

\begin{equation}
\begin{aligned}
&\rho_{c}=\frac{2s_{x_y}}{s_{x}^2 + s_{y}^2 + (\Bar{x}-\Bar{y})^2}
\end{aligned}
\label{CCC} 
\end{equation}

where ${s_{x}^2}$ and ${\Bar{x}}$ are the variances and mean value of the predicted values, ${s_{y}^2}$ and ${\Bar{y}}$ are the variances and mean value of the ground truth values. And ${s_{x_y}}$ is the covariance value of predicted and ground truth value. 

The CNN-M \citep{conf/bmvc/ChatfieldSVZ14} architecture is used as baseline architecture with two fully connected layers having 4096 and 2 neurons respectively on top of it. Pre-trained weights on the FaceValue dataset \citep{albanie2016learning} are used for the CNN part and Truncated Normal distribution with 0 mean and 0.1 variance is used for fully connected layers. The network was trained either in a way of fixing the CNN part and only fine-tuning the fc layers or in an approach of fine-tuning the whole network. For training the network, Adam optimizer is used to minimize the cost function defined using MSE value. The hyper-parameters were: batch size of 80 and 0.001 fixed learning rate.

\begin{figure}[H]
\centering
\includegraphics[height=1.5cm,width=7cm]{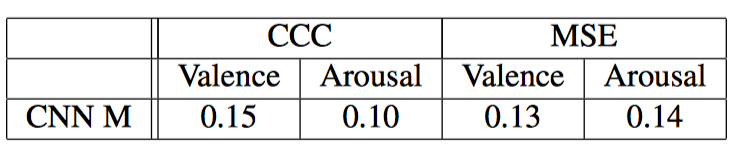}
\caption{Concordance (CCC) and Mean Squared Error (MSE) evaluation of valence and arousal predictions provided by the CNN-M baseline architecture \citep{kollias1}. }
\label{bl} 
\end{figure}

The resulting baseline value of CCC and MSE evaluation of valence and arousal predictions are shown in figure \ref{bl}.

Among the participants, the winning method was FATAUVA-Net \citep{chang2017fatauva} which proposed an integrated deep learning framework for facial attribute recognition, AUs detection and V-A estimation. The crucial design was to utilize AUs as mid-level representation to predict the valence and arousal values. In addition, the AU detector was trained based on the CNN part for facial attribute recognition. The final performance of FATAUVA-Net was 0.396 of valence and 0.282 of arousal measured by CCC and 0.123 of valence and 0.095 of arousal measured by MSE. 

Aff-Wild Challenge organizers developed a method VA-CRNN \citep{kollias2,kollias3} during the challenge as well and achieved better results (which have 0.57 CCC for valence and 0.43 CCC for arousal) than the best result achieved by the participants. In the data pre-processing procedure, the cropped face for each frame was detected by the method in \citep{mathias2014face} and 68 facial landmarks also tracked using method in \citep{chrysos2018comprehensive} for corresponding frame. The VA-CRNN method was based on an end-to-end architecture consists of CNN and RNN where CNN part can extract invariant features for each frame and RNN part can model temporal information for the sequence of frames. The details of the CRNN architecture is shown in figure \ref{crnn}:

\begin{figure}[H]
\centering
\includegraphics[height=3cm, width=6cm]{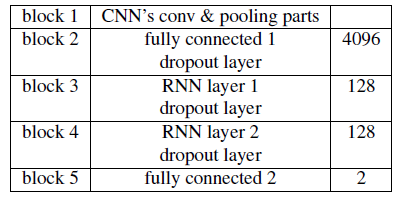}
\caption{Architecture for CNN-RNN network.}
\label{crnn}
\end{figure}

where VGG-Face \citep{parkhi2015deep} network with pre-trained weights on Face Value \citep{albanie2016learning} dataset was used for CNN part and GRU cell having 128 hidden units was used in RNN part. Facial landmarks were fed into the fully connected 1 layer, together with the output from the last pooling layer of the CNN part. The training was performed in the way of freezing the CNN part and retraining the rest network. Adam optimizer was used to minimize the cost function defined using CCC value defined in \ref{CCC}. Other hyper-parameters are batch size of 100, dropout probability value of 0.5 and learning rate of 0.001.

\section{Architectures}
\label{architecturesSection}
\subsection{CNN}
\begin{figure}[H]
\centering
\includegraphics[height=4cm,width=16cm]{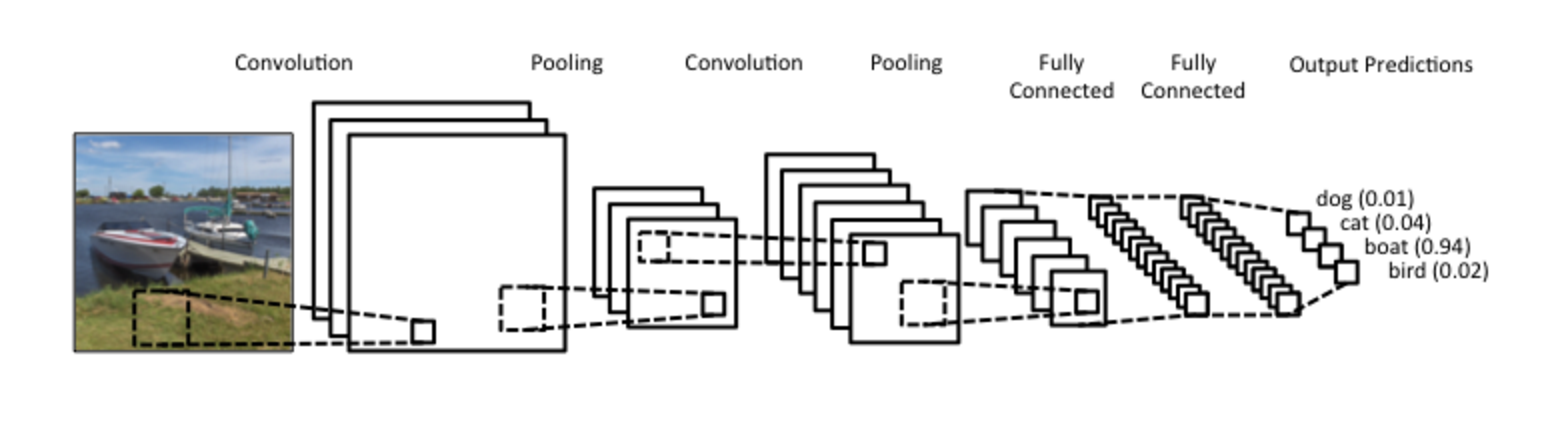}
\caption{This image is obtained from https://www.clarifai.com/technology. There are two structures of convolutional layer followed by a pooling layer, with two fully connected layers at the end before output layer of four outputs.}
\label{cnn}
\end{figure}

Convolutional Neural Networks (CNNs) usually consists of a sequence of layers including input layer, convolutional layer, activation layer, pooling layer and fully connected layer. Each layer of CNNs transforms one amount of activated results from previous layer to another by passing through a differentiable function. Figure \ref{cnn} shows an example of a CNN architecture. 

\subsubsection{Input Layer}
The input for CNN can be raw pixel values of the image (\cite{simou2008image}) which has width $W$ and height $H$ as spatial dimension parameter and channel number $D$ as depth parameter. If we take RGB image as input then the channel value is 3. If we take gray scale image, then the channel value is 1. Basicly, the dimension of the input can be describe as $ W \times H \times D$. 

\subsubsection{Convolutional Layer}
Convolutional Layer is the key design of CNN. Convolutional layer has a set of learnable filters. Instead of connecting every neuron in the previous hidden layer, each neuron $z$ in convolutional layer is only connected to a local region of previous layer neurons' activation $a$ by using filters. The spatial size of this connectivity is the filter size $F$ and the depth of this connectivity is the input depth $D_{1}$. The dimension of weights $w$ used in each filter is $F \times F \times D_{1}$. Each neuron $z$ in conv layer is computed by performing dot product between the set of weights of filter and corresponding local region in its input, adding bias $b$:

\begin{equation}
\label{equForline}
\begin{aligned}
&z^{l}=\sum_{d=0}^{D_{1}-1}\sum_{x,y=0}^{F-1}w_{d,x,y}a_{d,x,y}^{l-1} + b\\
\end{aligned}
\end{equation}

All neurons in the same depth slice in conv layer uses the same weights and bias. One depth slice is computed by sliding the filter spatially with stride $S$ pixels. So a conv layer whose depth is $D_{2}$ has the weights of dimension $D_{2}\times F \times F \times D_{1}$ and $D_{2}$ biases. To control the output size, we can pad zeros which number is $P$ to preceding layer's activation around its border, using $P_{1}$ representing number of zeros padded along width dimension and $P_{2}$ for height dimension. So if we have input volume with dimensions $W_{1} \times H_{1} \times D_{1}$ and output volume having dimensions W2 × H2 × D2. We can get following relation between them :

\begin{equation*}
\label{equForline}
\begin{aligned}
&W_{2} = \frac{W_{1} − F + 2P_{1}}{S} + 1,\\
&H_{2} = \frac{H_{1} − F + 2P_{2}}{S} + 1
\end{aligned}
\end{equation*}

\subsubsection{Activation Layer}
A rectified linear unit (ReLU) function is commonly used as activation function in deep neural networks. Relu perform element-wise operation for each neuron $z$ in a hidden layer ${l}$:

\begin{equation*}
\label{equForline}
\begin{aligned}
&a^{l} = max(0, z^{l}) \\
\end{aligned}
\end{equation*}

After applying Relu, the dimension of the input remains unchanged. Other activation functions, like sigmoid function, can also be applied. But ReLU is especially suitable for very deep neural network for addressing gradient vanishing problem.

\subsubsection{Pooling Layer}
It is common to periodically insert a pooling layer between consecutive conv layers in a CNN architecture performing down-sampling operation for each depth slice of input volume, leaving depth of input volume unchanged. There are many pooling functions. A popular one of them is max pooling. Accepting input volume having dimension $W_{1} \times H_{1} \times D_{1}$, max pooling layer perform a $MAX$ function with filter having dimension $F \times F$ sliding by stride $S$ spatially for every depth slice of the input volume independently. In detials, every sub region of size $F \times F$ in each depth slice in input volume, will be transformed to a single value, which is the maximum value in this sub region. Resulting output volume having dimension $W_{2} \times H_{2} \times D_{2}$ having following relations:

\begin{equation*}
\label{equForline}
\begin{aligned}
&W_{2} = \frac{W_{1} − F}{S} + 1,\\
&H_{2} = \frac{H_{1} − F}{S} + 1,\\
&D_{2} = D_{1}
\end{aligned}
\end{equation*}

\subsubsection{Fully Connected Layer}
Several fully connected layers usually function as the final part of CNNs. Here for illustrative convenience, a figure indicating fully connected layer is attached here. 
\begin{figure}[H]
\centering
\caption{Figure of Fully Connected Layer}
\includegraphics[width=5cm,height=5cm]{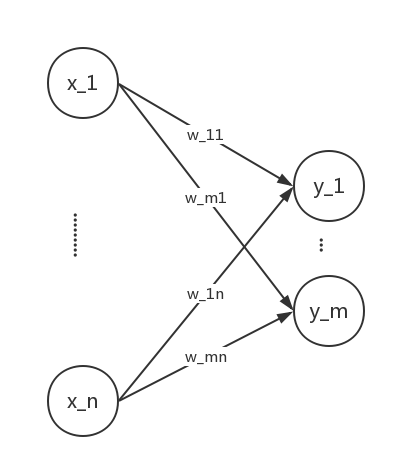}
\end{figure}
Each neuron in a fully connected layer have full connections to all activation in preceding layer. Let's say we have $n$ input units and $m$ output units. Also for each output units in fully connected layer, there is a bias term $b_j$. So the formulation for producing output activation in fully connected layer is:

\begin{equation}
\label{equForline}
\begin{aligned}
&y_j=\sum_{i=1}^{n}w_{ji}x_{i} + b_j, j \in [1,m]\\
\end{aligned}
\end{equation}

where $y_j$ is each unit in the output layer. So for all neurons in fully connected layer, the weights needed have dimension of $m \times n$. The output of the last fully connected layer in CNN function as the the output of the entire CNN.

\subsection{VGG16}
\label{VGG16section}
VGGNet \citep{simonyan2014very} was designed to have deeper architecture, more non-linearities and fewer parameters by using stack of small convolutional filters. Receptive field of size $3 \times 3$ is used for convolutional filters in VGGNet.

\begin{figure}[H]
\centering
\includegraphics[width=16cm,height=6cm]{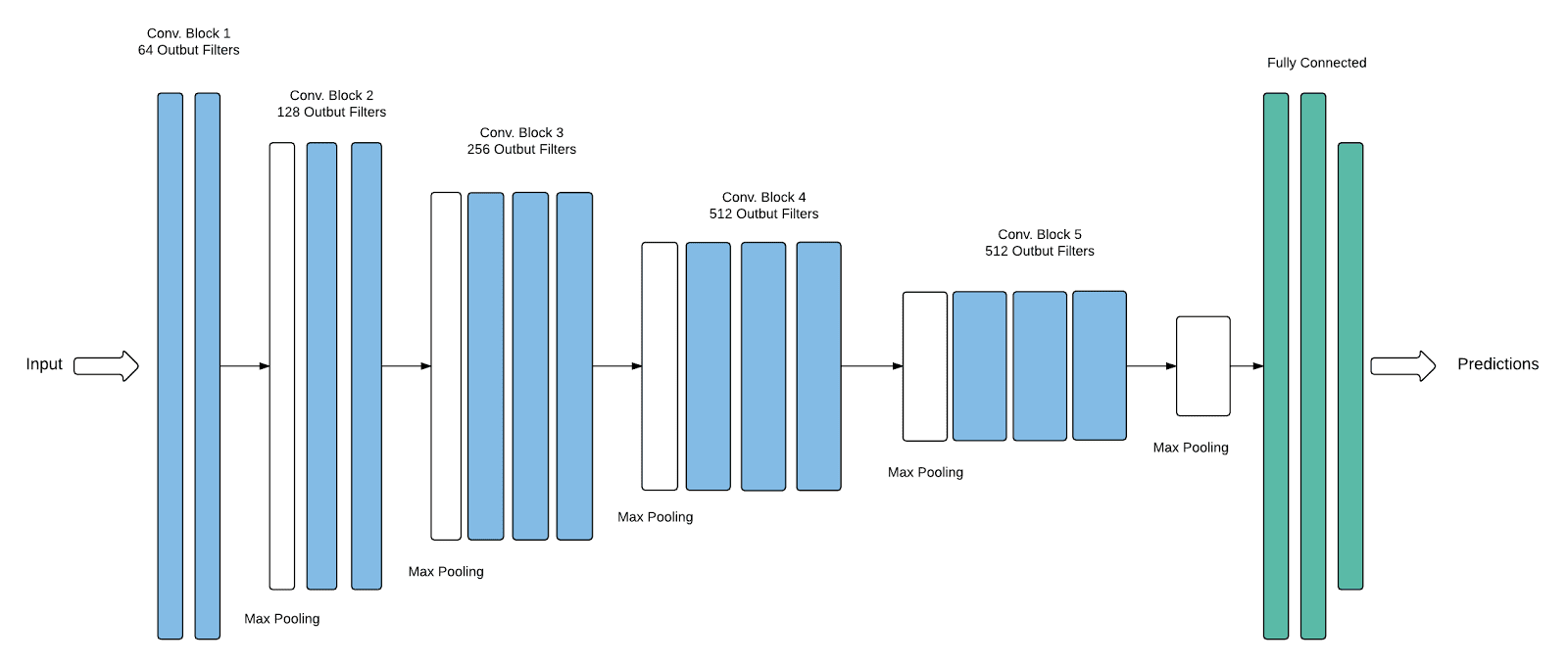}
\caption{ VGGNet16 Architecture obtained from \citep{CodesDeeper}}
\label{vgg}
\end{figure}

The architecture of VGG16 which has 16 weight layers, containing 13 convolutional layers and 3 fully connected layers, is shown in figure \ref{vgg}. Described in \citep{simonyan2014very}, the input image size is $224 \times 224$, which then is passed through a few convolutional (conv) layers, whose filters have $3 \times 3$ kernel size which is the smallest size for extracting spatial features. In which design, more non-linearity is introduced with fewer parameters. The stride size is fixed to 1 pixel and the padding size is 1 pixel for ${3 \times 3}$ conv layers so that the input spatial dimension remains the same after transformation. Max pooling layers are adopted interleaving in conv blocks which consists of two or three conv layers and operated over a $2 \times 2$ pixel region, having stride size of 2. Finally, three fully connected (fc) layers are attached on top of conv net. The first two fc layers have 4096 neurons each and the third fc layer performs 1000 categories ILSVRC\footnote{http://www.image-net.org/challenges/LSVRC/} classification and thus contains 1000 neurons. The final layer is the softmax layer. All hidden layers are followed by ReLU activation layer.


\subsection{ResNet50}
\label{ResNet50section}
As far as we know, deeper networks, like VGG16 and GoogleNet, usually have better performance in ImageNet chellenge. But as the depth goes further, the network can not be trained sufficiently due to gradient vanishing problem. ResNet \citep{he2016deep} has extremely deeper architecture which has up to 152 layers tried for ImageNet classification. This is mainly implemented by a residual connection design so that it can have deeper network but with lower risk of having gradient vanishing problem.

\begin{figure}[H]
\centering
\includegraphics[width=7cm,height=3cm]{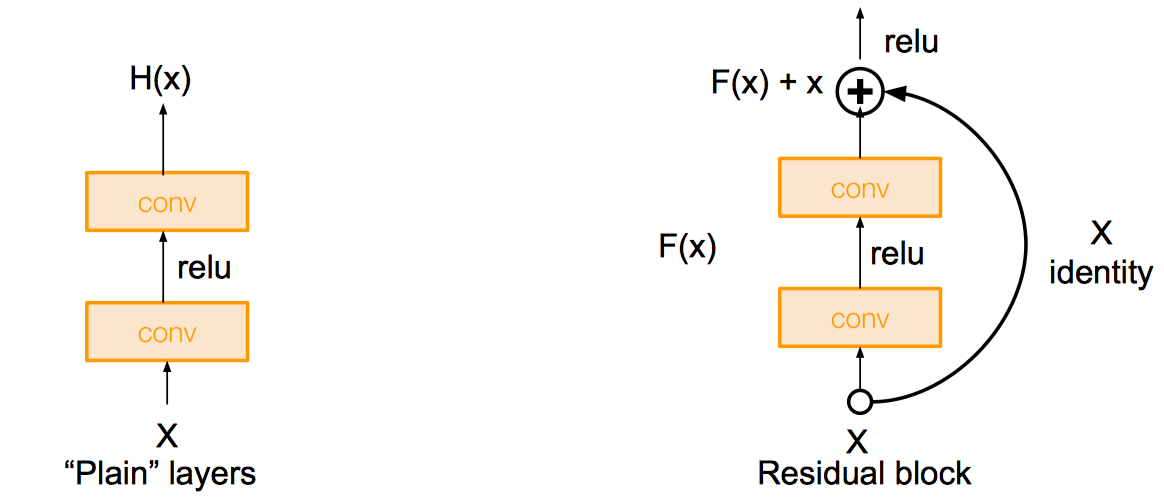}
\caption{ Residual connection from http://cs231n.stanford.edu.}
\label{res}
\end{figure}

As Figure \ref{res} shown, instead of directly trying to fit a desired underlying mapping $H(x)$, ResNet is designed to fit a residual mapping $F(x)$ where:

\begin{equation}
\label{equForline}
\begin{aligned}
&F(x)= H(x) - x
\end{aligned}
\end{equation}

However, sometimes the shortcuts need to transform the input to have the same dimension with the residual mapping, in which case the following formulation is used as residual connection:

\begin{equation}
\label{equForline}
\begin{aligned}
&H(x) = F(x, {W_{i}}) + W_{s}x
\end{aligned}
\end{equation}

where the ${W_{s}}$ is the linear projection matrix used by shortcuts connection.

\begin{figure}[H]
\centering
\includegraphics[width=7cm,height=3cm]{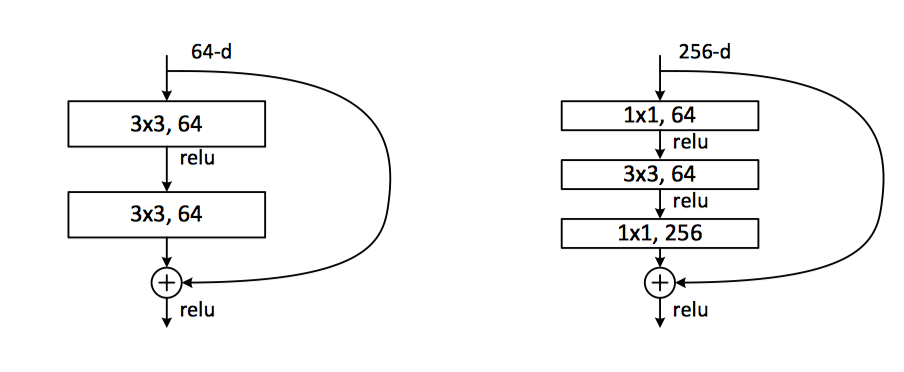}
\caption{ Left is a residual block has two conv layers. Right is Bottleneck architecture. \citep{he2016deep} }
\label{bottleneck}
\end{figure}

In order to control the training time, a deeper bottleneck architecture is designed to have three layers, which consists of ${1 \times 1}$, ${3 \times 3}$ and ${1 \times 1}$ conv layers so that the two ${1 \times 1}$ conv layer can first reducing then increasing the dimension of input and output. In this case, the ${3 \times 3}$ conv layer can have less computation cost. This bottleneck architectures shown in Figure \ref{bottleneck} are used to form ResNet50. 

After receiving the input, the first layer of ResNet50 is a convolutional layer having 64 filters with size ${7 \times 7}$ and stride 2, followed by a ${3 \times 3}$ max pooling layer with stride 2. Then the preceding outcome will be fed into the stack of replicated bottleneck architectures whose details are described in Table \ref{tab:resnet50}. After the bottleneck structures, there is one average pooling layer before final fully connected layer having 1000 dimensions followed by soft-max to output predictions. For ResNet50, projection shortcuts are applied to increase dimensions, and other shortcuts are identity mapping.

\begin{table}[H]
    \centering
    \begin{tabular}{|c|c|}
	\hline
	layer name & ResNet50  \\
	\hline
	conv1
	&
    ${
    7 \times 7, 64, stride 2
    }$ \\
    \hline
	conv2\_x
	&
    ${3 \times 3}$ max pool , stride 2
    \\
	&
    ${
    \left[\begin{array}
    {cc} 1 \times 1, & 64 \\  3 \times 3, & 64 \\ 1 \times 1, & 256
    \end{array}\right] \times 3
    }$\\
    \hline
    conv3\_x
	&
    ${
    \left[\begin{array}
    {cc} 1 \times 1, & 128 \\  3 \times 3, & 128 \\ 1 \times 1, & 512
    \end{array}\right] \times 4
    }$\\
    \hline
    conv4\_x
	&
    ${
    \left[\begin{array}
    {cc} 1 \times 1, & 256 \\  3 \times 3, & 256 \\ 1 \times 1, & 1024
    \end{array}\right] \times 6
    }$\\
    \hline
    conv5\_x
	&
    ${
    \left[\begin{array}
    {cc} 1 \times 1, & 512 \\  3 \times 3, & 512 \\ 1 \times 1, & 2048
    \end{array}\right] \times 3
    }$\\
    \hline
	&
    average pool, 1000-d fc, softmax
    \\
    \hline
    \end{tabular}
    \caption{ResNet50 Architecture. ${n \times n}$ represents the conv filter size, after which is the number of filters. The number after array structure means how many times this structure will be replicated. }
    \label{tab:resnet50}
\end{table}

\subsection{DenseNet}
\label{DenseNetsection}
\begin{figure}[H]
\centering
\includegraphics[width=11cm,height=6cm]{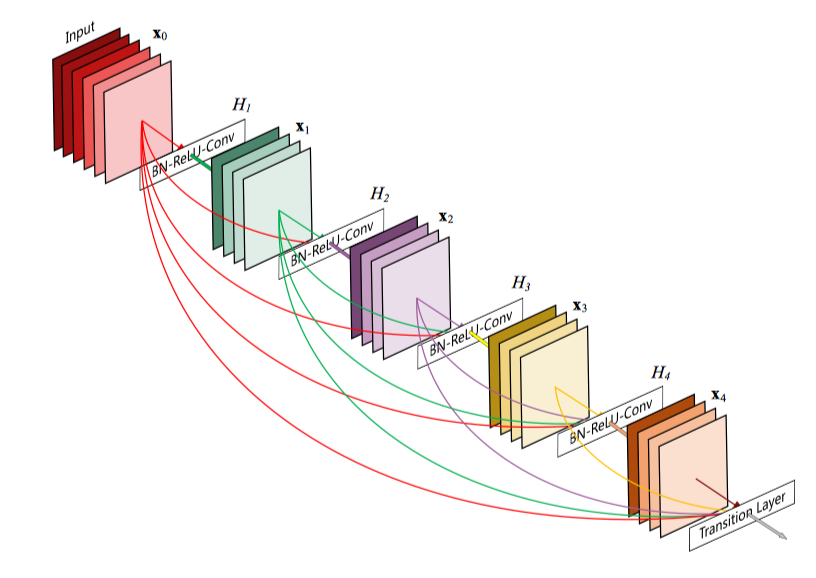}
\caption{A 5-layer dense block with a growth rate of $k = 4$ which is the number of feature-maps produced by $H_{l}$.\citep{huang2017densely}}
\label{denseblock}
\end{figure}

Densely Connected Convolutional Networks (DenseNet) \citep{huang2017densely} has achieved state-of-the-art performance in image recognition task recently. The Dense Block is crucial design in DenseNet. The key structure of Dense Block shown as in figure \ref{denseblock} is introducing "direct connections from any layer to all subsequent layers" \citep{huang2017densely}. The $l^{th}$ layer accepts the feature maps, $x_0, x_1, ..., x_{l-1}$, produced by all its prior layers as input:

\begin{equation}
\label{equForline}
\begin{aligned}
&x_l=H_l([x_0, x_1, ..., x_l])\\
\end{aligned}
\end{equation}

where $[x_0, x_1, ..., x_l]$ represents the concatenation of all the feature maps generated by layers $0, ..., l-1$ and $H_l$ refers to a composite function of three subsequent actions: batch normalization (BN), rectified linear unit (ReLU) and a $3 \times 3$ convolution (Conv).

\begin{figure}[H]
\centering
\includegraphics[width=15cm,height=2cm]{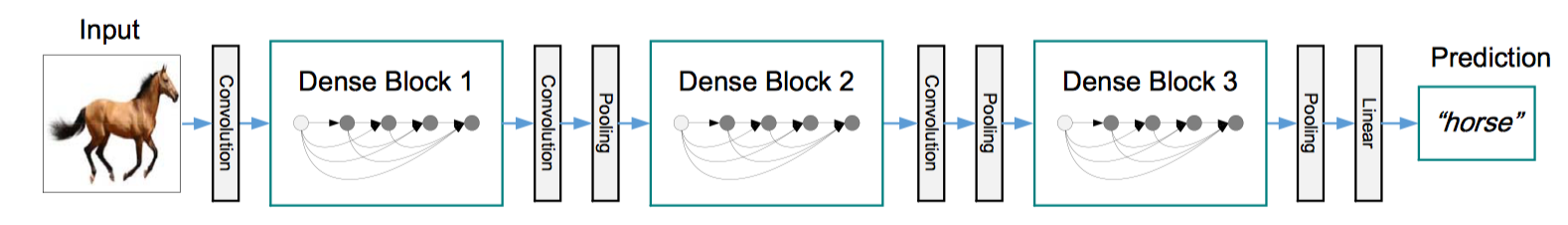}
\caption{A deep DenseNet with three dense blocks.\citep{huang2017densely}}
\label{denseArch}
\end{figure}

A DenseNet having 3 dense blocks is shown in figure \ref{denseArch}. The $1 \times 1$ conv layer and $2 \times 2$ max pooling layer with stride 2 between two neighboring dense blocks function as transition layer which is used to change feature map size. Bottleneck layers are also used to change the all inputs from different prior layers before every $3 \times 3$ conv layer.

DenseNet121 and DenseNet169 have growth rate ${k = 32}$ and each conv layer has structure of BN-ReLU-Conv. Details can be refered to Table 1 in \cite{huang2017densely}.

\subsection{RNN}
\begin{figure}[H]
\centering
\includegraphics[width=3cm,height=5cm]{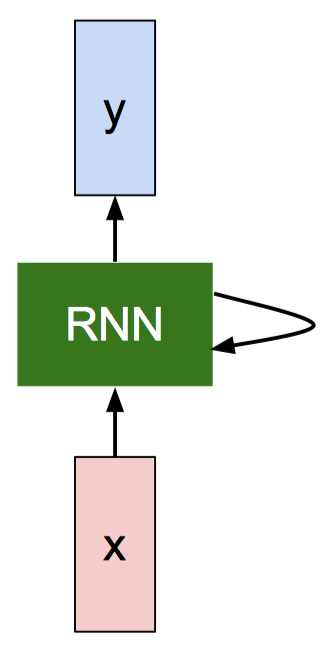}
\caption{RNNs Architecture from http://cs231n.stanford.edu.}
\label{rnn}
\end{figure}

Recurrent neural networks (RNNs) is good at processing sequential data. The architecture of RNNs has a loop in time dimension as shown in Figure \ref{rnn}. The internal state will be fed back to the model at next time stamp together with the input read in.

\begin{figure}[H]
\centering
\includegraphics[width=12cm,height=4cm]{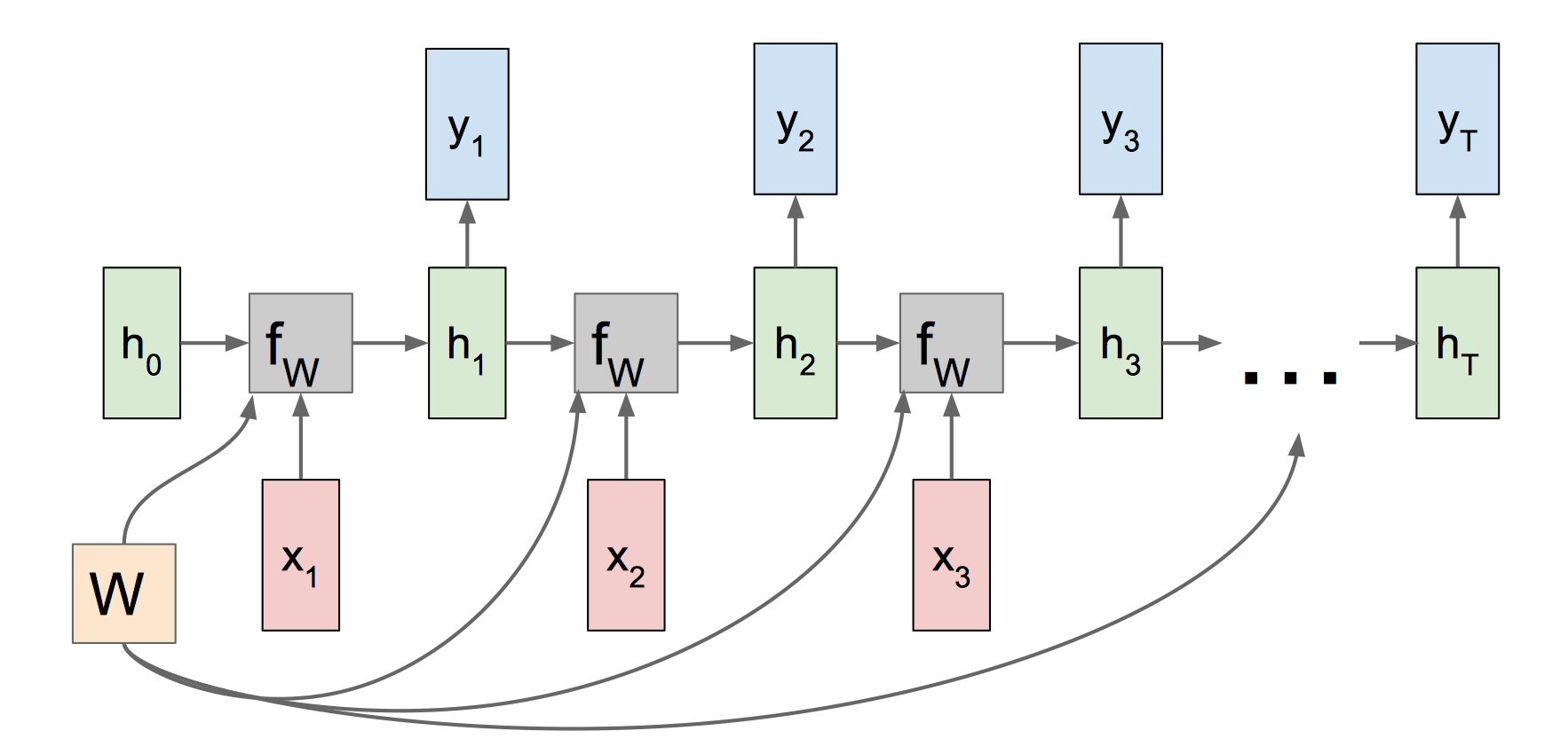}
\caption{Unrolled RNNs Architecture from http://cs231n.stanford.edu.}
\label{unroll}
\end{figure}

An unrolled version is shown in figure \ref{unroll}.

where we have input data $x_{t}$ and every hidden state vector $h_{t}$ at time stamp $t$, which can be updated by previous time stamp hidden state ${h_{t-1}}$ as:

\begin{equation}
\label{equForline}
\begin{aligned}
&h_{t}= f_{W}(h_{t-1}, x_{t})\\
&=tanh(W_{hh}h_{t-1} + W_{xh}x_{t} + b)\\
\end{aligned}
\end{equation}

in which trainable parameters $W_{hh}$, $W_{xh}$ are weight matrices and ${b}$ is a bias vector. Although RNN is good at modelling sequential data, long term dependency problem can not be solved.

\subsection{LSTM}
\label{LSTMsection}
Long short-term memory (LSTM) is a special designed RNN which has the ability to avoid the gradient vanishing problem in conventional RNN, which refer to loss gradients getting close to zero values in back propagation during training process. Similar to vanilla RNN, LSTM also has repeating modules. But rather than has single ${tanh}$ layer in one module, LSTM has more complex structure in one module. Apart from having hidden state $h_{t}$, LSTM also has cell state $c_{t}$ at every time stamp, where cell state is the memory unit of the network and is the key design of LSTM. Gates are used to control how information is passed to the cell state, namely decide what to keep and what to discard. Following Figure \ref{lstm} show how the internal operations interact with each other in a single LSTM module.

\begin{figure}[H]
\centering
\includegraphics[width=14cm,height=6cm]{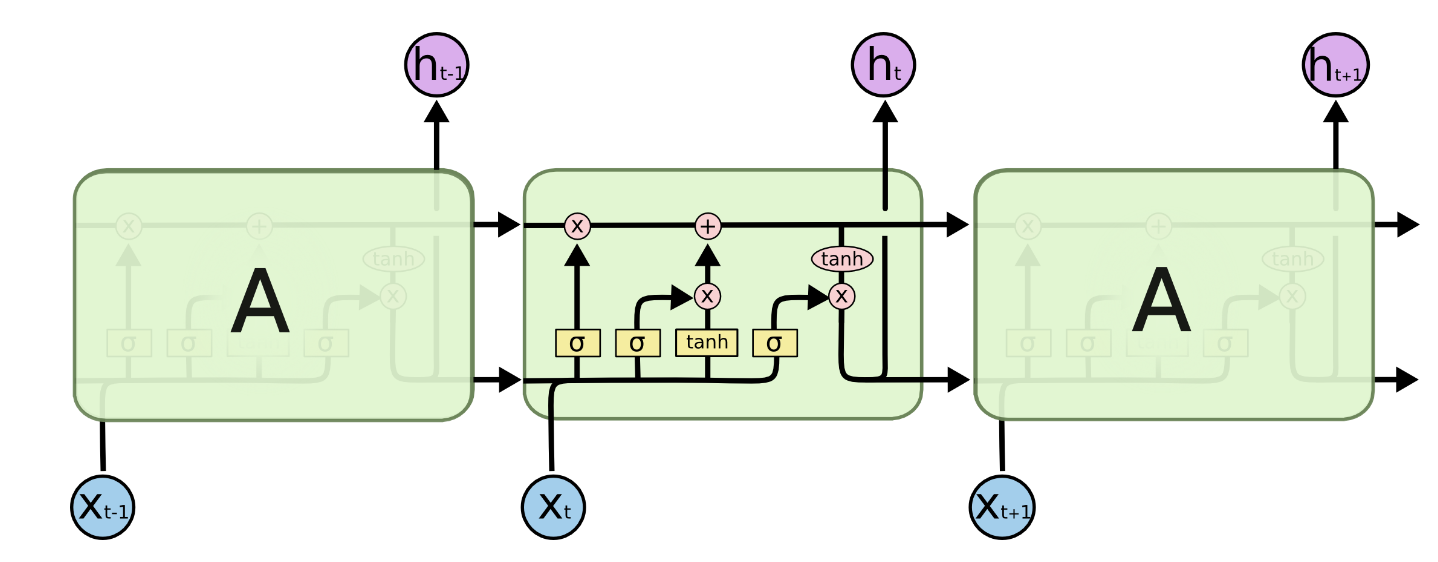}
\caption{An LSTM module has four internal layers interacting with each other at time step t. The figure is taken from: http://colah.github.io/posts/2015-08-Understanding-LSTMs/. }
\label{lstm}
\end{figure}

Forget gate vector $f_{t}$ at time stamp $t$ controls to what extent to erase content of previous cell state ${c_{t-1}}$ :

\begin{equation}
\label{equForline}
\begin{aligned}
&f_{t}= \sigma(W_{f}x_t + U_{f}h_{t−1} + b_{f})
\end{aligned}
\end{equation}

Input gate vector $i_{t}$ controls what part of candidate values will be updated:

\begin{equation}
\label{equForline}
\begin{aligned}
&i_{t}= \sigma(W_{i}x_t + U_{i}h_{t−1} + b_{i})
\end{aligned}
\end{equation}

Additionally, a vector of candidate values which means something that could be added to the cell state is computed as:

\begin{equation}
\label{equForline}
\begin{aligned}
&g_{t}= tanh(W_{g}x_t + U_{g}h_{t−1} + b_{g})
\end{aligned}
\end{equation}

The input gate and candidate values together decide what new information is going to be stored in cell state. Then the new cell state will be constituted by the outcome from the previous cell state and the new candidate values as shown below:

\begin{equation}
\label{equForline}
\begin{aligned}
&c_{t}= f_{t} \bigodot c_{t-1} + i_{t} \bigodot g_{t}
\end{aligned}
\end{equation}

where $\bigodot$ represents operation of element-wise product. Output gate vector controls how much to reveal to outside world at every time stamp ${t}$:

\begin{equation}
\label{equForline}
\begin{aligned}
&o_{t}= \sigma(W_{o}x_t + U_{o}h_{t−1} + b_{o})
\end{aligned}
\end{equation}

Finally, the new hidden state will be updated as:

\begin{equation}
\label{equForline}
\begin{aligned}
&h_{t}= o_{t} \bigodot tanh(c_{t})
\end{aligned}
\end{equation}

Therefore the set of trainable parameters of the LSTM is: 
\newline
${\theta = W_{f} , W_{i} , W_{g} , W_{o} , U_{f} , U{i} , U_{g} , U_{o} , b_{f} , b_{i} , b_{c} , b_{o}}$.

\subsection{GRU}
\label{GRUsection}
Gated Recurrent Unit (GRU) is an improved version of standard LSTM. GRU uses update gate and reset gate which controls what information should be passed to the output to avoid gradient vanishing problem in RNNs. Compared with LSTM, GRU uses update gate replaces the forget and input gate. In addition, the cell state and hidden state are merged together. The advantage of GRU is that the calculation is simplified compared with LSTM and the expression of the model is also strong, so GRU becomes more and more popular.

Here is the Figure \ref{gru} for the structure of  GRU:
\begin{figure}[H]
\centering
\includegraphics[width=11cm,height=6cm]{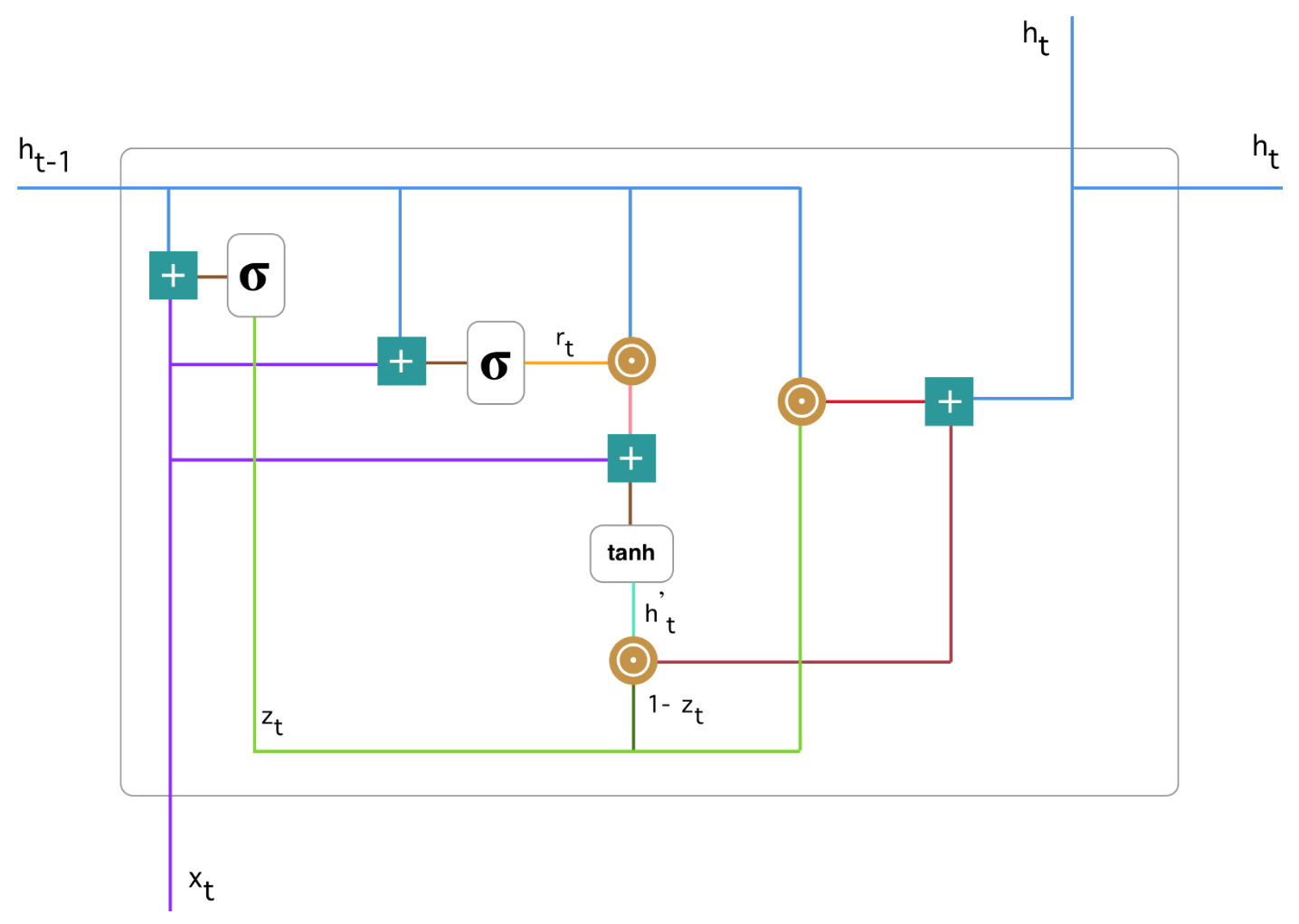}
\caption{GRU structure from : https://towardsdatascience.com/understanding-gru-networks-2ef37df6c9be. }
\label{gru}
\end{figure}

In this GRU structure, update gate $z_{t}$ for time stamp $t$ helps the model to determine how much of the past information needs to be passed along to the future by calculating:

\begin{equation}
\label{equForline}
\begin{aligned}
&z_{t}= \sigma(W_{z}x_{t} + U_{z} h_{t-1} + b_{z})
\end{aligned}
\end{equation}

where $x_{t}$ is the current time stamp input, $h_{t-1}$ is the previous time stamp memory content.

Reset gate $r_{t}$ is used to decide how much of the past information to forget from the model by computing:

\begin{equation}
\label{equForline}
\begin{aligned}
&r_{t}= \sigma(W_{r}x_{t} +U_{r} h_{t-1}+b_{r})
\end{aligned}
\end{equation}

Although it has same function form with update gate but it has different trained weights and usage.

Current memory content $h_{t}^{'}$ is calculated as :
\begin{equation}
\label{equForline}
\begin{aligned}
&h_{t}^{'}= tanh(W_{c}x_{t} + r_{t}\bigodot U_{c} h_{t-1})
\end{aligned}
\end{equation}

which determines what to remove from previous time stamp. Here, $\bigodot$ stands for Hadamard product.

Finally, the final memory at current state is updated as:
\begin{equation}
\label{equForline}
\begin{aligned}
&h_{t}= z_{t}\bigodot h_{t-1} + (1 - z_{t} )\bigodot h_{t}^{'}
\end{aligned}
\end{equation}

Therefore the set of trainable parameters of the GRU is: 
\newline
$\theta = {W_{z}, W_{r} , W_{c}, U_{z}, U{r}, U_{c}, b_{z}, b_{r}}$.


\section{Summary}
\label{backgroundSummary}
Emotion recognition can be solved by using categorical and dimensional way. AUs is also used to describe facial expressions. Compared with Seven Basic Emotions and Action Units, Valence and Arousal, as dimensional measurements, are able to capture more subtle, complex and diverse facial expressions. Since every emotion can be represented by a coordinate in 2-D Emotion Wheel shown in Figure \ref{EW}. Existing databases annotated in terms of valence and arousal has somewhat limited  drawbacks. SEMAINE database and RECOLA database are recorded in well-controlled conditions not close to real wolrd; SEWA database are captured in limited scenario; AFEW-VA database has annotations of 21 levels, which is not very capable of expressing various emotions, for valence and arousal. In this case, Aff-Wild database is more close to real world environment containing spontaneous emotions annotated with continuous valence and arousal. But compared with AFEW-VA, in which the annotations are produced frame by frame, the annotations for Aff-Wild may be less accurate. But it is more efficient to produce annotations using method similar to FEELTRACE.\newline

What is more, almost all the winning methods in emotion recognition challenges use deep learning approaches, which shows deep neural network \cite{kollia2009interweaving} is very satisfactory to be used in this affect analysis area. In \cite{jaiswal2016deep}, \cite{kollias4} and \cite{Chen2017MultimodalRecognition}, temporal model, like LSTM mentioned in \ref{LSTMsection} and GRU mentioned in \ref{GRUsection}, is shown to have outstanding performance in modelling dynamics in sequence of data. Moreover, deep CNNs, like VGGNet, ResNet and DenseNet, usually have state-of-the-art performance at image recognition task. When it comes to particular domain like human face recognition, pre-trained model, like VGGFace network, is also verified that has excellent performance in extracting facial features in similar task as said in \cite{kollias2} and \cite{Chen2017MultimodalRecognition}. 
\chapter{Database}
\label{databaseChapter}
\begin{figure}[H]
    \centering
    \subfigure[]
    {\includegraphics[height=1.5cm,width=1.5cm]{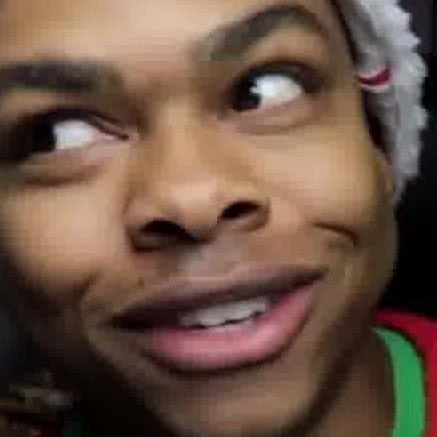}}
    \subfigure[]
    {\includegraphics[height=1.5cm,width=1.5cm]{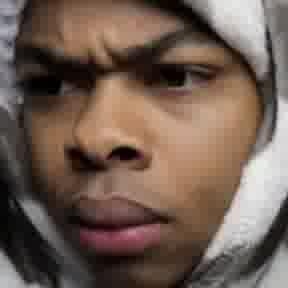}}
    \subfigure[]
    {\includegraphics[height=1.5cm,width=1.5cm]{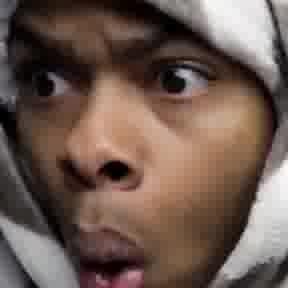}}
    \subfigure[]
    {\includegraphics[height=1.5cm,width=1.5cm]{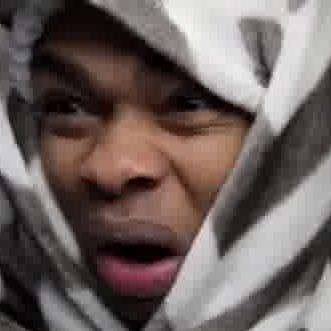}}
    \subfigure[]
    {\includegraphics[height=1.5cm,width=1.5cm]{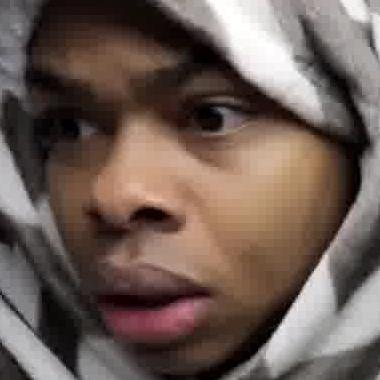}}
    \subfigure[]
    {\includegraphics[height=1.5cm,width=1.5cm]{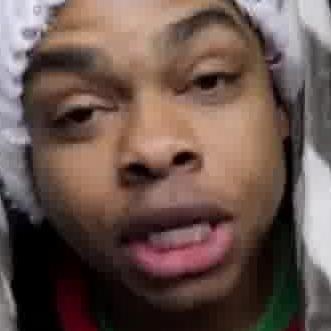}}
    \subfigure[]
    {\includegraphics[height=1.5cm,width=1.5cm]{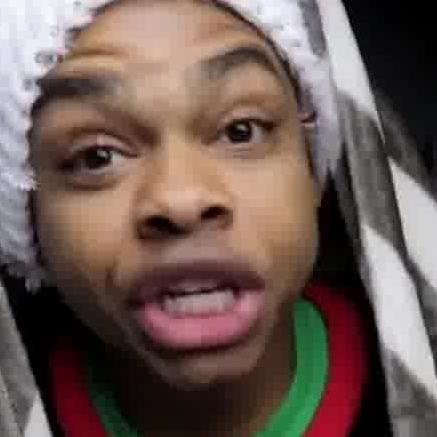}}
    \subfigure[]
    {\includegraphics[height=1.5cm,width=1.5cm]{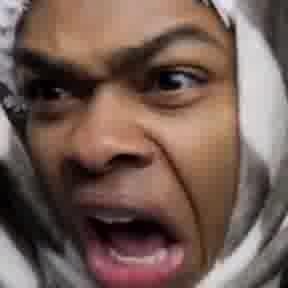}}
    \caption{Annotated Facial Expressions}
    \label{sequence}
\end{figure}

This chapter mainly demonstrates the process for extending the Aff-Wild Database \citep{kollias1}. Extending the Aff-Wild Database aims to enrich the training data so that the well-trained deep neural networks can be more generalized. The resulting built database consists of the train, validate and test parts, having 100, 29 and 30 videos respectively. Similar to Aff-Wild database, the collected videos contains spontaneous facial behaviours under arbitrary conditions which refers to real life, uncontrolled conditions such as various brightness environments and face occlusion. Some representative frames are shown in Figure \ref{challenging}:

\begin{figure}[H]
\centering
\subfigure[]
{\includegraphics[height=1.5cm,width=1.5cm]{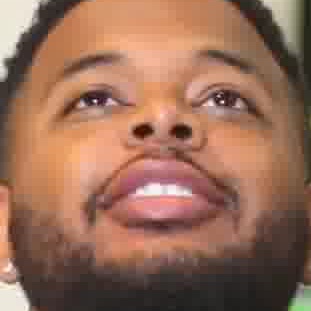}}
\subfigure[]
{\includegraphics[height=1.5cm,width=1.5cm]{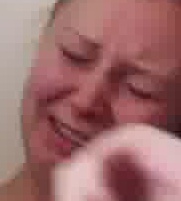}}
\subfigure[]
{\includegraphics[height=1.5cm,width=1.5cm]{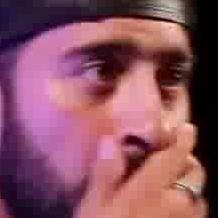}}
\subfigure[]
{\includegraphics[height=1.5cm,width=1.5cm]{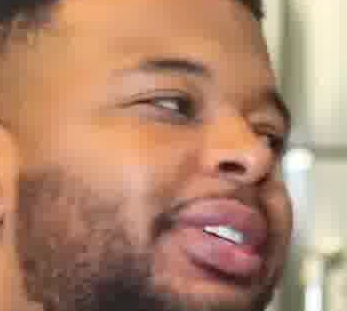}}
\subfigure[]
{\includegraphics[height=1.5cm,width=1.5cm]{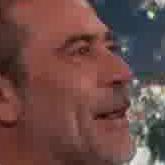}}
\subfigure[]
{\includegraphics[height=1.5cm,width=1.5cm]{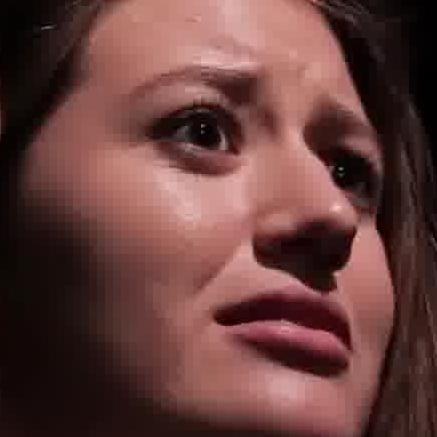}}
\subfigure[]
{\includegraphics[height=1.5cm,width=1.5cm]{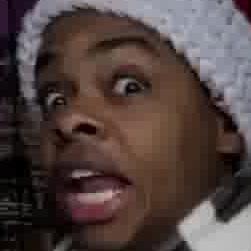}}
\subfigure[]
{\includegraphics[height=1.5cm,width=1.5cm]{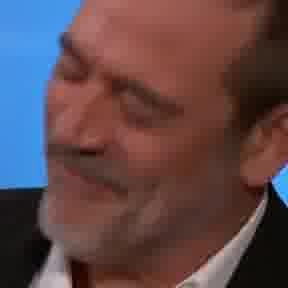}}
\caption{Challenging Frames}
\label{challenging}
\end{figure}
All the subjects in videos were annotated with valence and arousal for each frame. In Figure \ref{sequence}, there is a sequence of frames with valence and arousal annotations as shown in Table \ref{annotationForSequence} ranging in [-1000, 1000]. It can be verified that the annotations reflect the facial expressions well.

\begin{table}[H]
    \centering
    \begin{tabular}{|c|c|c|c|c|c|c|c|c|}
	\hline
    Frame & a & b & c & d & e & f & g & h \\
	\hline
    Valence & 274 & -151 & -311 & -431 & -103 & -103 &-151 & -351 \\
    \hline
    Arousal & 385 & 156 & 385 & 487 & 187 & 100 & 85 & 416 \\
    \hline
    \end{tabular}
    \caption{FPS values for 140 videos.}
    \label{annotationForSequence}
\end{table}

Since the Database Creation task aims to extend the Aff-Wild Database, the main approach of collecting and pre-processing the videos is similar to the method used to create Aff-Wild Database as stated in \citep{kollias1}. In this chapter, section \ref{collectData} illustrates the pipeline of database creation. Pre-processing of face detection is specified in section \ref{preprocess}, including the method of dealing with the detection results and the approach of matching the annotation with the corresponding frame. In section \ref{partition}, the detailed steps of partitioning the entire data set to train, validate and test set are depicted. \newline

This chapter mainly contributes to extending Aff-Wild database. In addition, the database built in this chapter is also used to train the deep neural network architecture constructed in Chapter \ref{designChapter}.

\section{Data Collection Procedure}
\label{collectData}
\begin{figure}[H]
\centering
\includegraphics[height=5cm,width=13cm]{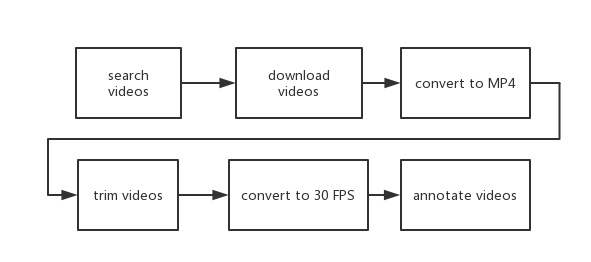}
\caption{Data Collection Procedure}
\label{procedure}
\end{figure}

The database creation procedure, as shown in Figure \ref{procedure}, mainly consists of searching suitable YouTube videos, downloading selected videos, converting to MP4 format, trimming of videos, converting all videos having different fps to chosen fps 30 and annotating videos with respect to valence and arousal.

\begin{enumerate}

    \item \textbf{Searching videos}: Similar to Aff-Wild database source, YouTube is the searching source for collecting videos because the spontaneous facial expressions under uncontrolled conditions are needed like figures shown in Figure \ref{challenging}. And one video should mainly contain one person's facial expression or at most two people. In this case, it is reasonable to analyze a sequence of facial expressions. The videos needed should contain facial expressions having various valence and arousal value since the resulting database should have balanced ground truth annotations. Regarding this, videos containing people reacting to diverse stimulation are searched by keywords like "reaction". For example, videos containing people doing meditation was selected since it is likely to have neutral valence and smaller arousal expression in these videos; videos about people watching Bear Grylls's wilderness survival television series was selected for there should be facial expressions with negative valence and bigger arousal value; videos for people receiving awards were chosen since positive valence expression should exist. What is more, the ethnicity and age diversity of the subjects in videos were also taken into consideration. The results at this stage are 140 links to the selected videos. 
    
	\item \textbf{Downloading videos} : With the links gotten from the searching stage, the videos had been downloaded from YouTube by using youtube-dl \citep{youtubedl} which is an open source tool from GitHub. For one particular video, best audio and best video having highest resolution were downloaded separately then merged into MKV format which can be the container for all formats of audio and video. For example, the download video could be in the format of webm which can not be merged into MP4 format but can be merged into MKV format. The results at this stage are 140 videos with best possible quality in MKV format. 
    
    \item \textbf{Converting format} : Since the downloaded videos were in MKV format but our annotator program \citep{kollias1} only accept MP4 format videos, the videos were converted from MKV format to MP4 format using the Format-Factory \citep{formatFactory} tool. The results at this stage are 140 videos in MP4 format. 
    
    \item \textbf{Trimming videos} : After that, I started to trim the videos in order to make videos must have human faces at the beginning and end of the videos which because in the later detecting process as stated in \ref{pick} the first and last frame is very important to be detected. For example, if a tracking technique was applied then there must be a human face in the first frame. Basically, the irrelevant content like advertisement and caption only frames were discarded. In terms of the time dimension, the videos which contain intermittent content were trimmed into a separate part. For example, if one video consists of two periods of videos with one recorded in day and another recorded at night,  then this video should be trimmed at the time where separate day and night content since when train the neural network using RNNs, it is very confused to recognize the pattern which is not consecutive in fact. For the convenient to annotate the videos later, long videos which have duration more than 10 minutes were also trimmed into separate parts since it is difficult to memorize that long content of the video to give appropriate annotation with respect to valence and arousal. It's worth noting that when two people appear in the video, only the part of two people appearing at the same time will be retained. The videos containing two subjects have been included twice in the video set for later annotating procedure. So each video with two subjects will add one more video to the entire data set and will affect the total number of frames and the male to female ratio. And both subjects appearing in the same video will be annotated with respect to valence and arousal. Here, FFmpeg \citep{ffmpeg} and iMovie \citep{iMovie} were used as trimming tool. The results at this stage are 159 videos. 
    
    \item \textbf{Converting FPS} : Then all videos were converted to have 30 FPS (which stands for frames per seconds) so that it is reasonable for RNN to accept a sequence of data having the same frequency. Following is the FPS information investigated before trimming:
    \begin{table}[H]
        \centering
        \begin{tabular}{|c|c|}
    	\hline
    	Frames Per Seconds & Total Number \\
    	\hline
        30 & 82 \\
        \hline
        24 & 27 \\
    	\hline
        25 & 17 \\
        \hline
        60 & 8 \\
        \hline
        29 & 2 \\
        \hline
        15 & 2 \\
        \hline
        14 & 1 \\
        \hline
        17 & 1 \\
        \hline
        \end{tabular}
        \caption{FPS values for 140 videos.}
        \label{fps-table}
    \end{table}
    From table \ref{fps-table}, we can see the majority has FPS around 30. Since trimming procedure does not affect FPS value for videos, the videos were converted after trimming. The results at this stage are 159 videos all having 30 FPS. 
    
    \item \textbf{Annotating arousal and valence values}: In order to train the deep learning networks, the ground truth value of valence and arousal for each frame are needed. To do the annotation, the videos were first watched thoroughly, then analysis of facial expression was done, finally, the videos were annotated using an annotator program \citep{kollias1} with a joystick. The pipeline is shown in Figure \ref{annotationPipeline}.
    
    \begin{figure}[H]
        \centering
        \includegraphics[height=3cm,width=13cm]{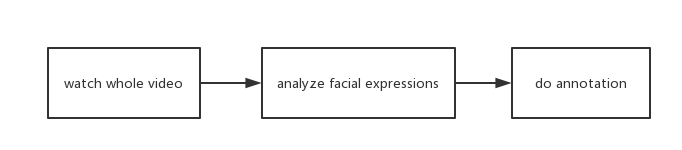}
        \caption{Annotation Pipeline}
        \label{annotationPipeline}
    \end{figure}

    During watching stage, the main task was to understand the content of the videos so that when analyzing the facial expressions, the context can help to judge the valence and arousal value. Here I also used Tampermonkey Chrome Plugin \citep{TampermonkeyChrome} to download the subtitle of YouTube videos so that I can study the precise (\cite{simou2007fire}) content of the videos.
    
    The most important part was the analysis of facial expression. Apart from the definition defined in \ref{VAdefinition}, here is the explanation with an example of valence and arousal considering the content in the videos: 
    
    For valence:
    \begin{itemize}
        \item Positive valence can be represented by expression like happy or exciting shown in Figure \ref{positive}.
            \begin{figure}[H]
            \centering
            \subfigure[Happy]
            {\includegraphics[height=2cm,width=2cm]{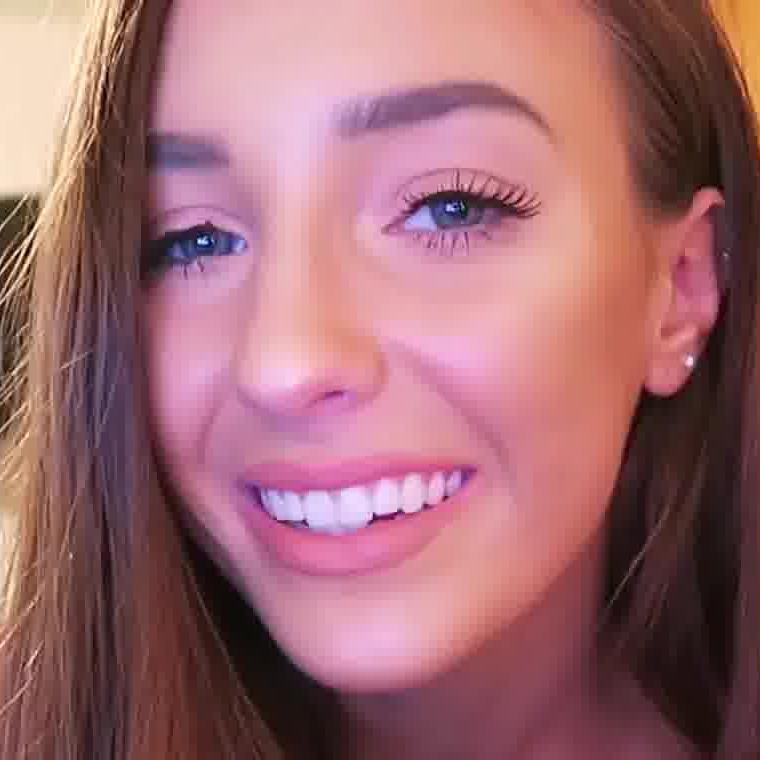}}
            \subfigure[Happy]
            {\includegraphics[height=2cm,width=2cm]{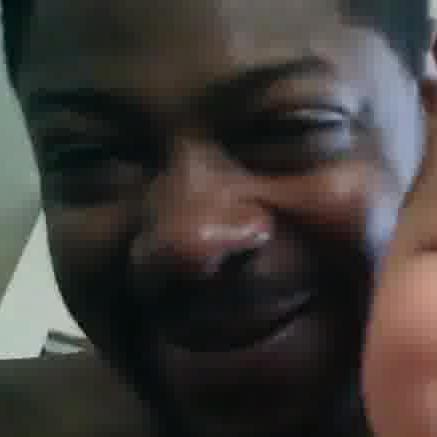}}
            \subfigure[Happy]
            {\includegraphics[height=2cm,width=2cm]{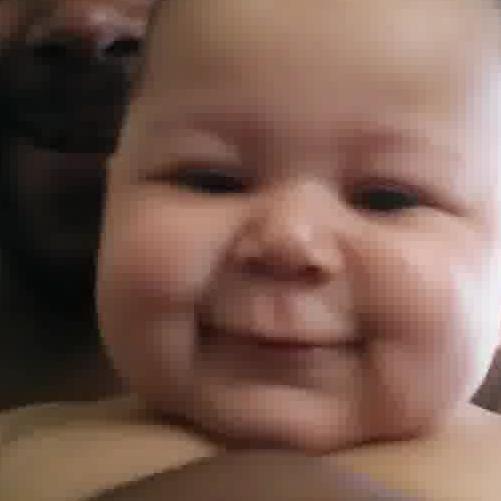}}
            \subfigure[Exciting]
            {\includegraphics[height=2cm,width=2cm]{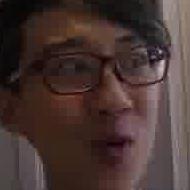}}
            \caption{Positive Valence Facial Expression}
            \label{positive}
            \end{figure}
        \item Neutral valence stands for no apparent positive or negative emotion like when people just talking but do not have any emotion shown up on face like in Figure \ref{neutral}.
            \begin{figure}[H]
            \centering
            \subfigure[Neutral]
            {\includegraphics[height=2cm,width=2cm]{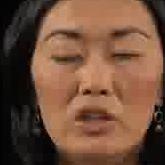}}
            \subfigure[Neutral]
            {\includegraphics[height=2cm,width=2cm]{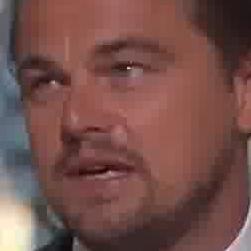}}
            \subfigure[Neutral]
            {\includegraphics[height=2cm,width=2cm]{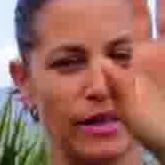}}
            \subfigure[Neutral]
            {\includegraphics[height=2cm,width=2cm]{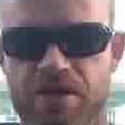}}
            \caption{Neutral Valence Facial Expression}
            \label{neutral}
            \end{figure}
        \item Negative valence can be disgusting, angry, sad or fear shown in Figure \ref{negative}.
            \begin{figure}[H]
            \centering
            \subfigure[Disgusting]
            {\includegraphics[height=2cm,width=2cm]{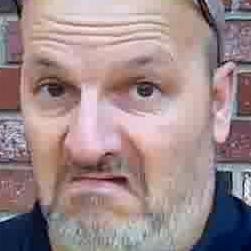}}
            \subfigure[Sad]
            {\includegraphics[height=2cm,width=2cm]{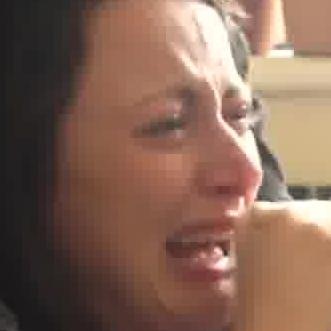}}
            \subfigure[Angry]
            {\includegraphics[height=2cm,width=2cm]{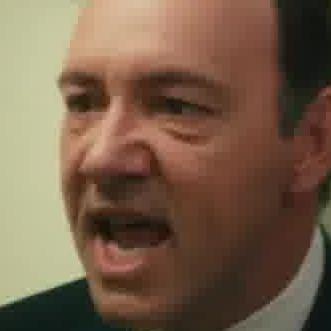}}
            \subfigure[Fear]
            {\includegraphics[height=2cm,width=2cm]{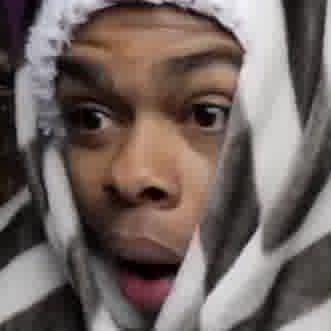}}
            \caption{Negative Valence Facial Expression}
            \label{negative}
            \end{figure}
        
    \end{itemize}
    
    For arousal:
    \begin{itemize}
        \item Positive arousal means the looking of a particular face is to some extent expressive. Like in Figure \ref{expressive}: 
         \begin{figure}[H]
            \centering
            \subfigure[Expressive]
            {\includegraphics[height=2cm,width=2cm]{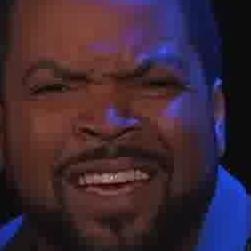}}
            \subfigure[Expressive]
            {\includegraphics[height=2cm,width=2cm]{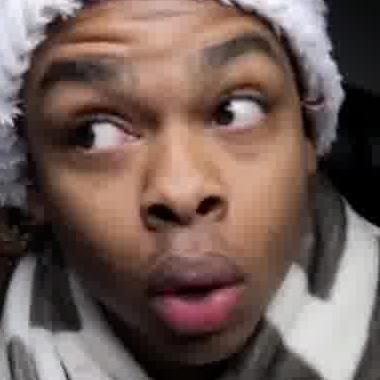}}
            \subfigure[Expressive]
            {\includegraphics[height=2cm,width=2cm]{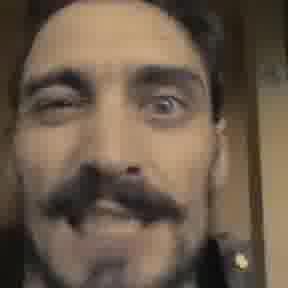}}
            \subfigure[Expressive]
            {\includegraphics[height=2cm,width=2cm]{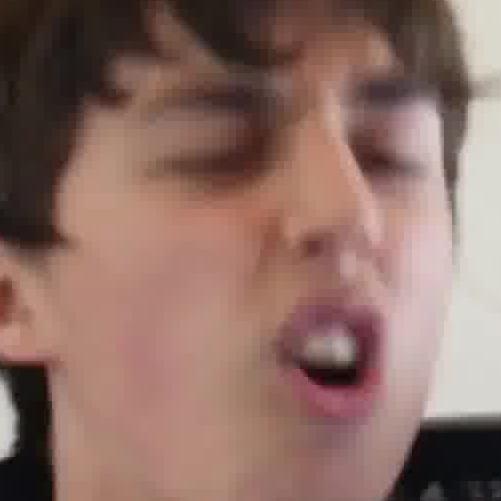}}
            \caption{Expressive Facial Expression}
            \label{expressive}
            \end{figure}
        \item 0 arousal refers to a situation that people have some feeling inside but do not show it. For example, when people watch videos they are active to what they see although they do not show.
            \begin{figure}[H]
            \centering
            \subfigure[Watching Video]
            {\includegraphics[height=2cm,width=2cm]{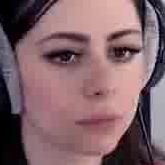}}
            \subfigure[Watching Video]
            {\includegraphics[height=2cm,width=2cm]{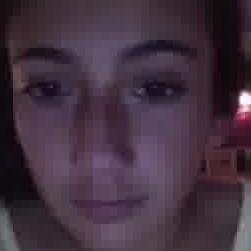}}
            \subfigure[0 Arousal]
            {\includegraphics[height=2cm,width=2cm]{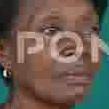}}
            \subfigure[Watching Video]
            {\includegraphics[height=2cm,width=2cm]{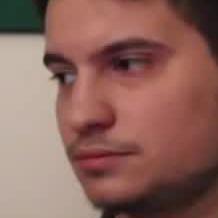}}
            \caption{Zero Arousal Facial Expression}
            \end{figure}
        \item Examples for negative arousal can be the facial expression of people who is sleeping. But in the database collected by myself, passive expression is rare. 
    \end{itemize}
     
    And during the process of analyzing the facial expression, more attention should be paid under some special circumstances.
    \begin{itemize}
        \item Situation One: When annotating the videos, the value of valence and arousal should be given just for every moment not overall sequence of frames. Like when giving annotation for people watch videos, most of the time they may show no emotion. Only when they see a funny or sad plot, will they suddenly laugh or show sad expression.
        \item Situation Two: Regarding valence, sometimes the annotation cannot be given directly based on what they show. Instead, context information should be considered to help make a judgment. Like positive value should be given to the expression of people burst into tears when they are receiving awards. Although they are crying, they feel so happy and grateful. As shown in Figure \ref{happyTears}.
        \begin{figure}[H]
            \centering
            \subfigure[]
            {\includegraphics[height=2cm,width=2cm]{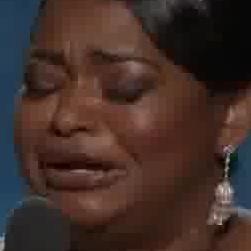}}
            \subfigure[]
            {\includegraphics[height=2cm,width=2cm]{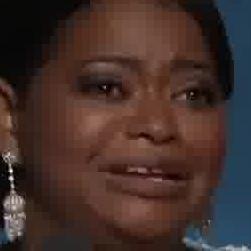}}
            \caption{Cry When Receiving Awards}
            \label{happyTears}
        \end{figure}
        
        Another example is negative value should be given to the expression of people smiles when they are talking about something in a sarcastic tone. Since this kind of smile is irony shown in Figure \ref{irony}.
        \begin{figure}[H]
            \centering
            \subfigure[]
            {\includegraphics[height=2cm,width=2cm]{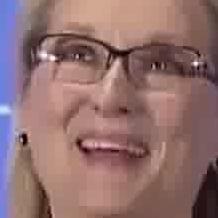}}
            \subfigure[]
            {\includegraphics[height=2cm,width=2cm]{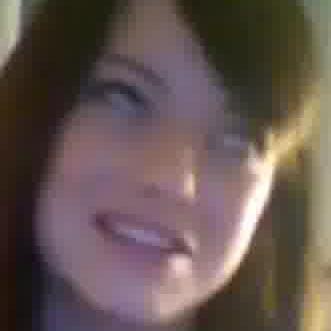}}
            \caption{Irony Smile}
            \label{irony}
        \end{figure}
    \end{itemize}
    
    Finally, the videos were annotated with an annotator program tool\citep{kollias1}. First, the joystick \citep{LogitechAttacker3} need to be selected and the identifier was provided in order to create the corresponding folder to save the annotation results. As shown in Figure \ref{login}.
    
    \begin{figure}[H]
    \centering
    \includegraphics[height=3cm,width=7cm]{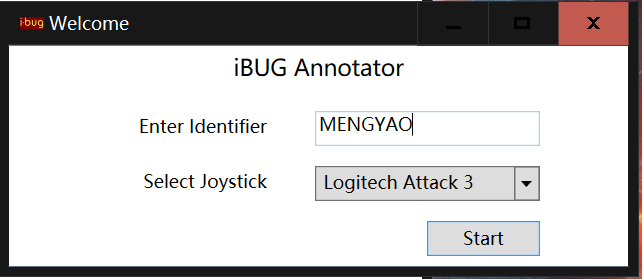}
    \caption{Login Interface}
    \label{login}
    \end{figure}

    Then the videos ready for annotating were listed on the left-hand  side and the videos already annotated were on the right-hand side as in Figure \ref{videoslist}.
    
    \begin{figure}[H]
    \centering
    \includegraphics[height=8cm,width=12cm]{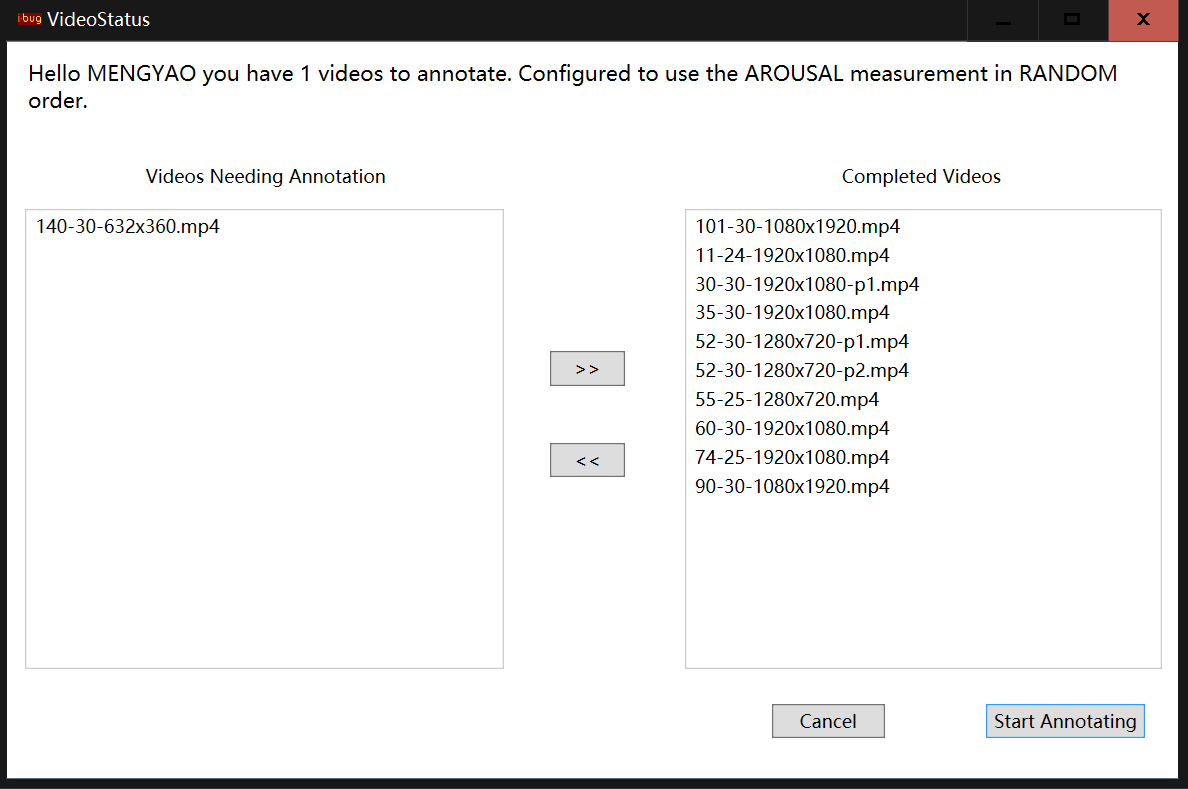}
    \caption{Videos List}
    \label{videoslist}
    \end{figure}
    
    For annotating the videos, every time we only focus on one dimension, valence or arousal. And while the video displaying, the annotation was given by moving the joystick at the same time. The interface is shown in Figure \ref{annotating}.
    
    \begin{figure}[H]
    \centering
    \includegraphics[height=7cm,width=10cm]{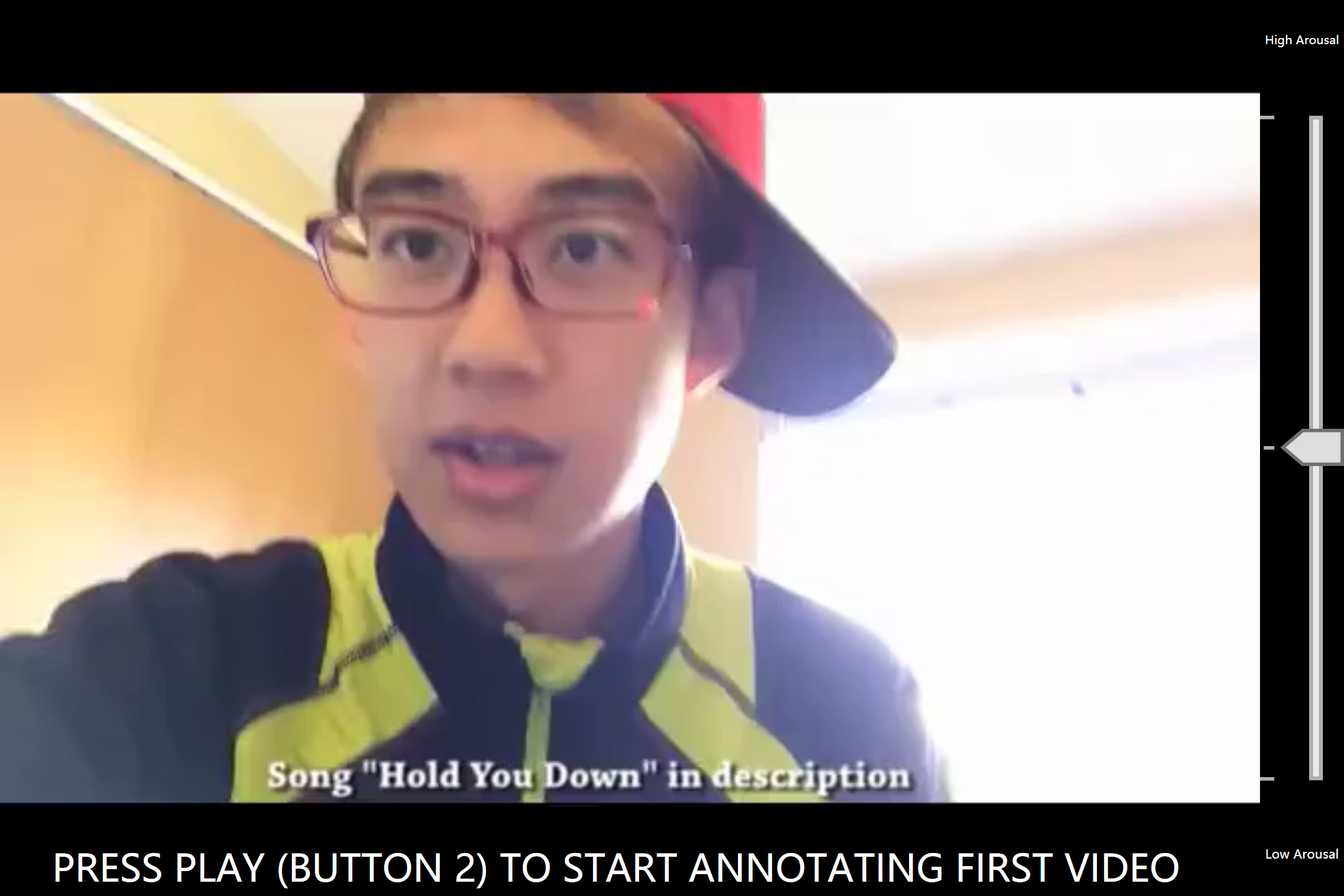}
    \caption{The GUI of the annotation tool when annotating arousal (the GUI for valence is exactly the same).}
    \label{annotating}
    \end{figure}
    
    With the tool, time-continuous annotation ranged from -1000 to 1000 were generated into one text file for each video for each dimension. The results for this stage are 159 annotation files for valence and 159 annotation files for arousal. Resulting example annotations are shown in Figure \ref{sequence}. 
\end{enumerate}

\section{Data Pre-processing}
\label{preprocess}
\subsection{Face Detection}
After finished the annotation for the videos, a pre-processing task aimed to find the human faces in all frames of the videos. There are many detectors \cite{avrithis2000broadcast} evaluated for deformable face tracking "in the wild" in \citep{chrysos2018comprehensive}, through the thorough experiments, the weakly supervised Deformable Parts Model(DPM) provided by \citep{mathias2014face} implemented in \citep{alabort2014menpo} has the best performance in terms of face detection tasks. So the FFLD2Detector which is the corresponding implementation class in Menpo Software\citep{alabort2014menpo} was used as a detector to detect the subjects faces in each frame of videos. So the pipeline for cropping the human faces was first to extract all the frames using the FFmpeg program \citep{ffmpeg} from all videos. Then FFLD2Detector was applied successively for every frame extracted and generated bounding boxes representing the face locations in one frame. All the detected objects (not all detected objects are the human face) were cropped using bounding boxes and saved. Since this DPM is extremely computationally expensive, the videos collected were partitioned into four parts and the detection programs were running on four computers in Doc lab for these four parts video data. This procedure cost around five days. However, there was still a problem. 

\subsection{Filer Cropped Objects}
\label{pick}
\begin{figure}[H]
    \centering
    \subfigure[Wrongly detected face]
    {\includegraphics[height=2cm,width=2cm]{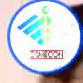}}
    \subfigure[True face]
    {\includegraphics[height=2cm,width=2cm]{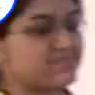}}
    \caption{Detected objects for one frame.}
    \label{faces}
\end{figure}

The videos collected most have only one subject, with several videos having at most two subjects. For most videos, the detected results were perfect. But for several videos, other than human faces, other objects were considered as human faces by the detector and saved. One example is shown in Figure \ref{faces}. 

At first, I wanted to apply the top performance trackers like SRDCF as evaluated in \citep{chrysos2018comprehensive}. But they are implemented in Matlab. Considering the time limitation and convenience, I tried all trackers provided in OpenCV \citep{OpenCV:API} instead. However, the detected faces produced by the trackers were of poor quality, namely the position of the bounding boxes produced by the trackers were not as correct as the ones produced by FFLD2Detector. So I decided to come up with a solution to pick the right face among all detected objects produced by FFLD2Detector for one frame. It is obvious that features extracted from images can be used here to compare the similarity for the different objects. For the simplicity, I also used the tools provided in OpenCV library \citep{opencv_library}. The different feature extracting tools can be selected for different situations. For the case shown in Figure \ref{faces}, the API used to calculate the histograms, \textit{cv2.calcHist()}, can be used to calculate the RGB channels histogram since these two objects have quite different RGB distribution apparently. Then use \textit{cv2.compareHist()} function to compute the similarities between all detected objects from one frame and the real face. Then the object which has the highest similarity value is chosen for this frame. In this approach, most detected results can be filtered correctly. If the problem was still unsolved, the faces were picked manually.   

After this procedure, the 159 videos became 159 folders of cropped faces images whose name are their corresponding frame number generated by FFmpeg program \citep{ffmpeg} when extracted from the video.

\subsection{Matching Process}
At this stage, the cropped faces and annotations for every video were prepared well. The task then was to match the faces and their corresponding valence and arousal. The annotations are in the format of two attributes, with one for time stamp and another one value for valence or arousal. Like the data shown in Table \ref{tab:annotation}. 

\begin{table}[H]
    \centering
    \begin{tabular}{|c|c|}
	\hline
	Time Stamps & Annotation  \\
	\hline
    0.010 & 121 \\
	\hline
    0.030 & 122 \\
    \hline
    0.041 & 123 \\
    \hline
    0.057 & 124 \\
    \hline
    0.089 & 125 \\
    \hline
    0.102 & 126 \\
    \hline
    0.119 & 127 \\
	\hline
    \end{tabular}
    \caption{Example of Annotation File Content.}
    \label{tab:annotation}
\end{table}

The annotation was generated by the annotator program at random time stamp. While the frame number for the cropped face is an integer like 1, 2, 3 which stands for the relative position of the frame in the video. So the annotation cannot be used directly for the cropped face. The nearest neighbour algorithm described in \citep{kollias1} was applied. In more details, for one integer frame number, compute its corresponding time stamp by doing multiplication with the time interval of one frame. This time interval is 0.03333 because all the videos are in 30 FPS (stands for frames per second). Then find the closest time stamp in the annotation file for the computed time stamp for one particular frame and assign the corresponding annotation for this frame. For example, for frame number 2, the corresponding time stamp is 0.06667. The closest time stamp in Table \ref{tab:annotation} is 0.057. So the annotation for frame 2 is 124. For both valence and arousal annotation files, this nearest neighbour algorithm was applied. Then the valence and arousal annotations were merged into one annotation file for each video in a format of the frame number, valence annotation, arousal annotation like in Table \ref{tab:mergeAnnotation}.  

\begin{table}[H]
    \centering
    \begin{tabular}{|c|c|c|}
	\hline
	Frame Number & Valence & Arousal  \\
	\hline
    1 & 121 & 121 \\
	\hline
    2 & 122 & 122 \\
    \hline
    3 & 123 & 123 \\
    \hline
    4 & 124 & 124 \\
    \hline
    5 & 125 & 125 \\
    \hline
    6 & 126 & 126 \\
    \hline
    7 & 127 & 127 \\
	\hline
    \end{tabular}
    \caption{Example of Merged Annotation File For One Video.}
    \label{tab:mergeAnnotation}
\end{table}

The result after matching process was 159 annotation files for each video.

\section{Partition for Database}
\label{partition}
In total, the resulting dataset contains 159 videos having 756,424 frames. There are 135 non-repeating subjects with 73 females and 62 males. For the purpose of training the deep learning neural network, the entire data set was supposed to be divided into the train, validate and test sets according to a ratio of 64\%, 16\% and 20\% respectively. So with this ratio applied to the number of videos, there should be 103, 25 and 31 videos in the train, validate and test data set. In order to make the divided three data sets balanced, the following three specifications should be obeyed:

\begin{enumerate}
    \item Criterion 1 : Annotation distribution in terms of the value for valence and arousal should be similar in three data sets.
    \item Criterion 2 : The number of male and female should be approximately the same in three data set. Since in 159 videos, 80 videos have the female face and 79 videos have the male face. 
    \item Criterion 3 : The person appearing in the three datasets ought to be independent from each other. 
\end{enumerate}

What I did to achieve this is first partition all the videos based on their major valence and arousal distribution. To be concrete, the whole set of videos were classified into four categories: mainly positive, mainly negative, neutral and having both sides value of valence. The ratio for these four classes is 43:58:46:12 (with a total number is 159). Let's call this Partition Ratio. So in initial partition, for each database, the corresponding number of videos were selected from those four categories with respect to the Partition Ratio. The results at this stage is shown in Table \ref{tab:initPartition}:

\begin{table}[H]
    \centering
    \begin{tabular}{|c|c|c|c|c|c|}
	\hline
	No. of Videos & Mainly Positive & Mainly Negative & Both Valence & Neutral & Total \\
	\hline
    Train & 27 & 38 & 29 & 9 & 103 \\
	\hline
    Validate & 7 & 9 & 8 & 1 & 25\\
	\hline
	Test & 9 & 11 & 9 & 2 & 31\\
	\hline
	Total & 43 & 58 & 46 & 12 & 159\\
	\hline
    \end{tabular}
    \caption{Number of Videos Partition for Three Data Set.}
    \label{tab:initPartition}
\end{table}

Here the arousal was not paid too much attention since the arousal value should be relatively balanced in the positive range. But the final partition was modified a little based on the ratio of the number of frames in three data set. 

The final partition for three data set is shown in Table \ref{tab:frames} from which we can see the partition ratio is close to the initial target, namely 64\% for the train, 16\% for the validate and 20\% for the test. 
\begin{table}[H]
    \centering
    \begin{tabular}{|c|c|c|c|c|}
	\hline
	Database & Train & Validate & Test & Total \\
	\hline
    No. of Frames & 527056(69.67\%) & 94223(12.46\%) & 135145(17.87\%) & 756424 \\
	\hline
    \end{tabular}
    \caption{Number of Frames in Three Data Set.}
    \label{tab:frames}
\end{table}

\begin{figure}[H]
    \centering
    \subfigure[Train]
    {\includegraphics[height=5cm,width=6.5cm]{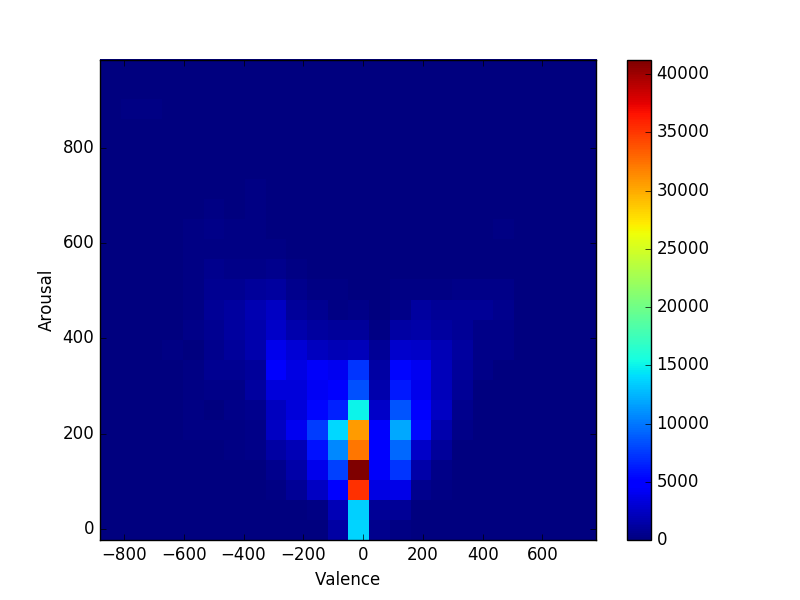}}
    \subfigure[Validation]
    {\includegraphics[height=5cm,width=6.5cm]{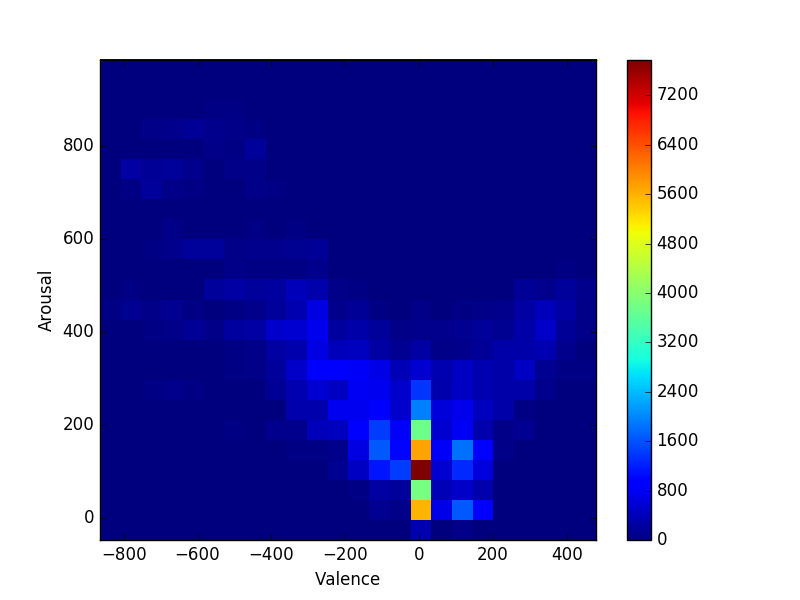}}
    \subfigure[Test]
    {\includegraphics[height=5cm,width=6.5cm]{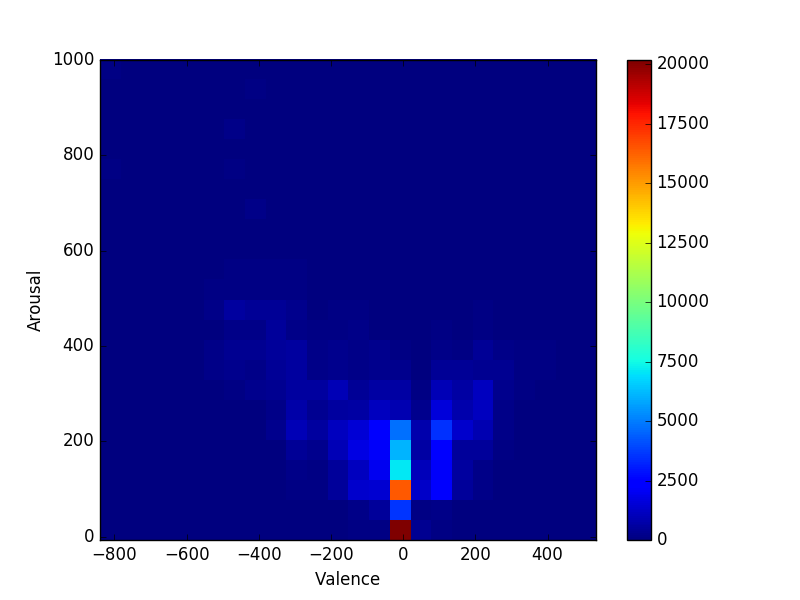}}
    \caption{Annotation Distribution}
    \label{annotationDistribution}
\end{figure}

\begin{figure}[H]
    \centering
    \subfigure[Valence Annotations]
    {\includegraphics[height=4cm,width=7cm]{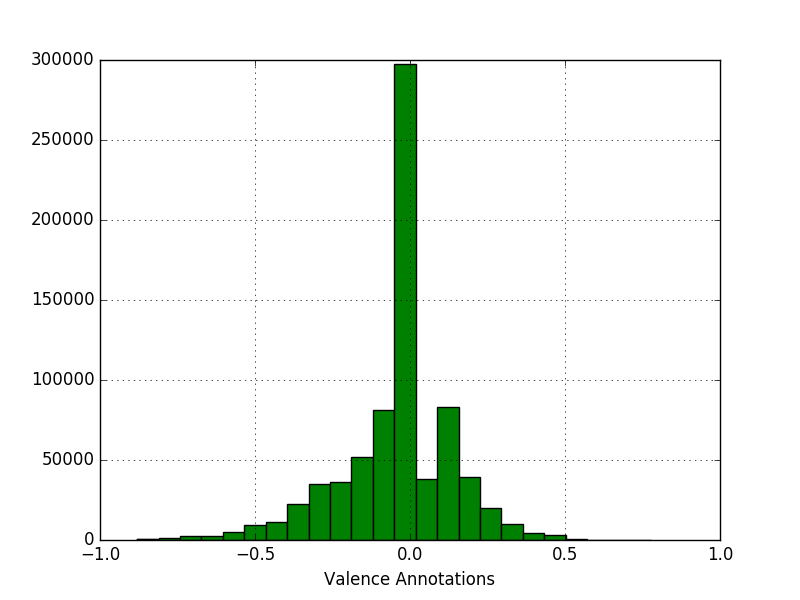}}
    \subfigure[Arousal Annotations]
    {\includegraphics[height=4cm,width=7cm]{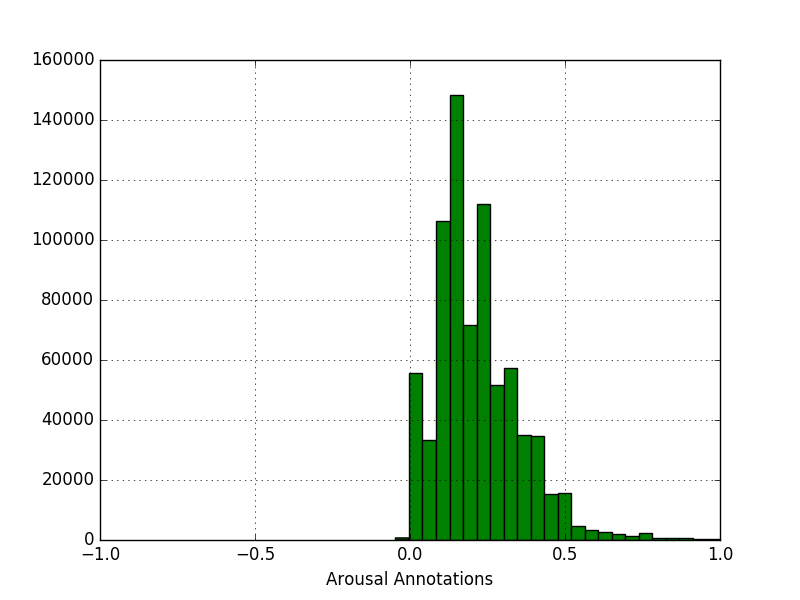}}
    \caption{Histogram of Annotations}
    \label{annotationHist}
\end{figure}

In Figure \ref{annotationDistribution}, the distribution of valence and arousal annotations for train, validation and test set is demonstrated. It is obvious that the three data sets have similar distributions in valence and arousal. Specifically, the arousal value is biased towards the positive value. While the distribution of valence value is more balanced in positive and negative range, the value mainly concentrated around neutral value. The expressions having extremely positive or negative value are very rare. From Figure \ref{annotationHist}, the distribution of valence and arousal of all frames are shown which can give more clear insight into the data distribution characteristics.

The gender distribution is shown in Table \ref{tab:gender} from which we can see the ratio of male to female in each data set is quite close to 1:1.

\begin{table}[H]
    \centering
    \begin{tabular}{|c|c|c|c|c|}
	\hline
	Database & Train & Validate & Test & Total \\
	\hline
    No of Male & 51 & 14 & 14 & 79 \\
    \hline
    No of Female & 49 & 15 & 16 & 80 \\
	\hline
	Total & 100 & 29 & 30 & 159 \\
	\hline
    \end{tabular}
    \caption{Number of Male/Female in Three Data Set.}
    \label{tab:gender}
\end{table}

In Table \ref{tab:attributes}, some attributes of database are listed.
\begin{table}[H]
    \centering
    \begin{tabular}{|c|c|}
	\hline
	Attributes & Description \\
	\hline
    Format & MP4 \\
    \hline
    Length of videos & 0.10-15.04 min \\
	\hline
	Total no of videos & 159 \\
	\hline
    \end{tabular}
    \caption{Attributes of the Database.}
    \label{tab:attributes}
\end{table}

\section{Summary}
\label{databasesummary}
The database built for extending Aff-Wild database were sourced from YouTube where videos usually have spontaneous facial expressions in the wild condition. After downloading with the best quality, videos were all converted to MP4 format so that they can be recognized by the annotator program. Then the videos were trimmed and converted into 30 FPS in order to be suitable for training purpose. During the annotation procedure, videos were watched first, then analyzed in terms of valence and arousal associated with their content. Later on, the videos were annotated while being displayed. Since facial expressions are the central analysis, human faces were cropped from every frame of videos using FFLD2Detector. The resulting cropped object is selected among all detected objects of one frame by comparing the similarity of RGB histogram of all objects detected and the real face. Till then, the annotations for every video and the faces for every frame were all collected. By executing the matching method, nearest neighbour algorithm, the database was finally built. Resulting database contains 159 annotations files representing 159 videos. Each file has valence and arousal annotation for each frame. In order to prepare the train, validate and test set for training deep neural network, the total data set was partitioned into 100, 29 and 30 videos, having 527056, 94223 and 135145 frames respectively. At this stage, the database is built and ready for training the neural network.

\chapter{Design and Alternatives}
\label{designChapter}




An end-to-end deep neural network is designed to predict on facial expression with respect to valence and arousal for every frame in video database collected in Chapter \ref{databaseChapter}. Inspired by the best possible deep neural network discussed in background \ref{AffWildChallenge}, joint CNN and RNN architecture design is adopted to construct facial expression recognition model. Through experiments on different combinations of CNNs and RNNs intoduced in later chapter \ref{experimentChapter}, the best performance architecture can be found. There exists a lot of state-of-the-art CNNs with pre-trained model and RNNs. To give reasonable comparisons, the experiment will first try different RNNs with VGGFace based CNN part to find the best performed RNN, which can be called as Best RNN. Then with this Best RNN fixed, different CNNs can be evaluated. Finally, the best CNN and RNN combination will become the resulting model to fulfill the emotion recognition task.\newline

This Chapter will first introduces the overview of the whole system, namely the joint CNN and RNN design in Section \ref{CRNNdedign}. Then the alternative CNNs and RNNs which can be used to constitute the whole network will be illustrated in Section \ref{alteCNNs} and \ref{alteRNNs}. For the output of the whole system, the FC layer design is explained in Section \ref{fcdegin}.

The contribution of this Chapter is giving the design of the joint CNN and RNN architecture and discussing alternative state-of-the-art CNNs and RNNs used in experiments.

\section{CNN-RNN Design}
\label{CRNNdedign}
\begin{figure}[H]
\centering
\includegraphics[height=10cm,width=8cm]{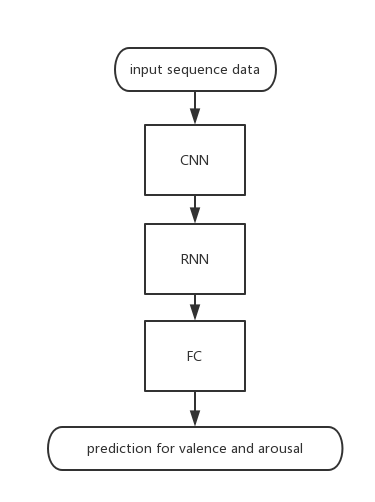}
\caption{Neural Network designed to predict valence and arousal for every frame in video data.}
\label{CRNN}
\end{figure}

The neural network used to recognize emotions in extended database built in Chapter \ref{databaseChapter} is an end-to-end architecture which is composed of CNN, RNN and FC layers subsequently \citep{kollias4,kollias7,kollias14,kollias15} as shown in Figure \ref{CRNN}.
The input sequence of data (video data) has shape of ${Sequence Length \times 96 \times 96 \times 3}$ where the last dimension 3 means there are three channel, RGB, for every frame and the spatial size of one image is ${96 \times 96}$. Then the input data will be first fed into CNN part to extract the visual features, before entering the RNN part which is used to model the dynamics in time dimension. At the end the network, the FC layer is used to give prediction of the valence and arousal for every frame in the sequence of data. So the prediction ${p}$ is of shape ${Sequence Length \times 2}$. For sequence of frames ${F \in {\rm I\!R}^{Sequence Length \times 96\times96\times3}}$, the prediction ${P \in [-1, 1]^{Sequence Length \times 2}}$ will be produced by the whole end-to-end neural network.

\section{Alternative CNNs}
\label{alteCNNs}
For CNN part in Figure \ref{CRNN}, Table \ref{tab:CNNmodels} shows the potential architectures together with corresponding pre-trained models which will be evaluated in later experimental Chapter \ref{experimentChapter}. In the following sub sections, these architectures and models are discussed.

\begin{table}[H]
\centering
\begin{tabular}{|c|c|}
\hline
Network         & Database Pre-trained on \\ \hline
VGGFace network & VGGFace         \\ \hline
ResNet-50       & VGGFace2                \\ \hline
ResNet-50       & ImageNet            \\ \hline
DenseNet-121    & ImageNet            \\ \hline
DenseNet169     & ImageNet            \\ \hline
\end{tabular}
\caption{CNN architectures with corresponding pre-trained model evaluated in experiments.}
\label{tab:CNNmodels}
\end{table}

\subsection{VGGFaceNet}
\label{VGGFaceDesign}
The VGGFace network \citep{Parkhi2015DeepRecognition} has similar configurations with VGG16 as described in \ref{VGG16section}. The CNN part configuration based on VGGFace network is shown in Table \ref{VGG16CNN}. 

\begin{table}[H]
\centering
\begin{tabular}{|c|c|}
\hline
Layer Name & Configuration \\ \hline
conv1\_x & {[}3 x 3, 64{]} x 2 \\ \hline
 & maxpool \\ \hline
conv2\_x & {[}3 x 3, 128{]} x2 \\ \hline
 & maxpool \\ \hline
conv3\_x & {[}3 x 3, 256{]} x3 \\ \hline
 & maxpool \\ \hline
conv4\_x & {[}3 x 3, 512{]} x3 \\ \hline
 & maxpool \\ \hline
conv5\_x & {[}3 x 3, 512{]} x3 \\ \hline
 & maxpool \\ \hline
\end{tabular}
\caption{CNN part configuration based on VGGFace network design. The parameters in this table are denoted as [kernel size, number of filters] times number of replication. All the conv layers in VGGFace have stride size ${1 \times 1}$ and all max pooling layers have kernel size ${2 \times 2}$ and stride size ${2 \times 2}$.}
\label{VGG16CNN}
\end{table}

\subsubsection{VGGFace Pre-trained Model}
The VGGFace network is well trained for face recognition task on VGGFace database created in \cite{Parkhi2015DeepRecognition} which contains  2.6 Million face images of 2,622 unique subjects. The resulting model was evaluated on LFW\citep{Huang2007LabeledEnvironments} benchmark (for automatic face verification) and
YTF\citep{YTFwolf2011face} benchmark (for face verification in video) and achieved comparable performance with the state-of-the-art methods on these two benchmarks. The main characteristics of VGGFace model is that it is trained on relatively smaller data set with much simpler network design while has comparable performance compared to other methods which were trained on larger data set or designed to have complex network. What is more, this VGGFace network and pre-trained model also achieved best performance in AFF-WILD challenge as experiemnted in \cite{kollias2}. Therefore, this VGGFace pre-trained model will be experimented on the extended database in Chapter\ref{databaseChapter}.

\subsection{ResNet50}
\label{ResNet50Design}
As described in background section \ref{ResNet50section}, ResNet50 has a stack of bottleneck architectures. Here for connecting with RNNs, the CNN part based on ResNet50 has configuration detailed in Table \ref{tab:resnet50CNN}.

\begin{table}[H]
    \centering
    \begin{tabular}{|c|c|}
	\hline
	Layer name & Configuration  \\
	\hline
	conv1
	&
    ${
    7 \times 7, 64, stride 2
    }$ \\
    \hline
	conv2\_x
	&
    ${3 \times 3}$ max pool , stride 2
    \\
	&
    ${
    \left[\begin{array}
    {cc} 1 \times 1, & 64 \\  3 \times 3, & 64 \\ 1 \times 1, & 256
    \end{array}\right] \times 3
    }$\\
    \hline
    conv3\_x
	&
    ${
    \left[\begin{array}
    {cc} 1 \times 1, & 128 \\  3 \times 3, & 128 \\ 1 \times 1, & 512
    \end{array}\right] \times 4
    }$\\
    \hline
    conv4\_x
	&
    ${
    \left[\begin{array}
    {cc} 1 \times 1, & 256 \\  3 \times 3, & 256 \\ 1 \times 1, & 1024
    \end{array}\right] \times 6
    }$\\
    \hline
    conv5\_x
	&
    ${
    \left[\begin{array}
    {cc} 1 \times 1, & 512 \\  3 \times 3, & 512 \\ 1 \times 1, & 2048
    \end{array}\right] \times 3
    }$\\
    \hline
	&
    average pool
    \\
    \hline
    \end{tabular}
    \caption{ResNet50 based architecture configuration used in joint CNN-RNN structure. ${n \times n}$ represents the conv filter size, after which is the number of filters. The number after array structure means how many times this structure will be replicated. }
    \label{tab:resnet50CNN}
\end{table}

\subsubsection{ImageNet Pre-trained Model}
ImageNet Large Scale Visual Recognition Challenge (ILSVRC) \citep{ILSVRC15} evaluates algorithms for object localization/detection from images/videos at scale. The ResNet50 based CNN model later used in experiments restores the model of ResNet-50 pre-trained on ILSVRC2012\citep{deng2012imagenet} which has performance as listed in Table \ref{tab:resnetImageNet}.

\begin{table}[H]
\centering
\begin{tabular}{|c|c|c|}
\hline
Network             & Top-1 & Top-5 \\ \hline
ResNet-50   & 75.2\% & 92.2\% \\ \hline
\end{tabular}
\caption{ The top 1 and top 5 accuracy rates on the ImageNet test set.}
\label{tab:resnetImageNet}
\end{table}

\subsubsection{VGGFace2 Pre-trained Model}
\label{subsubsec:vf2}
The VGGFace2 database is created in \cite{VGGFace2Cao18}. It is a dataset for recognizing faces across pose and age containing 3.31 million images collected for 9131 subjects. Four pre-trained models including
\begin{enumerate}
    \item ResNet-50 trained on VGGFace2 training set from scratch
    \item ResNet-50 model fine-tuned on VGGFace2 training set based on a pretrained model on Ms-Celeb-1M \citep{Ms-celeb-1m} dataset
    \item SE-ResNet-50 model \citep{hu2018senet} trained on VGGFace2 training set from scratch
    \item SE-ResNet-50 model \citep{hu2018senet} fine-tuned on VGGFace2 training set based on a pretrained model on Ms-Celeb-1M \citep{Ms-celeb-1m} dataset
\end{enumerate}
were evaluated on IJB-A \citep{IJB-A} and IJB-B \citep{IJB-B} face recognition benchmarks and all outperformed the existing state-of-the-art methods a lot. These four pre-trained model were all considered to be experimented later. However, the SE-ResNet-50 model was not converted successfully. So only the ResNet-50 based pre-trained models are experimented. 

\subsubsection{VGGFace VS VGGFace2}
Through the experiments in \cite{VGGFace2Cao18}, the ResNet-50 network trained on VGGFace2 has better performance than the one trained on VGGFace in face identification and probing across age and pose problem. All evidence shows that VGGFace2 has larger data variation (especially in terms of pose and age), better data quality and less label noise \cite{raftopoulos2018beneficial} than VGGFace. So in theory, the model pre-trained on VGGFace2 should have better performance in later experiments.

\subsection{DenseNet}
\label{DenseNetDesign}
The DenseNet architecture is illustrated in background section \ref{DenseNetsection}. There are two different configurations DenseNet evaluated later.

\begin{table}[H]
\centering
\begin{tabular}{|c|c|c|}
\hline
Layer Name                        & DenseNet-121            & DenseNet-169            \\ \hline
Conv layer                        & \multicolumn{2}{c|}{7 x 7 conv, stride 2}         \\ \hline
Pool                              & \multicolumn{2}{c|}{3 x 3 max pool, stride 2}     \\ \hline
Dense Block 1                     &    
    ${
    \left[\begin{array}
    {cc} 1 \times 1 & conv \\  3 \times 3 & conv
    \end{array}\right] \times 6
    }$                     
    &                         
    ${
    \left[\begin{array}
    {cc} 1 \times 1 & conv \\  3 \times 3 & conv
    \end{array}\right] \times 6
    }$ 
    \\ \hline
\multirow{2}{*}{Transition Layer} & \multicolumn{2}{c|}{1 x 1 conv}                   \\ \cline{2-3} 
                                  & \multicolumn{2}{c|}{2 x 2 average pool, stride 2} \\ \hline
Dense Block 2                     
&     
${
    \left[\begin{array}
    {cc} 1 \times 1 & conv \\  3 \times 3 & conv
    \end{array}\right] \times 12
    }$ 
&  
${
    \left[\begin{array}
    {cc} 1 \times 1 & conv \\  3 \times 3 & conv
    \end{array}\right] \times 12
    }$ 
\\ \hline
\multirow{2}{*}{Transition Layer} & \multicolumn{2}{c|}{1 x 1 conv}                   \\ \cline{2-3} 
                                  & \multicolumn{2}{c|}{2 x 2 average pool, stride 2} \\ \hline
Dense Block 3                     
&      
${
    \left[\begin{array}
    {cc} 1 \times 1 & conv \\  3 \times 3 & conv
    \end{array}\right] \times 24
    }$ 
&
${
    \left[\begin{array}
    {cc} 1 \times 1 & conv \\  3 \times 3 & conv
    \end{array}\right] \times 32
    }$ 
\\ \hline
\multirow{2}{*}{Transition Layer} & \multicolumn{2}{c|}{1 x 1 conv}                   \\ \cline{2-3} 
                                  & \multicolumn{2}{c|}{2 x 2 average pool, stride 2} \\ \hline
Dense Block 4                     
&
${
    \left[\begin{array}
    {cc} 1 \times 1 & conv \\  3 \times 3 & conv
    \end{array}\right] \times 16
    }$ 
& 
${
    \left[\begin{array}
    {cc} 1 \times 1 & conv \\  3 \times 3 & conv
    \end{array}\right] \times 32
    }$ 
\\ \hline
Pool                              & \multicolumn{2}{c|}{ global average pool}     \\ \hline
\end{tabular}
\caption{CNN part configuration based on DenseNet design. For DenseNet121 and DenseNet169, the growth rate for all dense blocks is ${k = 32}$ and each ${conv}$ layer in the table has the sequence BN-ReLU-Conv. \citep{huang2017densely}}
\label{tab:DenseNetCNN}
\end{table}

\subsubsection{ImageNet Pre-trained Model}
ImageNet Large Scale Visual Recognition Challenge (ILSVRC) \citep{ILSVRC15} evaluates algorithms for object localization/detection from images/videos at scale. The pre-trained models on ImageNet used later in experiments are listed in Table \ref{tab:denseImageNet}. 

\begin{table}[H]
\centering
\begin{tabular}{|c|c|c|}
\hline
Network             & Top-1 & Top-5 \\ \hline
DenseNet 121 (k=32) & 74.91\% & 92.19\% \\ \hline
DenseNet 169 (k=32) & 76.09\% & 93.14\% \\ \hline
\end{tabular}
\caption{ The top 1 and top 5 accuracy rates by using single center crop (crop size: 224x224, image size: 256xN) stated in \cite{denseRepo}. }
\label{tab:denseImageNet}
\end{table}

\section{Alternative RNNs}
\label{alteRNNs}

The whole RNN block in Figure \ref{CRNN} is configured as the details shown in Table \ref{tab:rnnblock}:
\begin{table}[H]
\centering
\begin{tabular}{|c|c|c|}
\hline
\multirow{2}{*}{RNN block} & RNN layer 1 & 128 hidden units \\ \cline{2-3} 
                           & RNN layer 2 & 128 hidden units \\ \hline
\end{tabular}
\caption{RNN block design}
\label{tab:rnnblock}
\end{table}

where the number of RNN layers is 2 and hidden units is 128. These two hyper-parameters are adopted based on experiment results mentioned in \cite{kollias7}. And the specific RNN unit evaluated in later experimental Chapter \ref{experimentChapter} can be alternatives among LSTM, GRU and Independently RNN. In addition, the Attention mechanism will also be considered. All of these will be introduced in following subsections. 

\subsection{LSTM}
The LSTM used in experiments later also allows the Peephole connection \citep{Beaufays2014LongHas}. With Peephole Connection, the current time stamp gate is allowed to "see" the cell state. Compared with details in Section \ref{LSTMsection}, the changes brought by allowing Peephole connection are mainly on input gate, forget gate and output gate, where it is necessary to add a variable indicating the state of the cell as shown below:

\begin{equation}
\label{equForline}
\begin{aligned}
&i_{t}= \sigma(W_{ix}x_t + U_{i}h_{t−1} + W_{ic}c_{t-1} +b_{i})\\
&f_{t}= \sigma(W_{fx}x_t + U_{f}h_{t−1} + W_{fc}c_{t-1} + b_{f})\\
&o_{t}= \sigma(W_{ox}x_t + U_{o}h_{t−1} + W_{oc}c_t + b_{o})
\end{aligned}
\end{equation}

where ${c_{t-1}}$ is the cell state for previous time stamp and ${c_{t}}$ is the current time stamp cell state. The newly added trainable parameters are: ${W_{ic}, W_{fc}, W_{oc}}$.

\subsection{GRU}
The GRU will be used in experiments is the version stated in section \ref{GRUsection}.

\subsection{IndRNN}
Although the gradient vanishing and exploding problem can be solved somewhat by LSTM and GRU design, the gradient decay problem still exist over layers due to the use of ${tanh}$ and ${sigmoid}$ activation functions. What is more, the gradient vanishing problem still exists in much longer sequence. In Independently RNN (IndRNN) design \citep{Li2018IndependentlyRNN}, the ReLU activation rather than ${sigmoid}$ or ${tanh}$ is used so as to make it easier to stack multiple recurrent layers. And independently RNN is designed to make longer RNN which can be described as:

\begin{equation}
\label{equForline}
\begin{aligned}
&h_{t}= \sigma(W x_t + u \bigodot h_{t−1} + b)\\
\end{aligned}
\end{equation}

where ${u}$ is a vector not a matrix and ${\bigodot}$ stands for Hadamard product. In this case, each neuron in hidden state only connect to itself in previous time stamp so that the function of every neuron in hidden state can be interpreted and visualized easily. The association between different neurons in hidden state is accomplished by stacking multiple layers where every neuron in next layer accepts the output from all neurons in previous layer. Last but not least, IndRNN regulate the recurrent weights of hidden neurons so that the gradient vanishing and exploding problem can be prevented. By using IndRNN, deeper and longer RNN can be constructed and work well. 

\subsection{Attention}
\label{sec:attention}
The Attention mechanism used in experiments through the API
\begin{lstlisting}[language=Python]
tf.contrib.rnn.AttentionCellWrapper
\end{lstlisting}

which is implemented in TensorFlow platform \citep{tensorflow2015-whitepaper}. The core idea for this attention mechanism is based on \cite{Bahdanau2014NeuralTranslate}. Basically, with the attention mechanism, every time the RNN generate the output for the current time step, it concentrates on the last ${n}$ time steps where the ${n}$ representing the window size of the attention mechanism. The key operation can be described as below:

\begin{equation}
\label{equForline}
\begin{aligned}
&s^{t}_{i}= v^{T} \tanh(W_{1}out_i + W_{2}c_{t})\\ 
&p^{t}_{i}=softmax(s^{t}_{i})\\
&atten_{t}=\sum_{i=t-n}^{t-1}p^{t}_{i}out_i\\
\end{aligned}
\end{equation}

where ${v^{T}}$, ${W_{1}}$ and ${W_{2}}$ are trainable parameters. ${out_i}$ represent one of the output from previous ${n}$ steps. ${c_{t}}$ stands for cell state for current time step. ${s^{t}}$ is a vector of length ${n}$ with one value for one step of total ${n}$ steps. After applying softmax function on  ${s^{t}}$, a probability distribution  ${p^{t}}$ is obtained which represents how much attention should receive for every time step in last ${n}$ steps. Finaly, the attention state ${atten_{t}}$ is the weighted sum of the output from previous ${n}$ steps with respect to the distribution  ${p^{t}}$. And the ${atten_{t}}$ state will also be used to constitute the input and output by concatenated with raw input and output then applied linear transformation.\newline

This attention mechanism is applied to the whole two layers RNN block in the experiments. 



\section{Fully Connected Layer}
\label{fcdegin}
As the last layer for the whole neural network used to recognize the facial expressions, fully connected layer here is designed to have two neurons with one for predicting valence and another for arousal as shown in Table \ref{tab:fcblock}.

\begin{table}[H]
\centering
\begin{tabular}{|c|c|c|}
\hline
FC block & Fully Connected Layer & 2 neurons \\ \hline
\end{tabular}
\caption{FC block has one fully connected layer with 2 neurons. }
\label{tab:fcblock}
\end{table}

\section{Summary}
\label{designsummary}
The neural network built for recognizing facial expressions in database created in Chapter \ref{databaseChapter} consists of CNN, RNN and FC block successively. The CNN block can be designed based on VGGFace, ResNet and DenseNet netoworks with pre-trained models. And RNN block is constructed to have 2 layers of RNN which can be LSTM allowing Peephole connection, GRU, IndRNN. Attention mechanism is also considered to be applied. With these state-of-the-art architectures, the best combination of CNN block and RNN block will be found through the later experiments.

\chapter{Implementation}
\label{implementationChapter}
\begin{figure}[H]
\centering
\includegraphics[width=10cm,height=11cm]{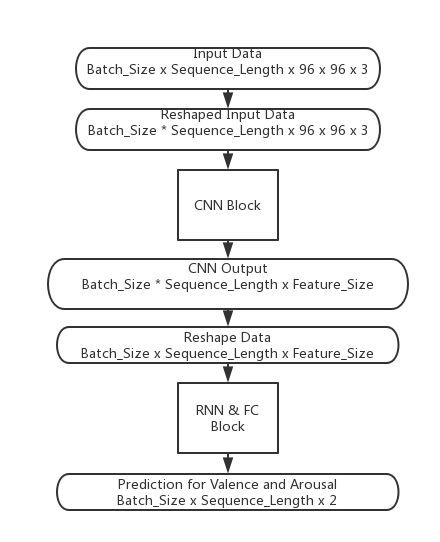}
\caption{Data Flow Shape during the procedure when pass along the network.}
\label{fig:datashape}
\end{figure}

This chapter gives the details of how the whole architecture is built and the environment of the experiments. \newline
Based on the design stated in Chapter \ref{designChapter}, the implementation of input data flow will be demonstrated in Section \ref{sec:data}, then the CNN and RNN implementation will be introduced in Section \ref{sec:CNNimp} and Section \ref{sec:RNNimp}. The definition of loss function will be illustrated in Section \ref{sec:lossfunction}. In Section \ref{sec:trainimp} and \ref{sec:evaimp}, the training and evaluating program will tell how the experiment running with these two programs. The environment details can be found in Section \ref{sec:envimp}. The data flow shape is illustrated in Figure \ref{fig:datashape}.

The contribution of this Chapter is setting up the basis of the experiments so that in later process the various experiments can be executed and the best architecture and best model can be found.

\section{Input Data}
\label{sec:data}
The database creation results are the generated 159 annotation files for 159 videos and cropped human faces from 159 videos stored in folders as described in Chapter \ref{databaseChapter}. For the simplicity of feeding data into networks later, the whole data sets including train, validate and test sets are all written into TFRecord files which will be explained in sub section \ref{sec:tfrecords}. Then in the DataLoader Class, the TFRecords files are imported into computational graph with the Dataset API in TensorFlow \citep{tensorflow2015-whitepaper} and are transformed into the suitable format so that the neural network can be trained properly on the data set which is displayed in section \ref{sec:dataloader}. After this whole procedure, the resulting data ready to be fed into the CNN block is detailed in Table \ref{tab:readyData}.
\begin{table}[h]
\centering
\begin{tabular}{|c|c|}
\hline
Attribute    & Content                 \\ \hline
Frame Image  & ${F \in [-1.0, 1.0]^{BatchSize \times SequenceLength \times 96\times96\times3}}$ \\ \hline
Label      & ${x \in [-1.0, 1.0]^{BatchSize \times SequenceLength \times 2}}$ \\ \hline
\end{tabular}
\caption{Details of processed data which is ready to be fed into the neural networks.}
\label{tab:readyData}
\end{table}

\subsection{Store in TFRecords}
\label{sec:tfrecords}
\begin{figure}[H]
\centering
\caption{Annotation files and cropped face images are all merged into TFRecords files.}
\label{fig:tfrecords}
\includegraphics[width=16cm,height=11cm]{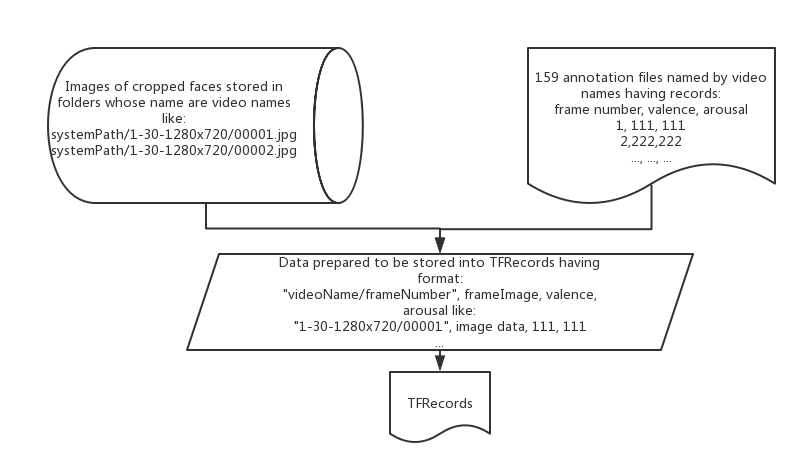}
\end{figure}

At this step, the cropped human faces and corresponding labels are all written into TFRecord files together. The procedure is shown in Figure \ref{fig:tfrecords}. And the data format for TFRecords can be seen in Table \ref{tab:tfdata}.
The reasons of choosing TFRecords are stated as following. 
\begin{enumerate}
    \item With TFRecords, data with different types can be stored together. It is convenient to read the frame data and respective labels from one single file rather than read the images and labels separately.  
    \item It is fast to read data from TFRecords since the data stored in TFRecords is of binary storage format.
    \item With TFRecords integrated with TensorFlow Dataset API, data can be read in small batches. Because the total storage of cropped faces is of 14.8 GB which can not be stored in memory totally.
\end{enumerate}

Now, the concrete details are explained. To write the data into TFRecords, the data should be processed first. The code below is the procedure of writing frame images of one video and corresponding valence and arousal values into one TFRecord file. 

\begin{scriptsize}
\estiloPython
\begin{lstlisting}[language=Python]
def _bytes_feature(value):
    return tf.train.Feature(bytes_list=tf.train.BytesList(value=[value]))
def _int64_feature(value):
    return tf.train.Feature(int64_list=tf.train.Int64List(value=[value]))
writer = tf.python_io.TFRecordWriter(tfRecordFilePath)
for record in getAnnotation(annotationFileForOneVideo):
    frameNumber = record[0]
    img = load_image(framesPath+frameNumber+'.jpg'),
    feature = {
        `frameNo':_bytes_feature(tf.compat.as_bytes(videoName+'/'+frameNumber)),
        `frame':_bytes_feature(tf.compat.as_bytes(img.tostring()),
        `valence':_int64_feature(record[1]),
        `arousal':_int64_feature(record[2])
    }
    example = tf.train.Example(features=tf.train.Features(feature=feature))
    writer.write(example.SerializeToString())
writer.close()
\end{lstlisting}
\end{scriptsize}
Data records in annotation file are iterated and processed one by one. The annotation file can be found by video name. Every record has frame number, valence and arousal value as stated in Table \ref{tab:mergeAnnotation}. Then the cropped face image can be imported by the method load\_image() with the path constituted by frame number and video name like : systemPath/videoName/frameNumber and the frame is read by OpenCV library \citep{opencv_library} and resized into ${96 \times 96}$. So every record in annotation file is transformed to data record having four attributes as shown in Table \ref{tab:tfdata}. Then every attribute is wrapped with tf.train.Feature so that TensorFlow can recognize them. Next, all attributes wrapped are organized by a dictionary structure `feature' so that the data stored in format of bytes can be read back later in the same structure. With this `feature' structure, data are wrapped again by tf.train.Example which is a protocol buffer used to serialize the data. Finally, the instance of tf.train.Example is written into TFRecord file by a writer created by tf.python\_io.TFRecordWriter with TFRecords path. The cropped faces of every video are stored into one TFRecord file in the same way. \newline
So after this stage, the database is converted into 159 TFRecord files representing 159 videos data. The attributes of each record are shown in Table \ref{tab:tfdata}.

\begin{table}[h]
\centering
\begin{tabular}{|c|c|}
\hline
Attribute    & Content                 \\ \hline
Frame Number & `videoName/frameNumber' \\ \hline
Frame Image  & ${F \in [0.0, 255.0]^{96\times96\times3}}$                        \\ \hline
Valence      & ${x \in [-1000, 1000]}$ and ${x \in}$ $\mathbb{Z}$   \\ \hline
Arousal      & ${x \in [-1000, 1000]}$ and ${x \in}$ $\mathbb{Z}$   \\ \hline
\end{tabular}
\caption{Data prepared for storing in TFRecords.}
\label{tab:tfdata}
\end{table}

\subsection{DataLoader Class}
\label{sec:dataloader}
\begin{figure}[H]
\centering
\includegraphics[width=12cm,height=21cm]{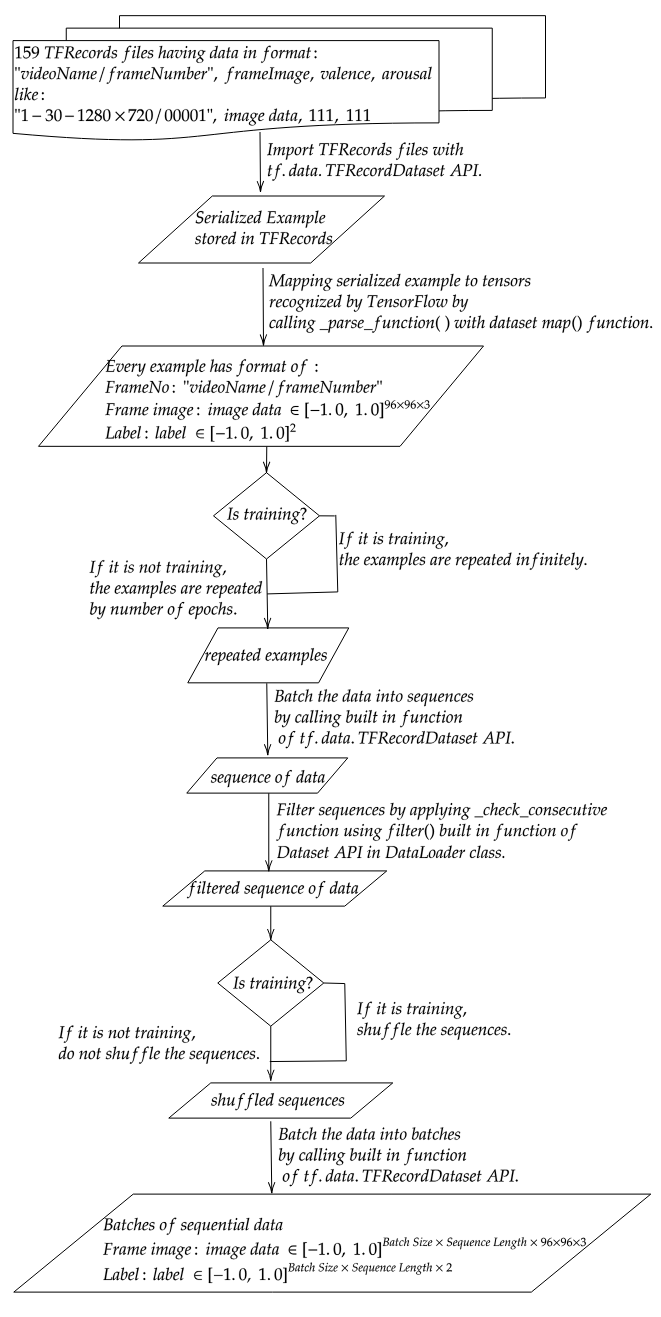}
\caption{Data process procedure implemented in DataLoader Class.}
\label{fig:DataLoader}
\end{figure}
An overview of the whole data process procedure is shown in Figure \ref{fig:DataLoader}. With the created 159 TFRecord files, batches of sequence of data used to fed into the neural network is generated by DataLoader class. In one data batch, the sequences of frames should come from different videos while the frames in one sequence should be from the same video. What is more, the frames in one sequence are ideal to be consecutive rather than have large gap between each other in terms of time stamps. Based on these requirements, the DataLoader Class in python is implemented as below:
\begin{scriptsize}
\estiloPython
\begin{lstlisting}[language=Python]
class DataLoader(object):
    def __init__(self, seq_length, batch_size, epochs):
        self.num_epochs = epochs
        self.batch_size = batch_size
        self.seq_length = seq_length
        self.buffer_size = 10 * self.batch_size * self.seq_length
        self.threshold = 15 * self.seq_length
    def getTFRecordsFiles(self, tfRecordsPath):
    def getTFRecordsPathList(self, tfRecordsPath):
    def _check_consecutive(self, frameNo, image, label):
    def _parse_function(self, serialized_example):
    def scale(self, image):
    def load_data(self, tfRecordsPath, is_training):
        dataset = tf.data.TFRecordDataset(self.getTFRecordsPathList(tfRecordsPath))
        ...
        dataset.make_one_shot_iterator()
        return iterator
\end{lstlisting}
\end{scriptsize}

Now the methods in DataLoader is illustrated:
\begin{itemize}
    \item \_\_init\_\_(self, seq\_length, batch\_size, epochs): By accepting the sequnce length, batch size and number of epochs, the DataLoader instance is created by initializing these attrubutes. Buffer size is also initialized used to set how much buffer used to shuffle the data. The threshold attribute is used to determine the maximum value allowed of the first and last frame number in one sequence.
    \item getTFRecordsFiles(self, tfRecordsPath): Given parameter tfRecordsPath which stands for the path of the folder which contains the expected TFRecord files, this method return a list of tfrecord files' name.
    \item getTFRecordsPathList(self, tfRecordsPath): Given the parameter tfRecordsPath which refers to the path of the folder containing the expected TFRecord files, this method returns the list of full path for tfrecord files.
    \item \_parse\_function(self, serialized\_example): Mapping the serialized example stored in TFRecords to tensors of frame number with type string, image data of type float and label data of type float. In addition, the valence and arousal data are converted in range of [-1.0, 1.0] by divided 1000.0 and merged together into one variable called `label'. And the image data is also scaled so that each element in image is in the range of [-1.0, 1.0] by subtracting 128.0 from them and dividing them by 128.0.
    \item \_check\_consecutive(self, frameNo, image, label): This function is designed to filter one sequence of data so that the frames in one sequence is close to each other regarding the time steps. Returning True means keeping this sequence, otherwise discard this sequence. This is implemented by checking the frameNo value which is a string of content `videoName/frameNo'. First, the first and last frame's video name are checked. If the the video names are the same then continue checking otherwise return false. Secondly, check the distance between the first frame number and the last one, if the distance is greater than attribute self.threshold, then return False, otherwise return True.
    \item load\_data(self, tfRecordsPath, is\_training): This method is the key of DataLoader class. With the TFRecords file path, the Dataset instance is created by tf.data.TFRecordDataset() API provied by TensorFlow \citep{tensorflow2015-whitepaper} which accepts the list of TFRecords files returned by method getTFRecordsPathList(self, tfRecordsPath) and returns the instance of Dataset. Then every instance of tf.train.Example in TFRecords is mapped by executing \_parse\_function() function invoked by map() function of Dataset instance. Then based on the value of boolean parameter is\_training, if this is training time, then the data will be repeated infinitely, other wise the data will be repeated based on the self.num\_epochs value. Then the \_check\_consecutive method is invoked to filter the sequence after the data is batched with sequence length. Finally, the data is batched into batch size after random shuffle process on sequences. At the end, the Iterator instance is returned by Dataset instance. With the Iterator instance, the data with the specified format can be read by invoking its get\_next() method.
\end{itemize}

The code below shows how to use the DataLoader class. To get batches of sequences, the DataLoader instance should be created by specifying the expected batch size, sequence length and number of epochs at first. Then by invoking the load\_data() method with the TFRecords Path and a boolean variable indicating whether or not it is training, an iterator created by Dataset API will be returned to provide access to generated data. In this way, the data fed into network can be gotten by invoking get\_next() method of Iterator Class instance.

\begin{scriptsize}
\estiloPython
\begin{lstlisting}[language=Python]
data_loader = DataLoader(SEQUENCE_LENGTH, BATCH_SIZE, NUM_EPOCHS)
iterator = data_loader.load_data(tfrecordsPath, True)
frameNo, image, label = iterator.get_next()
\end{lstlisting}
\end{scriptsize}

The resulting image data generated at this stage is of shape\newline ${batch\_size \times  sequence\_length \times 96 \times 96 \times 3}$.\newline The label data is of shape ${batch\_size \times sequence\_length \times 2}$. Now the data is ready for fed into the neural network.

\section{Neural Network}
With the data obtained from the Iterator generated by Dataset, the predictions for valence and arousal can be produced by applying neural network (\cite{wallace2003intelligent}) consisting of CNN and RNN as described in Chapter \ref{designChapter}. Now the implementation of CNN blocks and RNN blocks are introduced in the following sub sections.

\subsection{CNN Implementation}
\label{sec:CNNimp}
The implementation of CNN block is used to extract the visual features from the input data. After accepting the input cropped human face images, the output from CNN is the feature extracted for every image. The image data generated from the iterator returned by DataLoader has the shape of ${batch\_size \times  sequence\_length \times 96 \times 96 \times 3}$. Before fed into the CNN block, the data should be reshaped into shape of\newline ${batch\_size * sequence\_length \times 96 \times 96 \times 3}$. This reshape operation is fulfilled by tf.reshape() function which accepts the tensor to be reshaped and the target shape and returns the reshaped tensor.

\begin{scriptsize}
\estiloPython
\begin{lstlisting}[language=Python]
frameNo, image, label = iterator.get_next()
image_batch = tf.reshape(images, (BATCH_SIZE *SEQUENCE_LENGTH, 96, 96, 3))
\end{lstlisting}
\end{scriptsize}

So the data which is fed into the CNN block is contained in image\_batch variable having shape of ${batch\_size * sequence\_length \times 96 \times 96 \times 3}$.

\subsubsection{VGGFace}
The implementation for VGGFace network comes from \cite{kollias2}, together with the VGGFace\citep{Parkhi2015DeepRecognition} pre-trained model.

\begin{scriptsize}
\estiloPython
\begin{lstlisting}[language=Python]
VGGFace_network = VGGFace(SEQUENCE_LENGTH * BATCH_SIZE)
VGGFace_network.setup(image_batch, trainable=False)
cnn_output = VGGFace_network.get_face_fc0()
\end{lstlisting}
\end{scriptsize}

With the code above, an instance VGGFace\_network of VGGFace class is created by the constructor which requires the batch size for initializing the class. Here for CNN block, the batch size is SEQUENCE\_LENGTH * BATCH\_SIZE. Then the reshaped input image data, image\_batch, is fed into the setup function of VGGFace\_network together with a parameter trainable which determines whether or not the variables in the VGGFace network computational graph are trainable. In the implementation of setup function, the whole VGGFace network graph in TensorFlow is built. Finally, the get\_face\_fc0 function returns the output from the last conv block in the VGGFace network and assigns the output to cnn\_output variable. 

\subsubsection{ResNet50 with ImageNet model}
The code used to construct ResNet 50 with ImageNet pre-trained model is cloned from GitHub repository of Tensorflow-slim image  classification model library \citep{ResNet50imagenet}. 

\begin{scriptsize}
\estiloPython
\begin{lstlisting}[language=Python]
with slim.arg_scope(resnet_v1.resnet_arg_scope()):
    cnn_output, _ = resnet_v1.resnet_v1_50(
        inputs=image_batch, 
        num_classes=None,
        is_training=True,
        global_pool=True,
    )
\end{lstlisting}
\end{scriptsize}

With the code above, the CNN features which are assigned to cnn\_output can be produced from the API resnet\_v1\_50 provided by resnet\_v1.py source code. Some parameters are required to set up the network. The inputs parameter accepts the input image data. Since the output before logit layer is needed, the parameter num\_classes is set to None. The global\_pool parameter is set to True otherwise this API will perform dense prediction. With boolean parameter is\_training, the training mode for batch normalization layer is specified. Last but not least, other arguments are configured with resnet\_v1.resnet\_arg\_scope() function.

\subsubsection{ResNet50 with VGGFace2 model}
There is another ResNet50 based CNN block which is pre-trained on VGGFace2 database \citep{VGGFace2Cao18}. The code used for this implementation is found on the GitHub repository "VGGFace2 Dataset for Face Recognition" \citep{ResNet50vggface2}. However, the models provided only have Caffe and MatConvNet version. Therefore, the Caffe to TensorFlow conversion tool \citep{CaffeTF} is used to convert caffe model consisting of ".caffemodel" and ".prototxt" files to TensorFlow model consisting of numpy and python files. Then the model stored in numpy file is stored into check point file. For the four pre-trained models stated in section \ref{subsubsec:vf2}, only ResNet50 based models are converted successfully.

\begin{scriptsize}
\estiloPython
\begin{lstlisting}[language=Python]
image_batch = reshape_to_cnn(image)
resnet = ResNet50({'data':image_batch}, trainable=False)
cnn_output = resnet.get_output()
\end{lstlisting}
\end{scriptsize}

With the code above, to start with, the instance resnet of ResNet50 class is created with input image data, image\_batch, with parameter trainable specified. Then the features extracted from images are obtained from get\_output() function of resnet instance.

\subsubsection{DenseNet}
The code used for constructing DenseNet121 and DenseNet169 is cloned from GitHub repository \cite{denseRepo}. And the ImageNet pre-trained models for these two networks are also provided here. The way to use this repository is stated as below:
\begin{scriptsize}
\estiloPython
\begin{lstlisting}[language=Python]
network_fn = nets_factory.get_network_fn(
    name='densenet121',
    num_classes=None,
    data_format='NHWC',
    is_training=True
)
cnn_output, _ = network_fn(image_batch)
\end{lstlisting}
\end{scriptsize}

where the nets\_factory can provide the specific network function by receiving the network name of 'densenet121' or 'densenet169'. It is worth to be noted that setting num\_classs to None aims to get output from global average pooling layer rather than fc layers of DenseNet. The default\_image\_size attribute inside this implementation is also set to 96. So with the network\_fn, the output feature cnn\_output can be extracted from the input data image\_batch. 

\subsection{RNN Implementation}
\label{sec:RNNimp}
The RNN block is implemented to model dynamics in the sequential data. The number of hidden units used is 128 for all RNN units and the number of layers is 2 in all RNN block.

\begin{scriptsize}
\estiloPython
\begin{lstlisting}[language=Python]
feed_in = tf.reshape(cnn_output, [BATCH_SIZE, SEQUENCE_LENGTH, -1])
stacked_rnn = tf.contrib.rnn.MultiRNNCell([list_of_rnn_units])
rnn_outputs, _ = tf.nn.dynamic_rnn(stacked_rnn, feed_in, dtype=tf.float32)

\end{lstlisting}
\end{scriptsize}
The output features, cnn\_output, extracted from CNN block have shape of \newline ${batch\_size * sequence\_length \times feature\_length}$. In order to fit into the RNN block, the data is reshaped into variable feed\_in having shape of \newline ${batch\_size \times sequence\_length \times feature\_length}$. The two layers RNN block is constructed by tf.contrib.rnn.MultiRNNCell() API which accepts a list of RNN units and returns an RNN cell, stacked\_rnn, consisting of multiple simple RNN cells in order. Receiving the stacked\_rnn and feed\_in, the tf.nn.dynamic\_rnn() API builds a recurrent neural network based on the provided stacked\_rnn and performs dynamic unrolling version of RNN based on the length of the input. The return value of tf.nn.dynamic\_rnn() is a tuple of outputs and final states from RNN block for all frames in the input sequence.\newline

The output from the RNN block has shape: ${batch\_size \times sequence\_length \times hidden\_units}$. In the following sections, how the stacked\_rnn is constructed using different RNN units is demonstrated.

\subsubsection{GRU}
The GRU implementation uses the API tf.contrib.rnn.GRUCell(hidden\_units) provided in TensorFlow. 
\begin{scriptsize}
\estiloPython
\begin{lstlisting}[language=Python]
stacked_rnn = tf.contrib.rnn.MultiRNNCell([tf.contrib.rnn.GRUCell(hidden_units), tf.contrib.rnn.GRUCell(hidden_units)]
\end{lstlisting}
\end{scriptsize}
where the hidden\_units is 128.

\subsubsection{LSTM}
The LSTM implementation uses tf.nn.rnn\_cell.LSTMCell() API as shown below:
\begin{scriptsize}
\estiloPython
\begin{lstlisting}[language=Python]
def lstm_cell():
    lstm = tf.nn.rnn_cell.LSTMCell(hidden_units, use_peepholes=True, state_is_tuple=True)
    return lstm
stacked_rnn = tf.nn.rnn_cell.MultiRNNCell([lstm_cell(hidden_units) for _ in range(2)])
\end{lstlisting}
\end{scriptsize}

where the hidden\_units is 128, use\_peepholes is set to True representing allowing peephole connection. 

\subsubsection{Attention Mechanism}
The attention mechanism is implemented by tf.contrib.rnn.AttentionCellWrapper() API. Here, the AttentionCellWrapper() function applys the attention mechanism on the given input stacked\_rnn RNN cell which can be constructed in the way illustrated in previous sections, with the attn\_length parameter which stands for the attention window size specified in the Section \ref{sec:attention}. In the experiment, the attention length of 30 is used due to the video FPS value is 30 which means there are 30 frames in one second so it would not have much changes in one second in terms of the facial expression in the video. The wrapped RNN cell is then used in tf.nn.dynamic\_rnn() function to give the basic RNN unit constructed in the recurrent neural network.
\begin{scriptsize}
\estiloPython
\begin{lstlisting}[language=Python]
stacked_rnn = tf.contrib.rnn.MultiRNNCell([tf.contrib.rnn.GRUCell(hidden_units), tf.contrib.rnn.GRUCell(hidden_units)]
attentions = tf.contrib.rnn.AttentionCellWrapper(stacked_rnn, attn_length=30, state_is_tuple=True)
rnn_outputs, _ = tf.nn.dynamic_rnn(attentions, feed_in, dtype=tf.float32)
\end{lstlisting}
\end{scriptsize}

\subsubsection{IndRNN}
The IndRNNCell() API used for constructing IndRNN is cloned from \cite{IndRNNRepo}. The hidden\_units is 128 and the parameter recurrent\_max\_abs is used to regulate each neuron's recurrent weight as recommended in the paper\citep{IndRNNRepo}.

\begin{scriptsize}
\estiloPython
\begin{lstlisting}[language=Python]
TIME_STEPS = sequence_length
RECURRENT_MAX = pow(2, 1 / TIME_STEPS)
stacked_rnn = tf.contrib.rnn.MultiRNNCell([IndRNNCell(hidden_units, recurrent_max_abs=RECURRENT_MAX), IndRNNCell(hidden_units, recurrent_max_abs=RECURRENT_MAX)])
\end{lstlisting}
\end{scriptsize}







\subsection{FC} 
The output layer of the whole neural network is fully connected layer which has 2 neurons, with one predicting valence and another predicting arousal value. The output from the RNN block is of shape ${batch\_size \times sequence\_length \times hidden\_units}$. In order to be fed into the FC layer, the rnn\_output is reshaped to\newline ${batch\_size * sequence\_length \times hidden\_units}$ using tf.reshape() function. 

\begin{scriptsize}
\estiloPython
\begin{lstlisting}[language=Python]
fc_inputs = tf.reshape(rnn_outputs, (batch_size * seq_length, hidden_units))
fc_output = slim.layers.linear(fc_inputs, number_of_outputs)
prediction = tf.reshape(prediction, (batch_size, seq_length, number_of_outputs))
\end{lstlisting}
\end{scriptsize}

The slim.layers.linear() function is used to construct the fully connected layer by accepting the input tensor and the target number of predictions and returning the output tensor.
Here number\_of\_outputs is 2 which means there are two target values to be predicted, valence and arousal. The output from fc layer has shape of ${batch\_size * sequence\_length \times 2}$. Finally, the prediction is reshaped into\newline ${batch\_size \times sequence\_length \times 2}$ so that the prediction have corresponding shape with the ground truth values.

\section{Loss Function}
\label{sec:lossfunction}
The loss function used to evaluate the performance and applied to train the parameters consist of Concordance Correlation Coefficient (CCC) \citep{lawrence1989concordance} and Mean Squared Error(MSE). In particular, CCC is defined as:
\begin{equation}
\begin{aligned}
&\rho_{c}=\frac{2s_{x_y}}{s_{x}^2 + s_{y}^2 + (\Bar{x}-\Bar{y})^2}
\end{aligned}
\label{CCC}
\end{equation}
where ${s_{x}^2}$ and ${\Bar{x}}$ are the variances and mean value of the predicted values, ${s_{y}^2}$ and ${\Bar{y}}$ are the variances and mean value of the ground truth values. And ${s_{x_y}}$ is the covariance value of predicted and ground truth value. By improving the CCC value, predictions will become highly correlated with the ground truth, have smaller variation and get closer to the ground truth. In this way, the predicted values can better reflect the nature of ground truth values. Due to these reasons, ${1 - CCC}$ value is minimized dominantly in training process. The \textit{concordance\_cc2(predictions, labels)} function in \textit{losses.py} source code file is implemented to return the ${1 - CCC}$ value by receiving predictions and ground truth values for every batch of data.

\begin{scriptsize}
\estiloPython
\begin{lstlisting}[language=Python]
def concordance_cc2(predictions, labels):
    CCC = ...
    ...
    return 1 - CCC
\end{lstlisting}
\end{scriptsize}
MSE is defined as:
\begin{equation}
\begin{aligned}
&MSE=\frac{1}{N}\sum_{i=1}^{N}(x_{i}-y_{i})^2
\end{aligned}
\label{CCC}
\end{equation}
where ${x_{i}}$ represents one single predicted value and ${y_{i}}$ stands for one single ground truth value. N means the total number of samples. \newline

In training process, the ${MSE}$ and ${1-CCC}$ values are computed by \textit{compute\_loss()} function which accepts the prediction and ground truth values and computes the ${1-CCC}$ and ${MSE}$ value and stores in loss\_ccc and loss\_mse varaibles for valence and arousal separately. And the ${1-CCC}$ value is added by \textit{slim.losses.add\_loss()} function so that this loss can be managed together with the collection of loss functions in TensorFlow.

\begin{scriptsize}
\estiloPython
\begin{lstlisting}[language=Python]
def compute_loss(predictions, labels):
    for i, name in enumerate(['valence', 'arousal']):
        loss_ccc = compute_CCC(predictions, labels)
        loss_mse = ...
        slim.losses.add_loss(loss_ccc / 2.)
\end{lstlisting}
\end{scriptsize}

\section{Training Program}
\label{sec:trainimp}
In previous sections, the whole end-to-end system is built so that the predictions of the valence and arousal for each frame are generated. Training process aims to optimize (\cite{glimm2013using,horrocks2011answering}) the parameters used in this neural network so that the predictions produced by the system can be brought closer to the ground truth values. The deviation between ground truth and prediction is measured by ${1-CCC}$ value which is called the target loss function. With respect to this loss value, the trainable variables can be updated by the computed gradient so as to minimize the loss value. The trainable parameters can be updated every time a new batch of data fed into the network. The implemented training process for experiments is designed to feed infinite train data to the network so that the model can be trained continuously until stopped manually. Now the details are demonstrated as following.\newline

To implement training process, two functions, \textit{slim.learning.create\_train\_op} and\newline \textit{slim.learning.train} provided by slim library \citep{slimLibrary}, are used to perform the optimization.
To start with, the loss value and optimizer are needed.

\begin{enumerate}
    \item Total Loss: The total loss is gotten by invoking function \textit{slim.losses.get\_total\_loss()} which returns the value added in loss function compute\_loss() defined in section \ref{sec:lossfunction} by invoking function \textit{slim.losses.add\_loss()}.
    \item Optimizer: The optimizer is created by \textit{tf.train.AdamOptimizer()} function initialized by learning rate which implements the Adam algorithm. The choice is adopted since this Adam optimizer has best performance in experiments mentioned in \cite{kollias2}.
\end{enumerate}

With the total loss and the optimizer arguments, the \textit{slim.learning.create\_train\_op} function is used to create a \textit{train\_op} which when evaluated in session computes the loss value and apply the gradient defined as the code shown below.
\begin{scriptsize}
\estiloPython
\begin{lstlisting}[language=Python]
compute_loss(prediction, label_batch)
total_loss = slim.losses.get_total_loss()
optimizer = tf.train.AdamOptimizer(LEARNING_RATE)
train_op = slim.learning.create_train_op(total_loss, optimizer)
\end{lstlisting}
\end{scriptsize}

Then the \textit{slim.learning.train()} function is used to execute the train loop as shown below. The main work of this function is that evaluating the \textit{train\_op} every step of training until the data is exhausted.
\begin{scriptsize}
\estiloPython
\begin{lstlisting}[language=Python]
slim.learning.train(train_op,
                TRAIN_DIR,
                init_fn=init_fn,
                save_summaries_secs=60 * 15,
                log_every_n_steps=500, 
                save_interval_secs=60 * 15, 
                saver=tf.train.Saver(max_to_keep=10000)
                )
\end{lstlisting}
\end{scriptsize}
Considering the arguments used in this function:
\begin{enumerate}
    \item train\_op is stated above.
    \item TRAIN\_DIR represents the system path where to store the check points generated and the summary information.
    \item init\_fn is used to restore the variables at the beginning of the training process. Usually, this is used to restore variables in the pre-trained model. Function \textit{slim.assign\_from\_checkpoint\_fn()} is used to create this \textit{init\_fn} by specifying the arguments of \textit{variables\_to\_ restore} from the computational graph and the \textit{check\_point\_path} of pre-trained model.
    \item saver is defined by \textit{tf.train.Saver(max\_to\_keep=10000)} function which is used to save all the variables used in the training. With the max\_to\_keep argument, the maximum number of check points allowed to save is specified.
    \item During the process of training, the check points are stored every \textit{save\_interval\_secs} seconds and log information is output every \textit{log\_every\_n\_steps } steps. And the summary is saved every \textit{save\_summaries\_secs} seconds.
\end{enumerate}

In training process, the TFRecords of train data is fed into the \textit{load\_data()} function of \textit{DataLoader} instance with \textit{is\_training} parameter set to True so that the train data is fed into the neural network infinitely in training process. Then with the \textit{slim.learning.train()} function, the model is trained using Adam optimizer with respect to ${1-CCC}$ loss value continuously until the training process is stopped manually. So once the training program is executed, the check points for trained model are stored endlessly. 


\section{Evaluating and Testing}
\label{sec:evaimp}
\subsection{Evaluating}
Training process needs monitoring to know when the model is trained properly so that the training process can be stopped. Evaluating program is executed for this purpose. While the training program is running, the evaluation program is also running in parallel, waiting for the check points generated by training program. Following is the procedure of how this evaluating program works: 

\subsubsection{Initialize DataLoader class for evaluation}
To start with, the data input should be set up. As stated in section \ref{sec:dataloader}, the DataLoader class to be used in evaluating program is initialized by BATCH\_SIZE=1 and NUM\_EPOCHS=1. And the TFRecords files path of validation data is fed into the \textit{load\_data} function and the \textit{is\_training} parameter is set to False so that the data is repeated by the number specified by the NUM\_EPOCHS parameter and will not be shuffled.

\subsubsection{Build neural network to give inference}
With the validation data generated, the predictions can be produced by the model stored in check point. This is implemented by building the same computational graph as the graph stored in check point and restore the model weights on the graph. 

\subsubsection{Compute the performance measurement}
The performance measurement CCC and MSE value mentioned in section \ref{sec:lossfunction} can be computed on predictions and ground truth value by the metrics function\newline \textit{slim.metrics.streaming\_mean\_squared\_error()} and \textit{slim.metrics.streaming\_covariance()} provided by slim library \citep{slimLibrary}.

\subsubsection{Summary the performance value}
Use the \textit{tf.summary.scalar()} function to record the performance CCC and MSE value mentiond in section \ref{sec:lossfunction} of the model on validation data. In this way, the performance value can be monitored using TensorBoard\footnote{https://www.tensorflow.org/guide/summaries\_and\_tensorboard}.

\subsubsection{Evaluate continuously}
The function \textit{slim.evaluation.evaluation\_loop()} is used to execute the evaluation program. The evaluation program will keep waiting for the check points generated by training program and keep evaluating the performance and summarizing the CCC and MSE value. By looking at the TensorBoard evaluation line graph, when the evaluation CCC value stop increasing and begin to decrease, it is the time to stop the training process since the model begin over-fitting. And the check point which has the best performance on the validation data set in terms of CCC value is selected as the best performance model.

\subsection{Testing}
The test program is very similar to the evaluation program. The only difference between evaluate and test program is the input data. For test program, the TFRecords files path of test data is provided to \textit{DataLoader} class. After the evaluation program finished, namely the best performance check point is found, the test program is run by restoring the best performance check point model and evaluating the performance on test data. This test performance is used as the criteria of evaluating the generalization performance of one architecture. 

\section{Program environment}
\label{sec:envimp}
The python scripts of train, validate and test program run in python 2.7 environment. The tensorflow version is 1.4.0 and cuda version used is 8.0.61-cudnn.7.0.2. The GPU used is provided by CSG GPU cluster.

\section{Summary}
\label{implementationsummary}
The created database is stored into TFRecords files and imported into neural networks by Dataset API. The code used for constructing CNNs, RNNs and FC are illustrated. The experiments will be executed by running train and evaluate program at the same time on different GPUs. While the train program is running, the check points are produced and saved regularly. For evaluating the check point performance, the evaluate program will keep waiting for the check points and evaluate the check point's performance once the new check point is appeared. Finally, the test program is executed using the best validation performance check point on test data. The evaluate program performing evaluation process aims to find the best performance model. Test program running performance measurement process on test data aims to evaluate the generalization ability of the model.
\chapter{Experiments and Evaluation}
\label{experimentChapter}
The aim of this Chapter is to find the best performance CNN and RNN combination architecture and give appropriate evaluations on the results of the experiments. The details of how my experiments executed are demonstrated in Section \ref{sec:exproc}. The whole architecture of the deep neural network is designed as stated in Section \ref{CRNNdedign}. The basic idea is all the CNNs and RNNs mentioned before in Design Chapter \ref{designChapter} will be combined together and examined. First, I will try different RNNs on top of VGGFace network which will be detailed in section \ref{sec:expVggface}. After the best performance RNN is decided, different CNNs were tried out in section \ref{sec:expattention}. Finally, with the best CNN and RNN architecture, little hyper-parameter tuning was executed in section \ref{sec:exphyper}. For each part of the experiments, the detailed evaluation is provided. \newline 

The contribution of this Chapter is the best performance architecture of the neural network is obtained through the experiments and the best training strategy to get best performance model is also discovered based on the experiments so far. At the same time, the evaluation is also given as the experiments go.
 
\section{Experiment Procedure}
\label{sec:exproc}
\begin{figure}[H]
    \centering
    \includegraphics[height=8cm,width=14cm]{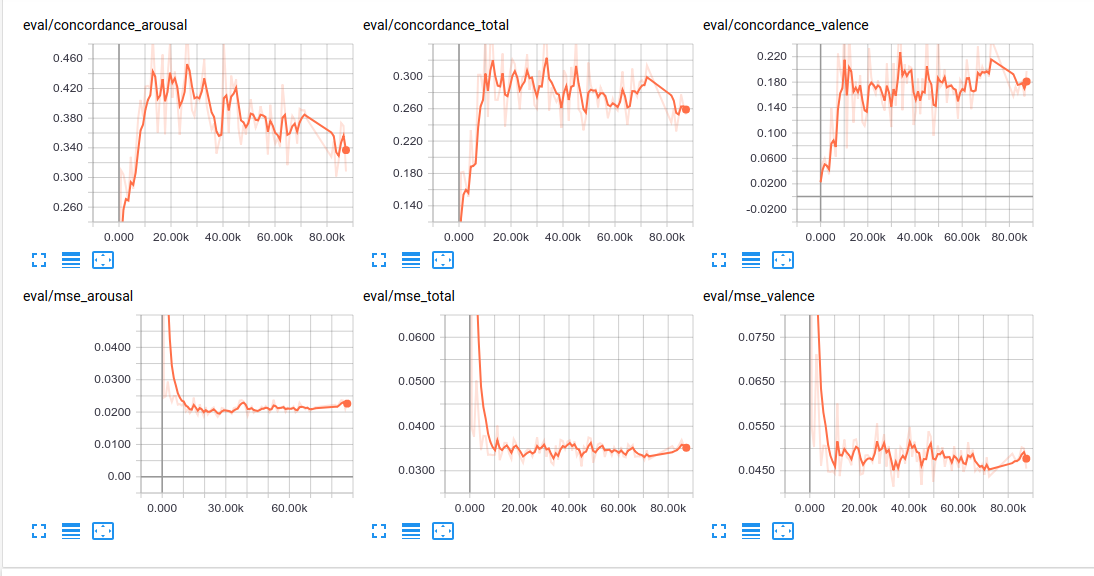}
    \caption{The graph shows the CCC and MSE performance on validation data set for the check points generated as training for VGGFace network combined with RNN block consisting of GRU with attention.}
    \label{evaluationbest}
\end{figure}

As stated in Chapter \ref{implementationChapter}, the train and evaluation programs are running on two GPUs at the same time, with the train program training on train data and evaluation program evaluating performance of generated check points on validation data. The CCC value is the dominant measurement for evaluating model performance. The model having a higher CCC value is considered to be better. Once the best validation performance model is found, the test program is run with this best performance check point on the test data set and gives a test performance as the final generalization performance for the corresponding architecture design. All the experiments are executed in this manner and use the test performance to compare the different architectures. The advantage of this training and evaluating approach is the training program can be executed in an efficient way in which the training process will not be interrupted by the evaluation process. The training process for finding the best validation performance usually costs around 2 days. As an example, with the CCC performance on validation data shown in Figure \ref{evaluationbest}, we can see the best performance for arousal appears around 25K global step and best performance for valence appears around the 35K global step. In this case, these two check points are all kept for later use in evaluating CCC performance on the test data set.

\begin{table}[H]
\centering
\begin{tabular}{|l|l|}
\hline
\multicolumn{2}{|c|}{Training Hyper-parameters} \\ \hline
Hyper-Parameters & Value \\ \hline
Image Size & 96x96x3 \\ \hline
Batch Size & 2 \\ \hline
Sequence Length & 80 \\ \hline
CNN configuration & VGGFace Network \\ \hline
Initialization for CNN & VGGFace Model \\ \hline
Number of Layers in RNN block & 2 \\ \hline
Number of Hidden Units & 128 \\ \hline
Initialization for RNN & random initialization \\ \hline
Initialization for FC & random initialization \\ \hline
Optimizer & Adam \\ \hline
\end{tabular}
\caption{Training parameters used for all experiments on RNNs.}
\label{tab:paramVGGface}
\end{table}

\section{Experiments Results on RNNs}
\label{sec:expVggface}
To start with, the different RNNs are experimented so as to find the best performed RNN architecture. To control the experiment conditions, all the experiments for selecting RNN are set under the same configurations, except using different RNN blocks. The CNN part uses VGGFace network and is initialized with the VGGFace pre-trained model as detailed in Section \ref{VGGFaceDesign} and the variables in the CNN part are set to not trainable. The rest of the conditions of the experiment is demonstrated in the Table \ref{tab:paramVGGface}. The number of hidden layer in RNN block is set to 2 and hidden units of RNN layer is set to 128 based on the experiments result in \cite{kollias2}. The learning rate is set to 0.001 at first for all RNN experiments. However, the GRU with attention mechanism and IndRNN models are not converged, so the learning rate for these two models is changed to 0.0001.

The experiments outcomes are recorded in Table \ref{tab:RNNsExp}. At first, the GRU, LSTM and IndRNN are experimented. Although the IndRNN has the best test performance in valence value in terms of CCC, the average test performance of GRU cell is the best. So the attention mechanism with attention length 30 is only applied on GRU and the model of GRU with attention mechanism has the best test performance with CCC value of valence is 0.412 and CCC value of arousal is 0.409.

\begin{table}[H]
    \centering
    \begin{tabular}{|c|c|c|c|c|}
	\hline
	RNN Block used & CCC-Valence & CCC-Arousal & MSE-Valence & MSE-Arousal \\
	\hline
	\rowcolor[HTML]{EFEFEF} 
    LSTM & 0.290 & 0.294 & 0.031 & 0.023\\
    \rowcolor[HTML]{EFEFEF} 
    LSTM & 0.303 & 0.249 & 0.037 & 0.029\\
	\hline
    GRU & 0.156 & 0.362 & 0.054 & 0.022\\
    GRU & 0.353 & 0.358 & 0.034 & 0.023\\
    \hline
    \rowcolor[HTML]{EFEFEF} 
    GRU-Attention & 0.347 & \textbf{0.409} & 0.033 & 0.020\\
    \rowcolor[HTML]{EFEFEF} 
    GRU-Attention & \textbf{0.412} & 0.342 & 0.032 & 0.026\\
    \hline
    IndRNN & 0.332 & 0.237 & 0.031 & 0.027\\
    IndRNN & 0.392 & 0.254 & 0.028 & 0.023\\
	\hline
\end{tabular}
    \caption{The results of CCC and MSE performance evaluated on test data set on different RNN cells with the same basic hyper-parameters shown in Table \ref{tab:paramVGGface}. Test performance of two check points are recorded for each RNN architecture. This is due to the fact that the best performing model of valence is different from the one of arousal on validation data sometimes.}
    \label{tab:RNNsExp}
\end{table}

\subsection{Evaluation on RNNs Performance}
\subsubsection{GRU vs LSTM}
The test performance of GRU has 0.353 CCC value of valence and 0.362 CCC value of arousal. And the CCC value for test performance on LSTM is 0.303 for valence and 0.294 for arousal. It is reasonable to see the result of GRU is better. From the perspective of the theory, GRU is the simplified version of LSTM so that GRU is more computationally efficient. And in the experiments stated in \cite{Jozefowicz2015AnArchitectures}, the GRU performs better than LSTM in almost all tasks. 

\subsubsection{Performance of IndRNN}
The CCC test performance of IndRNN is 0.392 for valence and 0.254 for arousal. The valence performance is better than LSTM's and GRU's. But the average performance on valence and arousal is worse than GRU's. In theory, the IndRNN is designed to have the ability to model much longer sequential data and build deeper RNNs block. One factor that may lead to the unexpected performance of IndRNN is the value of learning rate. In later experiments I found sometimes the learning rate may affect the model's performance a lot. Here the learning rate is set to 0.0001 and did not make much more experiments to find the best learning rate for IndRNN which may lead to poor performance of IndRNN. But due to the time limitation, this test performance is used to be the final performance of IndRNN in this experiments. 

\subsubsection{Attention mechanism on GRU}
Here the GRU cell with attention mechanism has the best performance, which has 0.412 for valence CCC and 0.409 for arousal CCC value. It is more suitable to give an explanation in terms of the video scenario. The facial expression for every frame in a video is changing every moment so that long-term memory may not have much effect on current emotion analysis. Let the RNN concentrate on last 30 frames, namely last 1 second, will make the analysis more accurate to some extent.

\section{Experiments on CNNs}
\label{sec:expattention}
With the best performance RNN block fixed, namely the 2 layers GRU with attention mechanism, different CNN blocks described in Chapter \ref{designChapter} are experimented. In details, the VGGFace network pre-trained on VGGFace database(VGGFace), ResNet50 network pre-trained on VGGFace2(ResNet50-VF2), ResNet50 network pre-trained on ImageNet(ResNet50-IN),  DenseNet121 network pre-trained on ImageNet(DenseNet121-IN) and DenseNet169 network pre-trained on ImageNet(DenseNet169-IN) are experimented. And for VGGFace, ResNet50-IN and DenseNet121-IN, there are three additional sub experiments are executed and the details are listed in Table \ref{tab:cases}. And the test performance for all examined architectures are listed in Table \ref{tab:CNNs}.

\begin{table}[H]
\centering
\begin{tabular}{|l|l|l|}
\hline
       & CNN Trainable Part      & RNN Initialization \\ \hline
Case 0(c0) & None                    & Random             \\ \hline
Case 1(c1) & Last Conv Layer         & Random             \\ \hline
Case 2(c2) & All Trainable Variables & Random             \\ \hline
Case 3(c3) & All Trainable Variables & Best RNN network from Case 2 \\ \hline
\end{tabular}
\caption{The four training strategies are described as four cases. The CNN block is initialized with respective pre-trained model and RNN block is all trainable for all these four cases.}
\label{tab:cases}
\end{table}

\begin{table}[H]
    \centering
    \begin{tabular}{|c|c|c|c|c|}
	\hline
	CNN Block used & CCC-Valence & CCC-Arousal & MSE-Valence & MSE-Arousal \\
	\hline
    VGGFace(c0) & 0.347 & 0.409 & 0.033 & 0.020\\
    VGGFace(c0) & 0.412 & 0.342 & 0.032 & 0.026\\
	\hline 
	VGGFace(c1) & 0.353 & 0.313 & 0.024 & 0.023\\
	\hline 
    VGGFace(c2) & 0.479 & 0.327 & 0.020 & 0.020\\
    \hline
    VGGFace(c3) & \textbf{0.555} & \textbf{0.456} & \textbf{0.018} & \textbf{0.017}\\
    \hline
    \rowcolor[HTML]{EFEFEF} 
	ResNet50-IN(c0) & 0.280 & 0.107 & 0.044 & 0.034\\
	\hline 
	\rowcolor[HTML]{EFEFEF} 
	ResNet50-IN(c1) & 0.162 & 0.071 & 0.068 & 0.046\\
	\rowcolor[HTML]{EFEFEF} 
	ResNet50-IN(c1) & 0.163 & 0.037 & 0.048 & 0.038\\
	\hline 
	\rowcolor[HTML]{EFEFEF} 
	ResNet50-IN(c2) & 0.303 & 0.393 & 0.033 & 0.019\\
	\hline 
	\rowcolor[HTML]{EFEFEF} 
	ResNet50-IN(c3) & 0.332 & 0.429 & 0.027 & 0.019\\
	\hline 
    DenseNet121-IN(c0) & 0.154 & 0.185 & 0.061 & 0.034\\
    \hline
    DenseNet121-IN(c2) & 0.468 & 0.294  & 0.024 & 0.022 \\
    \hline
    DenseNet121-IN(c3) & 0.449 & 0.306  & 0.022 & 0.021 \\
    DenseNet121-IN(c3) & 0.368 & 0.312  & 0.025 & 0.021 \\
    \hline
\end{tabular}
    \caption{ The test performance of experiments on different CNN blocks with the same RNN block. The RNN block for all experiments is set to have GRUs with attention mechanism. Test performance of two check points are recorded for some CNN architectures. This is due to the fact that the best performing model of valence is different from the one of arousal on validation data sometimes.}
    \label{tab:CNNs}
\end{table}

\subsection{Evaluation}
\subsubsection{Evaluation on DenseNet}
Here the CNN blocks based on DenseNet121-IN and DenseNet169-IN are experimented under c0 setting. Although the DenseNet169 has better performance in object detection task as stated in section \ref{DenseNetDesign}, here the CNN block based on DenseNet121 design has better performance than based on DenseNet169. There are two main factors. One is the image spatial size used in this project is ${96 \times 96}$ while the image size used in ImageNet is ${224 \times 224}$. So with the stronger capability to extract deeper features, DenseNet169 based CNN block may over-fit on the training data. The second factor is the amount of data collected in this project is much smaller than ImageNet. This fact can also lead to DenseNet169 over-fitting on the train data of the created database. So for DenseNet169, only c0 experiment is executed. For DenseNet121, c2 and c3 are also examined. 

\subsubsection{Evaluation on ResNet50-VF2}
Under case 0 setting, comparing with VGGFace network, this ResNet50-VF2 model has worse performance. In theory, experiments in \cite{VGGFace2Cao18} show that ResNet-50 network trained on VGGFace2 performs better than the one trained with VGGFace on face recognition problem. But here, the ResNet-50 network together with the VGGFace2 model is converted from Caffe version which may lead to the poor training experiment since the converted TensorFlow version code is not constructed correctly especially the batch norm layer implementation. So later on, this ResNet50-VF2 model will not be further discussed.

\subsubsection{Evaluation on Four Cases}
\begin{itemize}
    \item For case 0, the CNN block is set to not trainable so that the CNN block is used as a fixed visual feature extractor.
    
    \item For case 1, the difference compared with case 0 is that except for RNN block, the last convolutional layer in CNN block is also set to trainable. Case 1 experiment is only executed for VGGFace and ResNet50-IN and has lower CCC values on valence and arousal than their case 2 or case 3. This fact shows that the task here have much divergence with the task for ResNet50-IN even for VGGFace pre-trained model so that only fine tuning the dense layer is not sufficient to fit in the task of emotion recognition here. Moreover, the Case 1 experiment result in lower performance than case 0 experiment as well for ResNet50-IN and VGGFace. This mainly due to the fact that the pre-trained model used to extract original task image feature as a whole, if only last conv layer is updated, the feature extracted by shallow layers may not be compatible with the last conv layer.
    
    \item For case 2, the only difference is the whole CNN block is set to trainable based on the setting of case 0. And almost for all different CNN blocks used in the experiments, the test performance of case 2 is better than the corresponding case 0 experiment's result. The underlying principle of this phenomenon is that the variables in CNN block are updated based on the database collected which means the model learns the pattern in my database and adjust the parameters in the model so that the performance becomes better.
    
    \item For case 3, the only difference is the RNN block use the weights of the trained model from case 2 compared with the case 2 setting. Usually, for almost all CNN blocks, the test performance of case 3 is better than their own case 2 performance. Appropriate initialization for RNN is crucial to the performance which makes the model begin to learn at some point in the parameters space where close to the global minimum of the target loss function. By initializing the RNN model with the pre-trained model in case 2, the training process of case 3 usually converges faster and achieves better. One proof is shown in Figure \ref{case23comparison}. The figure \ref{case23comparison} demonstrates the evaluation graph for the model of ResNet50-IN connected with GRU-Attention. The blue line is for case 3 and the orange line is for case 2. It is clear that the blue line increases more steeply than the orange line and rises at an earlier global step. This phenomenon can be a good indication that proper initialization can make the loss function converges faster and better.  
    
    \begin{figure}[H]
        \centering
        \includegraphics[height=7cm,width=12cm]{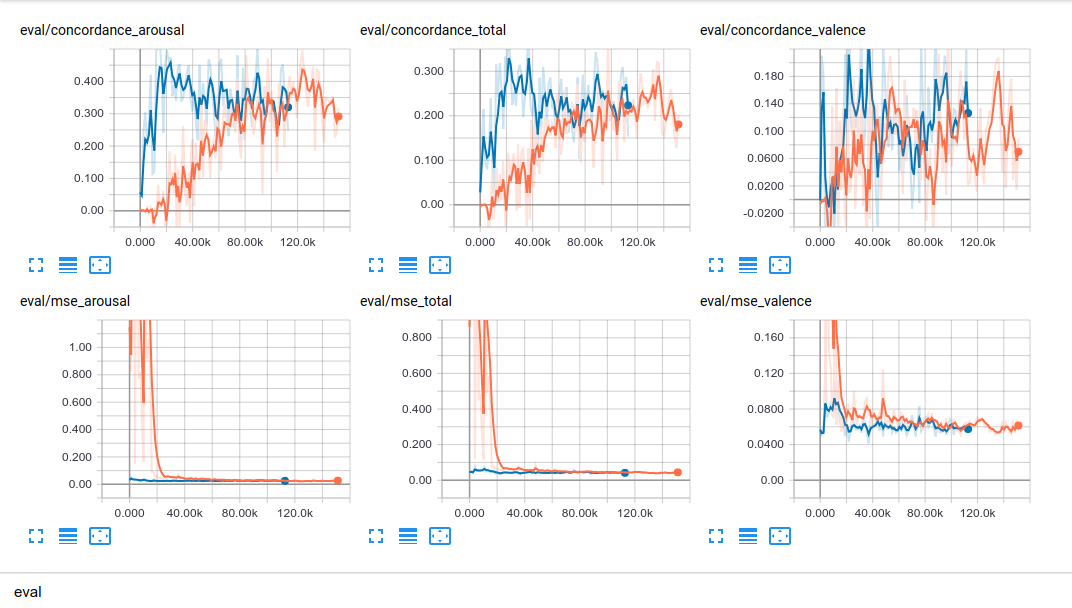}
        \caption{The graph shows the evaluation graph for model of ResNet50-IN connected with GRU-Attention. The blue line is for case 3 and orange line is for case 2.}
        \label{case23comparison}
    \end{figure}

    \item However, there exists some exception situation. For VGGFace network, the case 2 performance of arousal is worse than case 0. For DenseNet-121 performance, the case 3 performance of valence is worse than case 2. Sometimes the validation performance can not reflect test performance truly.

\end{itemize} 

\subsubsection{Evaluation on CNNs}
Comparing the best test performance of VGGFace, ResNet50-IN and DenseNet121-IN, it is reasonable to see the test performance of VGGFace network has better performance. The reason is that VGGFace network pre-trained on VGGFace database is used to fulfill face recognition task. While the other models trained on ImageNet are used to perform general object localization and detection task. So the task of VGGFace pre-trained model is much more similar with the task of the facial expression recognition task here, compared to general object recognition task. 

\section{Experiments on Hyper-parameters}
\label{sec:hyperparams}
After the previous experiments, the CNN and RNN block is selected based on the test performance. To be concrete, the VGGFace network pre-trained on VGGFace database is chosen as the CNN block and 2 layers GRU with attention mechanism is used as RNN block. What is more, the best test performance is trained under case 3 training strategy. So the following experiments aiming to find the best hyper-parameters are executed under these configurations and the only variable is the particular hyper-parameters needed to be experimented. Due to the time limitation, only sequence length is examined. The results of different sequence length experimented under case 2 and case 3 are listed in Table \ref{tab:seqlens}.

\label{sec:exphyper}
\begin{table}[H]
    \centering
    \begin{tabular}{|c|c|c|c|c|}
	\hline
	Sequence Length & CCC-Valence & CCC-Arousal & MSE-Valence & MSE-Arousal \\
	\hline 
	\rowcolor[HTML]{EFEFEF} 
    70 (c2) & 0.367 & 0.346 & 0.023 & 0.020\\
    \hline
    \rowcolor[HTML]{EFEFEF} 
    70 (c3) & 0.403 & \textbf{0.499} & 0.021 & \textbf{0.016} \\
    \rowcolor[HTML]{EFEFEF} 
    70 (c3) & 0.491 & 0.424 & 0.020 & 0.018 \\
	\hline 
	80 (c2) & 0.479 & 0.327 & 0.020 & 0.020\\
    \hline
    80 (c3) & \textbf{0.555} & 0.456 & \textbf{0.018} & 0.017\\
	\hline 
	\rowcolor[HTML]{EFEFEF} 
	100 (c2) & 0.403 & 0.459 & 0.025 & 0.016 \\
    \hline
    \rowcolor[HTML]{EFEFEF} 
    100 (c3) & 0.351 & 0.404 & 0.025 & 0.017 \\
    \rowcolor[HTML]{EFEFEF} 
    100 (c3) & 0.498 & 0.388 & 0.020 & 0.018 \\
	\hline 
    \end{tabular}
    \caption{Test performance of different Sequence Length for VGGFace-GRUs-Attention combination under case 2 and case 3 training strategy.}
    \label{tab:seqlens}
\end{table}

\subsubsection{Evaluation}
Although the best test performance of valence and arousal appear in different sequence length experiments, the sequence length of 80 is chosen as the final parameter value for further experiments in the future. Two factors are considered. Firstly, the average test performance of valence and arousal for sequence length 80 is 0.506 while this value for sequence length 70 is 0.495. Secondly, the best performance of valence and arousal appear in the same check point of sequence length 80 under case 3 in which situation the performance is more robust. On the contrary, the best test performance for valence and arousal appear in 2 different check point in sequence length 70 under case 3. What is more, the attention length 30 is used which may also affect on the performance of different sequence length. This needs further experiments. 

\section{Experiments Results}
Through all the experiments in this Chapter, the best performance architecture is found. VGGFace network pre-trained on VGGFace database is used as CNN block and 2 layers GRU having 128 hidden units with attention mechanism having attention length 30 is applied as RNN block. The other parameters are stated in Table \ref{tab:paramVGGface}. Adam optimizer is used with learning rate 0.00001. The way of training this architecture is first randomly initialized the RNN and load the VGGFace model for VGGFace network then training the whole model and produce the best performance check point, which is the case 2 strategy. Then initialize the RNN block with the best model in case 2 and initialize the CNN block with the VGGFace pre-trained model and train the whole model together generating the best performing model. The best test performance CCC value for valence is 0.555 which is produced with sequence length 80 while the best test performance CCC value for arousal is 0.499 which is generated with sequence length 70. We can denote the best performing model for valence as VGGFace-GRU-Atten-80 and denote the best performing model for arousal as VGGFace-GRU-Atten-70. \newline

\subsection{Evaluation on Experiment Result}
\begin{figure}[H]
    \caption{Comparison of ground truth and predictions of all test data for arousal. Here the predictions are produced by the model of 70 (c3) in Table \ref{tab:seqlens} which has 0.499 CCC value for arousal.}
    \label{arousal70_all}
    \centering
    \includegraphics[height=5cm,width=10cm]{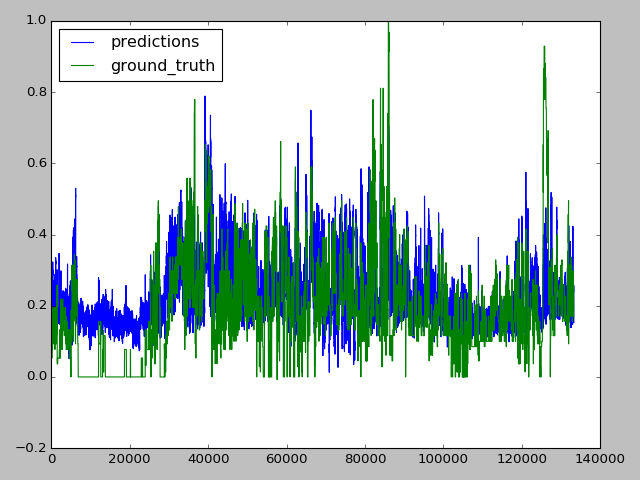}
\end{figure}

\begin{figure}[H]
    \centering
    \includegraphics[height=5cm,width=10cm]{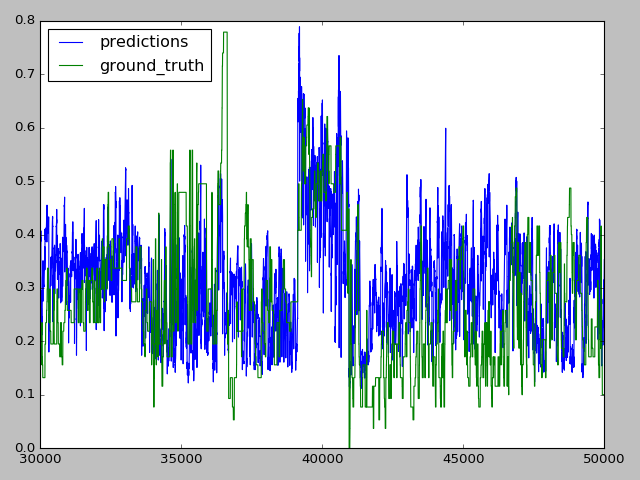}
    \caption{Comparison of ground truth and predictions of test data from frame number 30000 to 50000 for arousal. This figure shows high CCC value on prediction and true value. Here the predictions are produced by the model of 70 (c3) in Table \ref{tab:seqlens} which has 0.499 CCC value for arousal.}
    \label{arousal70_3_5}
\end{figure}

\begin{figure}[H]
    \centering
    \includegraphics[height=5cm,width=10cm]{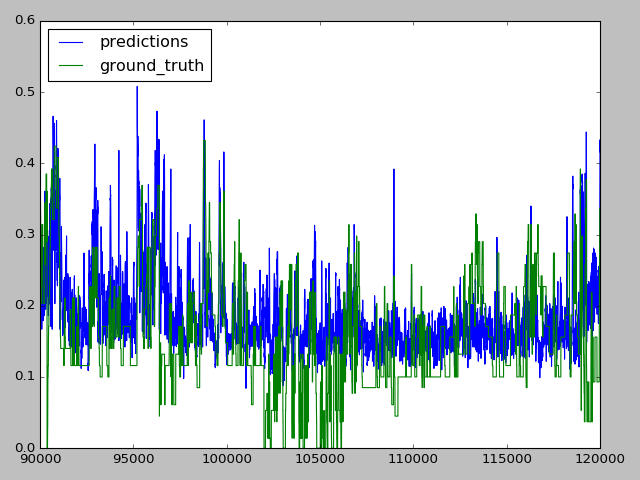}
    \caption{Comparison of ground truth and predictions of test data from frame number 90000 to 120000 for arousal. This figure shows high CCC value on prediction and true value. Here the predictions are produced by the model of 70 (c3) in Table \ref{tab:seqlens} which has 0.499 CCC value for arousal.}
    \label{arousal70_9_12}
\end{figure}

\begin{figure}[H]
    \caption{Comparison of ground truth and predictions of all test data for valence. Here the predictions are produced by the model of 80 (c3) in Table \ref{tab:seqlens} which has 0.555 CCC value for valence.}
    \label{valence_all}
    \centering
    \includegraphics[height=5cm,width=10cm]{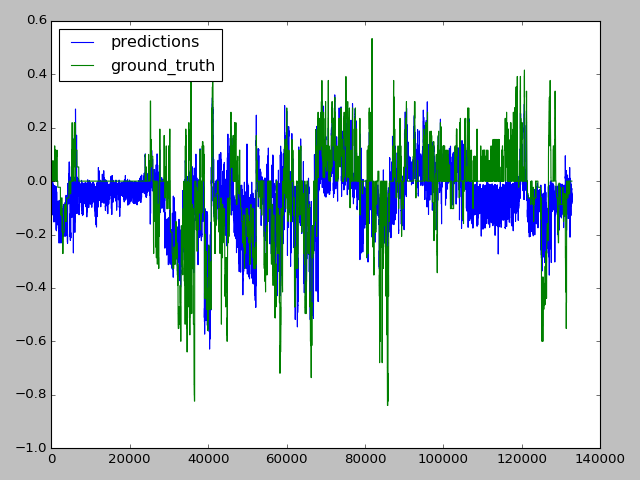}
\end{figure}

\begin{figure}[H]
    \caption{Comparison of ground truth and predictions of test data from frame number 50000 to 70000 for valence. This figure shows high CCC value on prediction and true value. Here the predictions are produced by the model of 80 (c3) in Table \ref{tab:seqlens} which has 0.555 CCC value for valence.}
    \label{valence_5_7}
    \centering
    \includegraphics[height=5cm,width=10cm]{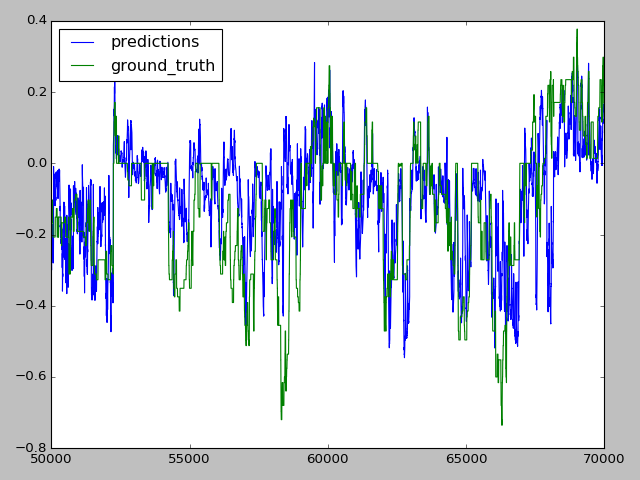}
\end{figure}

\begin{figure}[H]
    \caption{Comparison of ground truth and predictions of test data from frame number 70000 to 100000 for valence. This figure shows high CCC value on prediction and true value. Here the predictions are produced by the model of 80 (c3) in Table \ref{tab:seqlens} which has 0.555 CCC value for valence.}
    \label{valence_7_10}
    \centering
    \includegraphics[height=5cm,width=10cm]{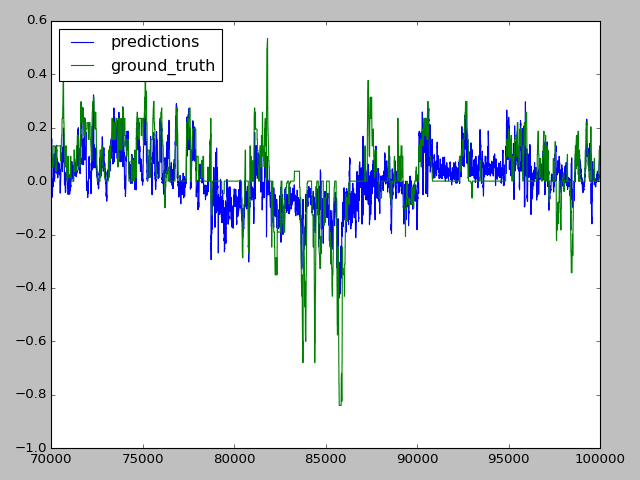}
\end{figure}

In the following graph, the predictions produced by VGGFace-GRU-Atten-80 and  VGGFace-GRU-Atten-70 are plot together with the corresponding ground truth values. In Figure \ref{arousal70_all} and \ref{valence_all}, the predictions and ground truth values for all frames in the test data set are plot. It is clear that by optimizing the loss function defined by CCC value, the predicted value produced by the best performance model is highly correlated with the true value and the predicted value has a small variance and small offset from the true value. In Figure \ref{arousal70_3_5}, \ref{arousal70_9_12}, \ref{valence_5_7} and \ref{valence_7_10}, the plot for predictions and ground truth values of selected period of frames is shown so that it is more clear to see what the high CCC value means.


\section{Summary}
\label{experimentsummary}
Rich experiments on different CNN and RNN blocks combinations are executed aiming to find the best architecture used to train the best performance model to fulfill the emotion recognition task. By running the training and evaluating program, the best validate performance check points are found and tested on test data. By comparing the test performance on valence and arousal, the best CNN or RNN blocks are chosen. The best RNN block is selected through the experiments on different RNN blocks by fixing the CNN block of VGGFace model. The resulting best RNN block is based on GRU with attention mechanism. Then different CNN blocks are experimented by fixing the chosen RNN block and the VGGFace block is chosen. Further experiments on different training strategies are explored. By analyzing the experiments results, the chosen architecture need trained on all trainable variables and proper initialization on RNN is also important. At the late experiments stage, the sequence length is also examined and 80 is found to have the best performance. From the experiments result, there are a lot interesting findings. Considering RNN block, it is more targeted to solve the emotion recognition in video problem with the attention mechanism rather than other advanced architecture aiming to improve the general performance. In terms of CNN block, the task of pre-trained model of VGGFace is more similar to our emotion recognition task than object detection problem solved by other pre-trained model. So the VGGFace based CNN block has the best performance. Nevertheless, the VGGFace plus GRU with attention architecture with appropriate initialization still need thoroughly training in order to get best performance rather than only fine tuning dense layers. In this way, the model can be trained much faster and achieve better test performance. With the final test performance, 0.555 CCC for valence and 0.499 CCC for arousal, it is verified that the Database created in Chapter \ref{databaseChapter} is well annotated and has tolerable label noises. The automatic emotion recognition system built on VGGFace plus GRU with attention is well trained on the extended database and achieves descent performance by observing the plot of prediction and ground truth comparison in Figure \ref{valence_all} and \ref{arousal70_all}.
\chapter{Conclusion and Future Work}
\label{conclusionChapter}
This chapter first summarizes the achievements of extending Aff-Wild database and building automatic system to recognize facial emotion in sequential data in this project in Section \ref{achievements}. Then list the challenges of creating the database and exploring different architectures in Section \ref{challenges}. An objective evaluation is given in Section \ref{evaluation}. In the end, the future work is discussed in Section \ref{futurework}.

\section{Achievements}
\label{achievements}
Building an automatic system to accomplish human emotion recognition task in real-world condition is very challenging. One dominant reason is the database containing spontaneous facial expressions annotated with valence and arousal under uncontrolled conditions is really rare. Existing databases have more or less problems, like SEMAINE and RECOLA are recorded in well-controlled conditions, SEWA contains facial expressions of limited scenarios as stated in Chapter \ref{backgroundChapter}. Aff-Wild database is satisfied with these requirements but still needs to be extended. Due to this motivation, the \textbf{first achievement} of this project is the accomplishment of extending Aff-Wild database, namely a new database was created following nearly the same procedure of generating the Aff-Wild database as stated in Chapter \ref{databaseChapter}. The basic steps of creating this database consist of searching videos from YouTube, downloading videos with best quality, converting videos to MP4 format to be recognized by annotator program, trimming the videos to be prepared for detecting and tracking faces, converting to 30 FPS so that it is suitable for RNN training and annotating the videos with valence and arousal. After the pre-processing and data set partition, the database was finally be created. To summarize the details, the produced database has 527056(69.67\%), 94223(12.46\%) and 135145(17.87\%) frames in train, validate and test set annotated with valence and arousal value which has continuous value ranging from -1000 to 1000 which is scaled to a range of [-1.0, 1.0]. The distribution of valence value is balanced to some extent while the arousal value is mainly distributed in the positive domain. In total 159 videos, there are 79 videos have male subjects and 80 videos have female subjects. The videos are all in MP4 format and have 30 FPS. The ethnic diversity and age divergence are also guaranteed. \newline

Deep learning model is widely adopted in emotion recognition tasks successfully like the methods used in \cite{jaiswal2016deep}, \cite{kollias2}, \cite{Chen2017MultimodalRecognition}. To analyze the sequence of facial expressions in the created database, an end-to-end deep learning architecture is designed based on the idea mentioned in \cite{kollias2} since this approach achieved the best performance on Aff-Wild Challenge \citep{kollias1} and described in Chapter \ref{designChapter}. The design of the whole architecture used for this project is first using CNN block to extract the visual feature from facial images, then with the extracted features, RNN block is used to model the dynamics in sequential frames before sending the output of RNN block to fully connected layer to perform valence and arousal prediction. To find the best CNN and RNN combination, adequate experiments of different CNN blocks with corresponding pre-trained models and RNN blocks are carried out in Chapter \ref{experimentChapter}. And the best performance CNN block is VGGFace based architecture and resulting RNN block is consisting of 2 layers GRU with attention mechanism with attention length 30. The number of hidden units is 128 in RNN cell. The best performance model is trained using transfer learning techniques meaning initializing CNN and RNN block with corresponding pre-trained model and training the whole architecture to fit the created database. The best-performed model has 0.555 CCC on valence with sequence length 80 and has 0.499 CCC on arousal with sequence length 70. Building an end-to-end deep learning system to recognize the facial expression emotion and finding out the best-performed model through rich experiments are the \textbf{second achievement} of this project.

\section{Challenges of Implementation}
\label{challenges}
\subsubsection{Challenges of Creating Database}
There are two significant challenges in creating the database. The \textbf{first challenge} is that it is challenging to give appropriate valence and arousal values to the facial expressions in the video. Even the videos subtitles are studied and the content is understood, there are still a lot of controversial facial expressions in the video which are difficult to judge. Moreover, the changes of facial expressions are sometimes really fast so that it is difficult to move the attacker in time to give the accurate annotations sometimes. Manual modification to the generated annotation files is needed to avoid spending more time. Due to these factors, the annotation process cost around two weeks. The \textbf{second challenge} is there are more than one cropped objects generated by the detector for one frame although there is only one human face in it. Since the number of images in one video is around 10 thousand, it is not applicable to pick the right detected human face from all detected objects of one frame for all wrongly detected videos manually. The solution used in this project is first to compute the right cropped face image's feature by \textit{cv2.calcHist()} API in OpenCV library \citep{opencv_library} and use this feature value as the standard feature. Then compute the feature of all detected objects of one frame and compare the similarity of these features with the standard feature using \textit{cv2.compareHist()} function and select the object which has the biggest similarity value. By this approach, most wrongly detected cases can be overcome.

\subsubsection{Challenges of Experimenting Architectures}
There are several challenges in building the deep neural network. The \textbf{first challenge} the cropped images have different data format with the frame name and the valence and arousal annotations so that it is complicated to read these data to feed into the neural network. To solve these problems, TFRecords is chosen as the format of storing train, validate and test data so that all the frame image data and annotations can be stored together and be dealt with together. The \textbf{second challenge} is the training time of finding the best performance on the validation set is quite long which is around two days. To solve this problem, the training and validating program are implemented separately using TensorFlow slim library and are run on 2 separate GPUs to gain the time so that more experiments on different CNN and RNN blocks can be executed.

\section{Evaluation on Achievements}
\label{evaluation}
In terms of the created database used in the experiments, the annotations are generated by one annotator. On the contrary, the annotation for AFF-WILD \citep{kollias1} was computed by the mean of the most correlated best annotations among all annotations provided by different annotators. And in other experiments like in \cite{AslanHumanStates}, the labels are provided by experts. So the annotations generated by myself may be less consistent or accurate. But considering the test performance of the final trained model, which has 0.555 CCC value for valence and 0.499 for arousal, the annotation quality of the created database is descent to some extent. With this test performance, it also proves that the model is well-trained as well. The actual prediction effect can be observed from Figure \ref{arousal70_all} for arousal and \ref{valence_all} for valence. Certainly, there still exists some drawbacks of the model trained in this project. From the annotation distribution shown in \ref{annotationDistribution} of the created database, the arousal value is severely biased. In this situation, the trained deep neural network is also biased to some extent.

\section{Future Work}
\label{futurework}
\subsubsection{Experiments on Other Advanced CNNs and RNNs}
Although quite a lot of architectures are experimented with in this project, there are still a lot state-of-the-art architectures and pre-trained models can be further examined. For example, the VGGFace2 pre-trained models\citep{VGGFace2Cao18} can be further explored. In terms of RNN blocks, the performance of IndRNN can be further examined by optimizing the hyper-parameters used. And there still exists a lot other CNN and RNN architectures.

\subsubsection{Hyper-parameters Optimization}
The hyper-parameters used during the training process are mainly adopted from the configuration stated in \cite{kollias2} since the task is quite similar. Due to the time limitation, the experiments on hyper-parameters in Section \ref{sec:hyperparams} only contains the test performance on different sequence length values. In the future work, these hyper-parameters can be further examined: attention length in attention mechanism, learning rate used to set the Adam optimizer, the number of hidden layers in RNN block and the number of hidden units of RNN cell.

\subsubsection{Evaluate on Aff-Wild}
The trained model for facial expression recognition tasks on the created database should be evaluated on the Aff-Wild database so that it is more clear to know the generalization performance of the model trained in this project.

\appendix
\chapter{Ethics Checklist}
The videos downloaded from YouTube involves subjects appearing in the videos. After downloaded, the videos are first converted to MP4 format. Then the videos are trimmed into several short videos. All these trimmed videos are converted to 30 FPS. These are all the operations applied on the videos collected from the YouTube.\newline
The collection has been conducted under the scrutiny and approval of Imperial College Ethical Committee (ICREC). The chosen videos were under Creative Commons License (CCL). \newline

  	\begin{tabularx}{\textwidth}{Xss}\toprule
                \emph{Section 1} & \emph{HUMAN EMBRYOS/FOETUSES} & \\\midrule
        & \emph{yes}  & \emph{no} \\ 
     Does your project involve Human Embryonic Stem Cells?
     &  &\checkmark\\ 
     Does your project involve the use of human embryos? & &\checkmark  \\
     Does your project involve the use of human foetal tissues / cells? & &\checkmark   \\\toprule
       \emph{Section 2} & \emph{HUMANS} & \\\midrule
        & \emph{yes}  & \emph{no} \\ 
     Does your project involve human participants? 
     & \checkmark & \\ \toprule
     \emph{Section 3} & \emph{HUMAN CELLS/TISSUES} & \\\midrule
        & \emph{yes}  & \emph{no} \\ 
     Does your project involve human cells or tissues? (Other than from “Human Embryos/Foetuses” i.e. Section 1)? 
     & &\checkmark \\ \toprule
     \emph{Section 4} & \emph{PROTECTION OF PERSONAL DATA} & \\\midrule
        & \emph{yes}  & \emph{no} \\ 
     Does your project involve personal data collection and/or processing?
     & \checkmark & \\ 
     Does it involve the collection and/or processing of sensitive personal data (e.g. health, sexual lifestyle, ethnicity, political opinion, religious or philosophical conviction)?
     & &\checkmark  \\  
     Does it involve processing of genetic information?
     & &\checkmark \\  
     Does it involve tracking or observation of particispants? It should be noted that this issue is not limited to surveillance or localization data. It also applies to Wan data such as IP address, MACs, cookies etc.
     & \checkmark & \\  
     Does your project involve further processing of previously collected personal data (secondary use)? For example Does your project involve merging existing data sets?
     & \checkmark & \\\toprule
     
     \emph{Section 5} & \emph{ANIMALS} & \\\midrule
        & \emph{yes}  & \emph{no} \\ 
     Does your project involve animals? &  & \checkmark \\
    \toprule
  \emph{Section 6} & \emph{DEVELOPING COUNTRY} & \\\midrule
        & \emph{yes}  & \emph{no} \\ 
     Does your project involve developing countries? & &\checkmark \\
     If your project involves low and/or lower-middle income countries, are any benefit-sharing actions planned? & &\checkmark \\\toprule
     \emph{Section 7} & \emph{ENVIRONMENT PROTECTION AND SAFETY} & \\\midrule
        & \emph{yes}  & \emph{no} \\ 
      Does your project involve the use of elements that may cause harm to the environment, animals or plants? & &\checkmark \\
      Does your project deal with endangered fauna and/or flora /protected areas? & &\checkmark \\     
      Does your project involve the use of elements that may cause harm to humans, including project staff? & &\checkmark \\          
      Does your project involve other harmful materials or equipment, e.g. high-powered laser systems? & &\checkmark \\\toprule
      \emph{Section 8} & \emph{DUAL USE} & \\\midrule 
             & \emph{yes}  & \emph{no} \\ 
       Does your project have the potential for military applications? & &\checkmark \\
       Does your project have an exclusive civilian application focus? & &\checkmark \\      
       Will your project use or produce goods or information that will require export licenses in accordance with legislation on dual use items? & &\checkmark \\\toprule
           \emph{Section 9} & \emph{MISUSE} & \\\midrule 
                  & \emph{yes}  & \emph{no} \\ 
Does your project have the potential for malevolent/criminal/terrorist abuse?& &\checkmark \\ 
Does your project involve information on/or the use of biological-, chemical-, nuclear/radiological-security sensitive materials and explosives, and means of their delivery?& &\checkmark \\            
Does your project involve the development of technologies or the creation of information that could have severe negative impacts on human rights standards (e.g. privacy, stigmatization, discrimination), if misapplied?& &\checkmark \\            
Does your project have the potential for terrorist or criminal abuse e.g. infrastructural vulnerability studies, cybersecurity related project?& &\checkmark \\\toprule
  \emph{Section 10} & \emph{LEGAL ISSUES} & \\\midrule
               & \emph{yes}  & \emph{no} \\ 
Will your project use or produce software for which there are copyright licensing implications?& &\checkmark \\\toprule 
Will your project use or produce goods or information for which there are data protection, or other legal implications?& &\checkmark \\\toprule
  \emph{Section 11} & \emph{OTHER ETHICS ISSUES} & \\\midrule
               & \emph{yes}  & \emph{no} \\ 
Are there any other ethics issues that should be taken into consideration?& &\checkmark \\
		\bottomrule
	\end{tabularx}
\chapter{Identities Appeared in Extended Database}
\begin{figure}[H]
\centering
\includegraphics[height=13cm,width=13cm]{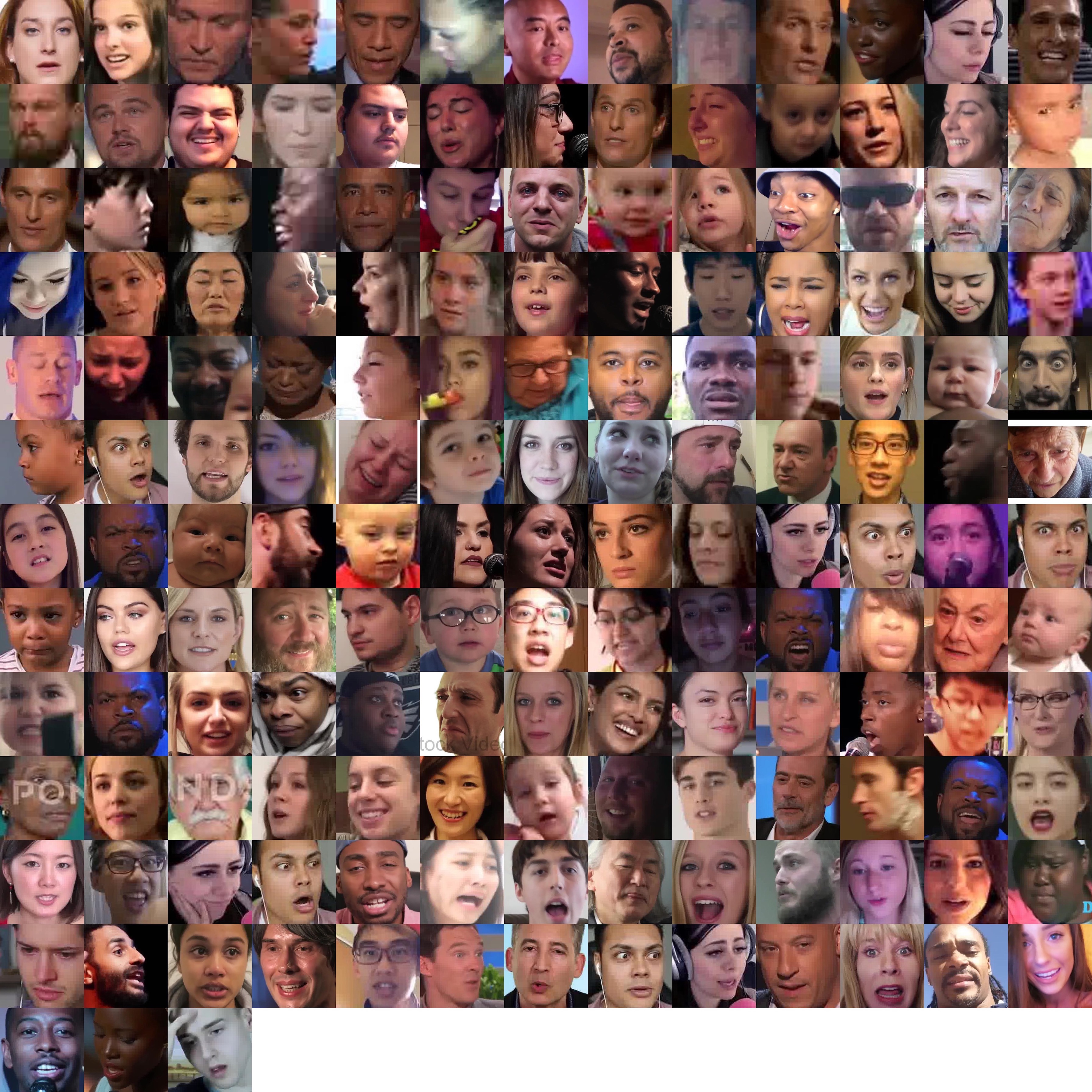}
\caption{All identities appeared in collected database.}
\end{figure}

\chapter{Link List of Videos}
1, \url{https://www.youtube.com/watch?v=enk0bOv-gF8} \newline
2, \url{https://www.youtube.com/watch?v=67SeR3LxtdI} \newline
3, \url{https://www.youtube.com/watch?v=2C4c7VR8X7I} \newline
4, \url{https://www.youtube.com/watch?v=9ECJaErtCIs} \newline
5, \url{https://www.youtube.com/watch?v=JlfYA0xcbVw} \newline
6, \url{https://www.youtube.com/watch?v=KJwgxhvNd9U} \newline
7, \url{https://www.youtube.com/watch?v=wnHOWnxUxO8} \newline
8, \url{https://www.youtube.com/watch?v=x2U9HZj5GLI} \newline
9, \url{https://www.youtube.com/watch?v=3EuDj5E1JWY} \newline
10, \url{https://www.youtube.com/watch?v=Jh6o8kMbWfg} \newline
11, \url{https://www.youtube.com/watch?v=K3YnaQ64ELY} \newline
12, \url{https://www.youtube.com/watch?v=aqu4ezLQEUA} \newline
13, \url{https://www.youtube.com/watch?v=XxoY1bGf_u4} \newline
14, \url{https://www.youtube.com/watch?v=Zr16eHb4YNU} \newline
15, \url{https://www.youtube.com/watch?v=-Ktdf2KQ58c} \newline
16, \url{https://www.youtube.com/watch?v=1gSCEImUey8} \newline
17, \url{https://www.youtube.com/watch?v=oEwBoxygBuY} \newline
18, \url{https://www.youtube.com/watch?v=P_-PxVv2gn4} \newline
19, \url{https://www.youtube.com/watch?v=lJBo9jdUJiY} \newline
20, \url{https://www.youtube.com/watch?v=uhfRn46mqLo} \newline
21, \url{https://www.youtube.com/watch?v=bSoITsaSs0M} \newline
22, \url{https://www.youtube.com/watch?v=NJXraKM94EE} \newline
23, \url{https://www.youtube.com/watch?v=OTYrFAElzww} \newline
24, \url{https://www.youtube.com/watch?v=8qqtC1KxSSo} \newline
25, \url{https://www.youtube.com/watch?v=tA5giyG8E7g} \newline
26, \url{https://www.youtube.com/watch?v=RDocnbkHjhI} \newline
27, \url{https://www.youtube.com/watch?v=XvgnOqcCYCM} \newline
28, \url{https://www.youtube.com/watch?v=Zi_k7H43Y-M} \newline
29, \url{https://www.youtube.com/watch?v=Y6_ZQeDGaaQ} \newline
30, \url{https://www.youtube.com/watch?v=xbIF9I-eV2w} \newline
31, \url{https://www.youtube.com/watch?v=vnKZ4pdSU-s} \newline
32, \url{https://www.youtube.com/watch?v=sq1l-19pwS4} \newline
33, \url{https://www.youtube.com/watch?v=Eu_Gl0woeOw} \newline
34, \url{https://www.youtube.com/watch?v=Ln0SCsZPZIc} \newline
35, \url{https://www.youtube.com/watch?v=CiITxUUtzLI} \newline
36, \url{https://www.youtube.com/watch?v=VMLVUkRB2PM} \newline
37, \url{https://www.youtube.com/watch?v=uQQA8qWYmu4} \newline
38, \url{https://www.youtube.com/watch?v=OJGJ8cFd3SY} \newline
39, \url{https://www.youtube.com/watch?v=nl_PRXqgiBY} \newline
40, \url{https://www.youtube.com/watch?v=Ei6kejNICWA} \newline
41, \url{https://www.youtube.com/watch?v=DHL7iOd_FXg} \newline
42, \url{https://www.youtube.com/watch?v=znFG8nYHd_8} \newline
43, \url{https://www.youtube.com/watch?v=FKFy_pyK7E8} \newline
44, \url{https://www.youtube.com/watch?v=_tME5zuHvjM} \newline
45, \url{https://www.youtube.com/watch?v=Qm8qKdp-AIs} \newline
46, \url{https://www.youtube.com/watch?v=apzXGEbZht0} \newline
47, \url{https://www.youtube.com/watch?v=7FC4qRD1vn8} \newline
48, \url{https://www.youtube.com/watch?v=ZtUPKekDY7M} \newline
49, \url{https://www.youtube.com/watch?v=kwCrQ5SQaL0} \newline
50, \url{https://www.youtube.com/watch?v=DwLIIqRBQoM} \newline
51, \url{https://www.youtube.com/watch?v=S3V3_PGJvVk} \newline
52, \url{https://www.youtube.com/watch?v=bfiFAbMLTwc} \newline
53, \url{https://www.youtube.com/watch?v=g1fODA0MoOI} \newline
54, \url{https://www.youtube.com/watch?v=qWw6TSQm3hg} \newline
55, \url{https://www.youtube.com/watch?v=_oTEvAasllc} \newline
56, \url{https://www.youtube.com/watch?v=e2R0NSKtVA0} \newline
57, \url{https://www.youtube.com/watch?v=wJiS0UrA68U} \newline
58, \url{https://www.youtube.com/watch?v=-gSYKgAMPBc} \newline
59, \url{https://www.youtube.com/watch?v=Sjn5uqDxLPs} \newline
60, \url{https://www.youtube.com/watch?v=S0ssoXyQ7Ek} \newline
61, \url{https://www.youtube.com/watch?v=wBgm-9YS74s} \newline
62, \url{https://www.youtube.com/watch?v=Cp7EiCDzZ6c} \newline
63, \url{https://www.youtube.com/watch?v=3aRAEy-hZWg} \newline
64, \url{https://www.youtube.com/watch?v=2bINpR70WsI} \newline
65, \url{https://www.youtube.com/watch?v=TxCw9gB7jQ4} \newline
66, \url{https://www.youtube.com/watch?v=KcWn1bPCZVs} \newline
67, \url{https://www.youtube.com/watch?v=5Z-RGqsSNos} \newline
68, \url{https://www.youtube.com/watch?v=aGEFtRwPhE4} \newline
69, \url{https://www.youtube.com/watch?v=IOvRksXtNOM} \newline
70, \url{https://www.youtube.com/watch?v=pQZpjW5m3qo} \newline
71, \url{https://www.youtube.com/watch?v=LhlDeCUQ6iE} \newline
72, \url{https://www.youtube.com/watch?v=4IR7A-xeWBo} \newline
73, \url{https://www.youtube.com/watch?v=9WuaBJA-e4U} \newline
74, \url{https://www.youtube.com/watch?v=C3hABRHmQoo} \newline
75, \url{https://www.youtube.com/watch?v=8m173M3mG1w} \newline
76, \url{https://www.youtube.com/watch?v=ahlm-91krew} \newline
77, \url{https://www.youtube.com/watch?v=wD2cVhC-63I} \newline
78, \url{https://www.youtube.com/watch?v=0lescn84_ck} \newline
79, \url{https://www.youtube.com/watch?v=JhRW8J3gTuo} \newline
80, \url{https://www.youtube.com/watch?v=aZCrB9jrAiY} \newline
81, \url{https://www.youtube.com/watch?v=mJBE_PkK6hM} \newline
82, \url{https://www.youtube.com/watch?v=_XfUUYK7Gkg} \newline
83, \url{https://www.youtube.com/watch?v=ubSWWqDpldE} \newline
84, \url{https://www.youtube.com/watch?v=LkoOCw_tp1I} \newline
85, \url{https://www.youtube.com/watch?v=e9yUXVzs0Qw} \newline
86, \url{https://www.youtube.com/watch?v=eGxPGgPdTVw} \newline
87, \url{https://www.youtube.com/watch?v=fcfQkxwz4Oo} \newline
88, \url{https://www.youtube.com/watch?v=m2eyq9qTOQY} \newline
89, \url{https://www.youtube.com/watch?v=aatoBIfYmEc} \newline
90, \url{https://www.youtube.com/watch?v=lb7_tPl1bOM} \newline
91, \url{https://www.youtube.com/watch?v=E1a231faW-k} \newline
92, \url{https://www.youtube.com/watch?v=5w3cYtJekpw} \newline
93, \url{https://www.youtube.com/watch?v=bAHfEYBeb-Q} \newline
94, \url{https://www.youtube.com/watch?v=3mOB_BCjiHw} \newline
95, \url{https://www.youtube.com/watch?v=NyBqfzs-Gzk} \newline
96, \url{https://www.youtube.com/watch?v=lFC-6rjjHo0} \newline
97, \url{https://www.youtube.com/watch?v=elPCZSmiEsU} \newline
98, \url{https://www.youtube.com/watch?v=xemqxOtX-OE} \newline
99, \url{https://www.youtube.com/watch?v=SigyAbMH2xA} \newline
100, \url{https://www.youtube.com/watch?v=P4ramoioWnw} \newline
101, \url{https://www.youtube.com/watch?v=SN0cLBv5AUc} \newline
102, \url{https://www.youtube.com/watch?v=9HuTKuMCUpU} \newline
103, \url{https://www.youtube.com/watch?v=rLrAkJVlXVY} \newline
104, \url{https://www.youtube.com/watch?v=OVx02fA6soc} \newline
105, \url{https://www.youtube.com/watch?v=RcfEr6jSbNw} \newline
106, \url{https://www.youtube.com/watch?v=3Tx5_STuU1s} \newline
107, \url{https://www.youtube.com/watch?v=8v8UfHL0v10} \newline
108, \url{https://www.youtube.com/watch?v=0vlHCR9IYQ8} \newline
109, \url{https://www.youtube.com/watch?v=syKWtVVvdEI} \newline
110, \url{https://www.youtube.com/watch?v=xvh1WFJRoGM} \newline
111, \url{https://www.youtube.com/watch?v=Bsnv7B77noY} \newline
112, \url{https://www.youtube.com/watch?v=vTyLSr_VCcg} \newline
113, \url{https://www.youtube.com/watch?v=xpyrefzvTpI} \newline
114, \url{https://www.youtube.com/watch?v=73fz_uK-vhs} \newline
115, \url{https://www.youtube.com/watch?v=3wUcC2QhqTg} \newline
116, \url{https://www.youtube.com/watch?v=lm6DFV81H6g} \newline
117, \url{https://www.youtube.com/watch?v=I21LyTgacqU} \newline
118, \url{https://www.youtube.com/watch?v=2QUacU0I4yU} \newline
119, \url{https://www.youtube.com/watch?v=pzYZ_sMW6Eo} \newline
120, \url{https://www.youtube.com/watch?v=iCwKM6uB71I} \newline
121, \url{https://www.youtube.com/watch?v=SCHK12XRqE4} \newline
122, \url{https://www.youtube.com/watch?v=Rr1mssQjF6I} \newline
123, \url{https://www.youtube.com/watch?v=osMhAT3pkT0} \newline
124, \url{https://www.youtube.com/watch?v=5z9aTSlmM7w} \newline
125, \url{https://www.youtube.com/watch?v=fTXhgr7I6Fg} \newline
126, \url{https://www.youtube.com/watch?v=NSFA7FQ7bYE} \newline
127, \url{https://www.youtube.com/watch?v=wyPWD1dPtPc} \newline
128, \url{https://www.youtube.com/watch?v=zeufCmtjnV4} \newline
129, \url{https://www.youtube.com/watch?v=tptUAgH22BU} \newline
130, \url{https://www.youtube.com/watch?v=sUDXHuv_JKk} \newline
131, \url{https://www.youtube.com/watch?v=sGnM0BoCZ0M} \newline
132, \url{https://www.youtube.com/watch?v=8xlMd5R2-pI} \newline
133, \url{https://www.youtube.com/watch?v=2p5SlSiGOMU} \newline
134, \url{https://www.youtube.com/watch?v=qROLM75gU-Q} \newline
135, \url{https://www.youtube.com/watch?v=xUp_G4BpwB4} \newline
136, \url{https://www.youtube.com/watch?v=aHTQwwZug9M} \newline
137, \url{https://www.youtube.com/watch?v=NgEwFnvVbLo} \newline
138, \url{https://www.youtube.com/watch?v=0mz59i898HU} \newline
139, \url{https://www.youtube.com/watch?v=Q13z_WCr1TU} \newline
140, \url{https://www.youtube.com/watch?v=bGvHKE0MYVQ} \newline

\bibliographystyle{apa}
\bibliography{bibs/sample,bibs/mendeley}

\end{document}